\documentclass[lettersize,journal]{IEEEtran}
\usepackage{amsmath,amsfonts}
\usepackage{float}
\usepackage{multirow}
\usepackage{makecell}
\usepackage{array}
\usepackage[caption=false,font=normalsize,labelfont=sf,textfont=sf]{subfig}
\usepackage{textcomp}
\usepackage{stfloats}
\usepackage{url}
\usepackage{verbatim}
\usepackage{graphicx}
\usepackage{cite}
\usepackage{stfloats}
\usepackage{booktabs}
\usepackage{subfig}
\usepackage[ruled,linesnumbered]{algorithm2e}
\usepackage{caption}
\usepackage{hyperref}
\hypersetup{hidelinks,
	colorlinks=true,
	allcolors=black,
	pdfstartview=Fit,
	breaklinks=true}

\usepackage[capitalize]{cleveref}
\crefname{section}{Sec.}{Secs.}
\Crefname{section}{Section}{Sections}
\Crefname{table}{Table}{Tables}
\crefname{table}{Tab.}{Tabs.}
\begin{document}

\title{ASF-Net: Robust Video Deraining via Temporal Alignment and Online Adaptive Learning}

\author{Xinwei Xue, Jia He, Long Ma, Xiangyu Meng, Wenlin Li, Risheng Liu,~\IEEEmembership{Member,~IEEE}
%\author{IEEE Publication Technology,~\IEEEmembership{Staff,~IEEE,}
%\thanks{This paper was produced by the IEEE Publication Technology Group. They are in Piscataway, NJ.}% <-this % stops a space
\thanks{This work is partially supported by the National Natural Science Foundation of China (Nos. 61806036, 61922019, 61932020 and 61733002), the National
Key R\&D Program of China(2020YFB1313503), LiaoNing Revitalization Talents Program (XLYC1807088), and the Fundamental Research Funds for the Central Universities (Corresponding author: rsliu@dlut.edu.cn).

X. Xue and R. Liu are with the DUT-RU International School of Information Science \& Engineering, Dalian University of Technology, and also with the Engineering and Key Laboratory for Ubiquitous Network and Service Software of Liaoning Province, Dalian University of Technology, Dalian, 116024, China. (email: xuexinwei@dlut.edu.cn, rsliu@dlut.edu.cn).

J. He, L. Ma and X. Meng are with the School of Software Technology, Dalian University of Technology, Dalian, 116024, China. (email: hejia@mail.dlut.edu, malone94319@gmail.com, lnmengxiangyu@gmail.com). 

W. Li is with the DUT-RU International School of Information Science \& Engineering, Dalian University of Technology, Dalian, 116024, China. (email: wenlinli661@gmail.com ). 

%L. Ma is with the School of Software Technology, Dalian University of Technology, Dalian, 116024, China. (email: malone94319@gmail.com).
}}

%R. Liu is with the DUT-RU International School of Information Science \& Engineering, Dalian University of Technology, and also with the Engineering and Key Laboratory for Ubiquitous Network and Service Software of Liaoning Province, Dalian University of Technology, Dalian, 116024, China. (email: rsliu@dlut.edu.cn).}}
%\thanks{Manuscript received April 19, 2021; revised August 16, 2021.}}

\markboth{Journal of \LaTeX\ Class Files,~Vol.~14, No.~8, August~2021}%
{Shell \MakeLowercase{\textit{et al.}}: A Sample Article Using IEEEtran.cls for IEEE Journals}


\maketitle
\begin{figure*}[!t]
     \centering
     \begin{tabular}{c@{\extracolsep{0.25em}}c@{\extracolsep{0.25em}}c@{\extracolsep{0.25em}}c@{\extracolsep{0.25em}}c}%@{\extracolsep{0.5em}}c
		\includegraphics[width=0.192\linewidth]{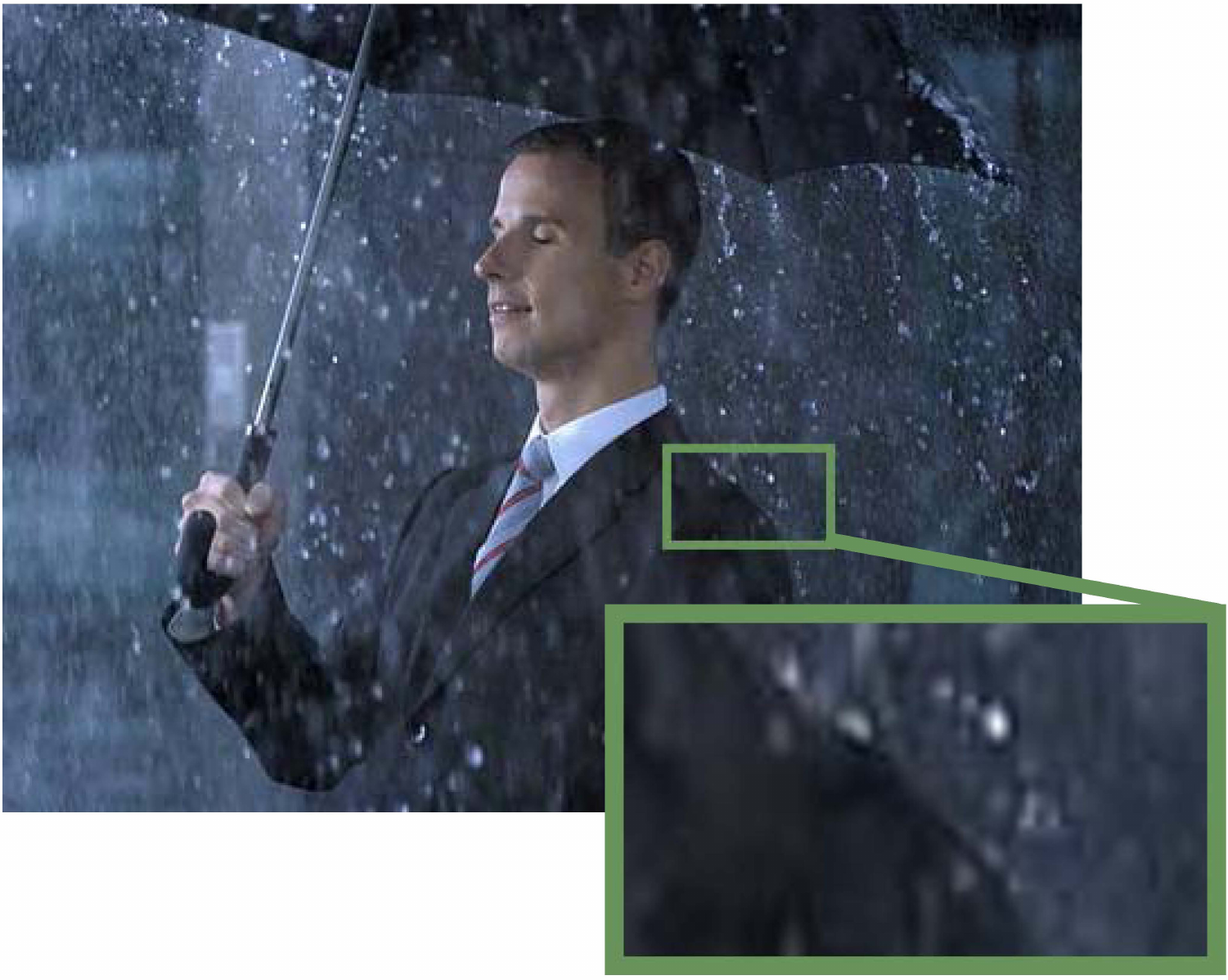}&
		\includegraphics[width=0.192\linewidth]{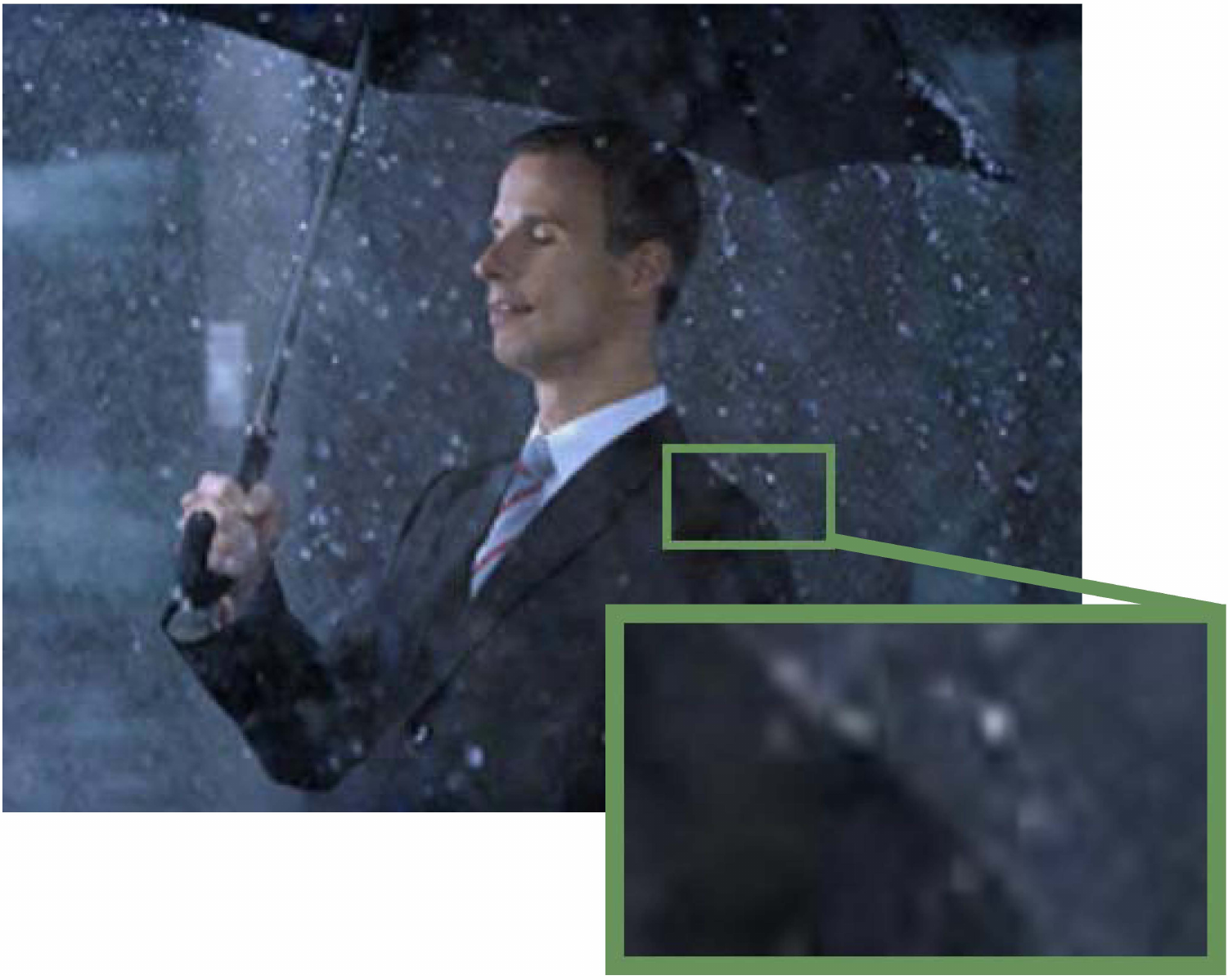}&
		\includegraphics[width=0.192\linewidth]{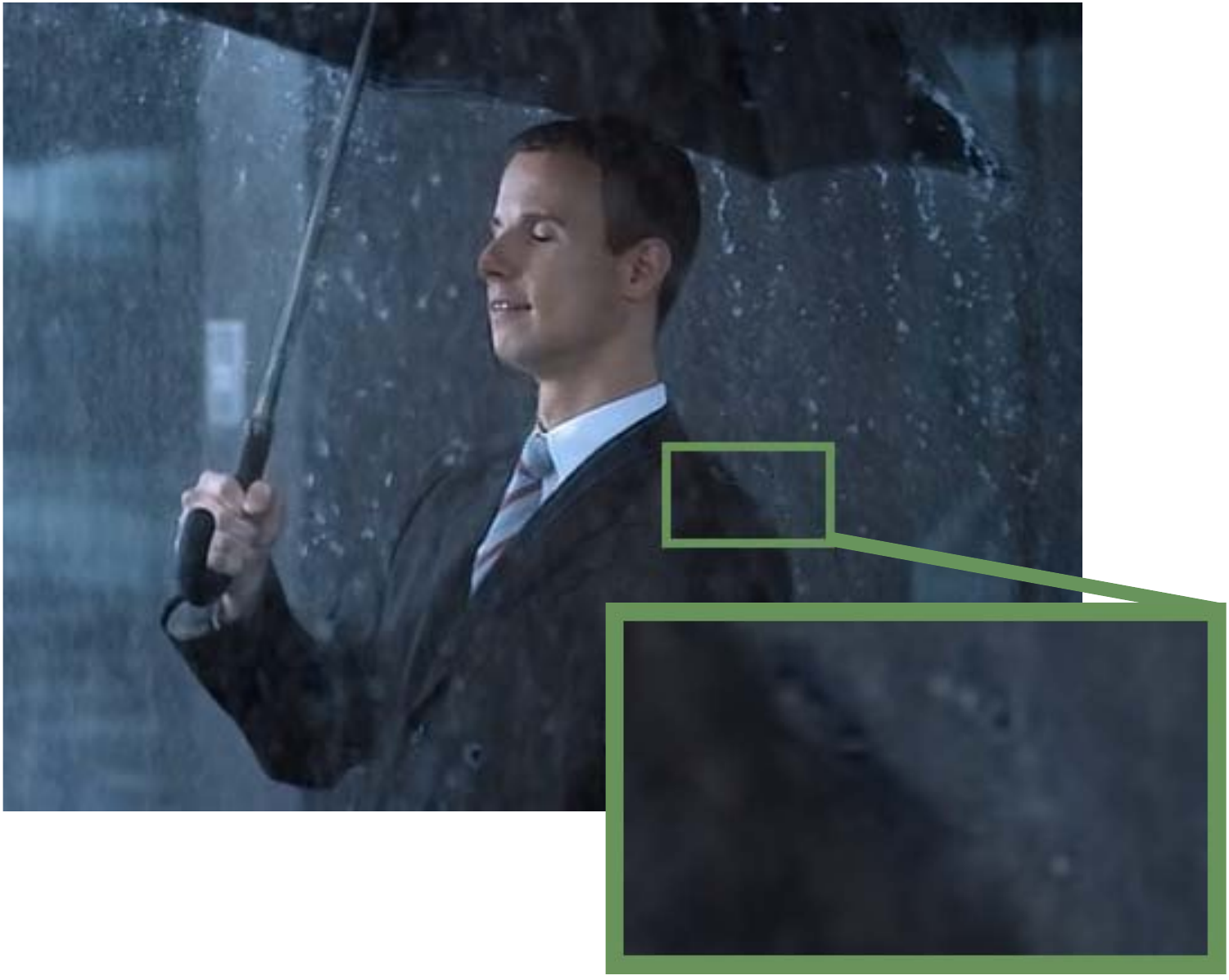}&
		\includegraphics[width=0.192\linewidth]{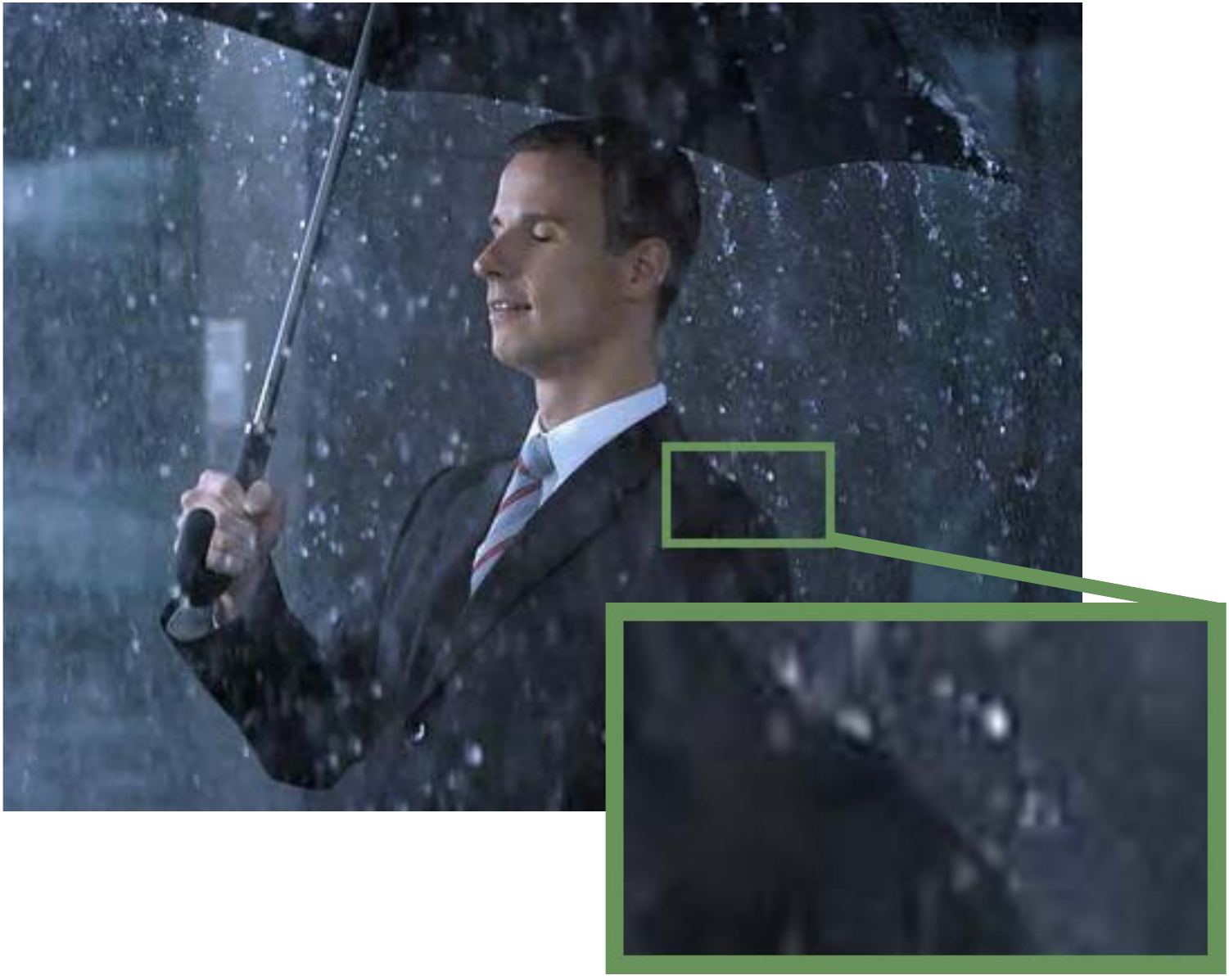}	&
		\includegraphics[width=0.192\linewidth]{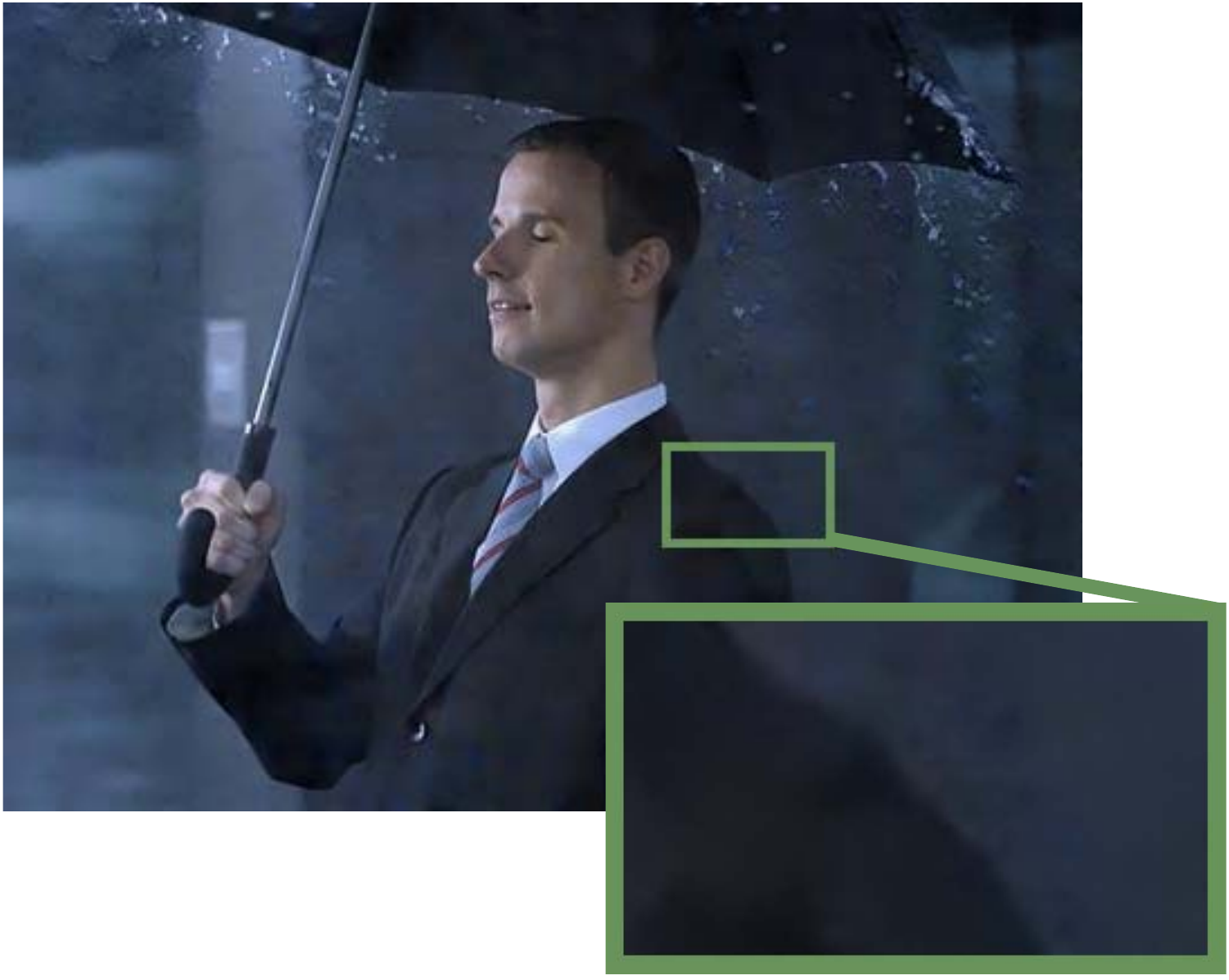}\\
		\footnotesize Rain Frame&\footnotesize J4RNet~\cite{yang2019joint} (\textit{TPAMI '19}) &\footnotesize SLDNet~\cite{yang2020self} (\textit{CVPR '20})  &\footnotesize S2VD~\cite{yue2021semi} (\textit{CVPR '21}) &\footnotesize Ours
	\end{tabular}
	%\vspace{-0.3cm}	
	\caption{Video deraining results on a real-world example. All advanced methods fail to remove visible rain streaks, while our method realizes a better visual quality with fewer rain streaks and clearer background.}
	%\vspace{0.2cm}
	\label{fig:first}
\end{figure*}

\begin{abstract}
%Recently, learning-based video deraining methods have achieved excellent results, however, there still exist two critical challenges that need to be solved, i.e., the temporal exploitation among adjacent frames and the model's adaptability towards unknown real-world scenarios. To handle these challenges, we investigate video deraining from paradigm design to learning strategy construction. Concretely, we establish a new computational paradigm (Alignment-Shift-Fusion Network, named ASF-Net) by introducing a temporal shift module that has never appeared in this area to better exploit temporal correlations among adjacent frames. The temporal shift module delves deep into temporal information by exchanging channel-level information on the feature space.
In recent times, learning-based methods for video deraining have demonstrated commendable results. However, there are two critical challenges that these methods are yet to address: exploiting temporal correlations among adjacent frames and ensuring adaptability to unknown real-world scenarios. To overcome these challenges, we explore video deraining from a paradigm design perspective to learning strategy construction. Specifically, we propose a new computational paradigm, Alignment-Shift-Fusion Network (ASF-Net), which incorporates a temporal shift module. This module is novel to this field and provides deeper exploration of temporal information by facilitating the exchange of channel-level information within the feature space. 
To fully discharge the model's characterization capability, we further construct a LArge-scale RAiny video dataset (LARA) which also supports the development of this community. On the basis of the newly-constructed dataset, we explore the parameters learning process by developing an innovative re-degraded learning strategy. This strategy bridges the gap between synthetic and real-world scenes, resulting in stronger scene adaptability. Our proposed approach exhibits superior performance in three benchmarks and compelling visual quality in real-world scenarios, underscoring its efficacy. 
The code is available at \url{https://github.com/vis-opt-group/ASF-Net}.
\end{abstract}


\section{Introduction}
%\IEEEPARstart{R}{ain}, as a common severe natural weather, often causes adverse effects such as visual quality degradation and content information obscuration during outdoor sports video shooting, which affects the quality of subsequent tasks on video such as video super-resolution, video frame interpolation, video segmentation, etc. And also videos captured by outdoor cameras are inevitably affected by rain streaks during rainy days, producing visibility degradation to limit the perception ability of some intelligence devices. So it is extremely essential to remove rain streaks from videos. In the following, we first introduce our contributions, then provide a comprehensive review of related works. 
\IEEEPARstart{R}{ain} streaks inevitably affect the clarity of videos captured in outdoor rainy scenes, causing visibility degradation and even occlusion of the background content. Moreover, with many deep learning-based works~\cite{liu2012fixed, wu2019essential, liu2021retinex, liu2021learning, liu2020real, piao2019depth, piao2020a2dele, zhang2020select, ma2023bilevel, LiuLZFL20, LiuLFZHL22, abs-2203-06810, ZhangLLLFL20, LiLLLF22} emerging in the computer vision field, rain streaks can also significantly impair the performance of some computer vision tasks that mainly rely on clean video frames, such as object detection, tracking, semantic segmentation, video instance segmentation, surveillance, and autonomous navigation. Consequently, there is a growing interest in developing effective techniques to remove rain streaks from videos. 

Different from single image deraining methods, video deraining methods remove rain streaks from the target frame with the help of its adjacent frames. For example,~\cite{jiang2018fastderain, li2018video, chen2018robust, liu2018erase, liu2018d3r, liu2019removing, yang2019frame, xue2020sequential, yang2020self} exploit spatial and temporal redundancy information for video rain streaks removal. Some works ~\cite{jiang2018fastderain, li2018video, xue2020sequential} are built on physical models, such as sparse coding, multi-scale characteristic of streaks and unrolling theory. Others~\cite{chen2018robust, liu2018erase, liu2018d3r, yang2019frame, yang2020self, xue2020sequential} designed specific deep networks to utilize the spatial and temporal information inherent in the rainy video.
%With the rapid development of deep learning and neural networks, video rain removal methods have been innovated from a rain detector based on the photometric appearance of rain~\cite{2004Detection} first proposed by Garg and Naya. 
%Nevertheless, there are still several specific problems with the main methods used in the current phase. First, due to the special inter-frame timing of the video, the utilization and recovery of temporal information have been the most important part of the video rain removal work, but the precise temporal information has been difficult to achieve a real fusion; second, the training strategy for the network is relatively limited, and the test results for real and synthetic rain data under the same conditions still differ greatly. 
Despite the significant progress made by learning-based video deraining methods, there are still several problems with these methods in the current phase. Primarily, the inter-frame difference of the video complicates the utilization and recovery of temporal information, which is a crucial issue of video rain removal work. However, achieving the precise integration of temporal information is very challenging, which may result in sub-optimal results. Additionally, most approaches still employ training strategies with some limitation, which can result in significant discrepancies between the synthetic and real rain data under the same conditions. 

%To solve the above challenges, we develop a new ASF-Net to delve deep into video deraining, especially in unknown real-world scenarios, which makes the full use of temporal information through the temporal shift module and utilizes a new learning strategy to improve the generalization of the network and a new dataset is constructed to improve the effect on both real and synthetic rain datasets. 
To address the aforementioned challenges, we propose a new framework called \textbf{ASF-Net}, which includes a temporal \textbf{A}lignment module, a novel temporal \textbf{S}hift module, and a temporal \textbf{F}usion module. In addition, we also incorporate a new learning strategy to enhance the generalization capability of ASF-Net, enabling it to process the previously unseen or unexpected scenarios better. Furthermore, we conducted a new video deraining dataset to improve the effectiveness of ASF-Net on both synthetic and real-world rain scenarios. These key advancements make ASF-Net a highly effective video deraining tool with remarkable versatility and robustness. 
As shown in Fig.~\ref{fig:first}, the methods J4RNet~\cite{yang2019joint} and S2VD~\cite{yue2021semi} cannot remove most rain streaks in the real-world scenario. SLDNet~\cite{yang2020self} indeed eliminates some rain streaks but produces unknown blurs and damages the visual quality. Compared with them, ours is clearer and more natural. 

In summary, our contributions can be concluded as the following four folds.
\begin{itemize}
	\item We introduce a novel video deraining paradigm, alignment-shift-fusion, which can make full use of the temporal information in the video data by a novel temporal shift module and exchange the feature space between adjacent frames at the channel-level. To the best of our knowledge, this is the first time such an approach has been utilized after motion-estimate alignment. 
%	We propose a new paradigm for video deraining named by alignment-shift-fusion. For the first time after motion-estimate alignment, the underutilized temporal information is fully exploited by the temporal shift module, exchanging the feature space of neighbor frames at the channel-level. 

	\item To fully leverage the capabilities of our video deraining paradigm, we have constructed a large-scale rainy video dataset, named \textbf{LA}rge-scale \textbf{RA}iny Video Dataset (\textbf{LARA}). LARA contains 1000 synthetic videos with diverse rain streaks, and 120 real-world videos collected from the Internet. 
%	To fully utilize the model capabilities, we built a \textbf{LA}rge-scale \textbf{RA}iny video dataset (\textbf{LARA}) that contains 1000 synthetic videos with different intensities of rain streaks and 120 real-world videos collected from the web. We demonstrate the effectiveness of LARA from different aspects. This will invigorate the video deraining. 
	
	\item We propose a new online learning strategy for video deraining based on the LARA dataset. This approach establishes a correspondence between the synthetic and real-world domains through a re-degraded augmentation mechanism, which can improve the performance of our deraining algorithm and simulate the real-world scenarios naturally. 
%	To enhance the adaptability to different scenarios and to delve into the parameter learning process, we are the first to propose a new learning strategy based on the LARA dataset. Unlike the end-to-end training approach, this new online learning strategy establishes the correspondence between the real and synthetic domains through a re-degraded augmentation mechanism, which is more applicable to real-world scenarios. 
	
	\item We conducted extensive experiments to evaluate the effectiveness of our alignment-shift-fusion paradigm for video deraining. Experimental results demonstrate  the superior performance and scene adaptability of ours compared to other state-of-the-art methods. A series of ablation experiments further validate the critical role of the temporal shift module in better utilizing temporal information. These results provide strong evidence for the effectiveness of our proposed method. 
%	Extensive experiments are performed to verify our superiority against other state-of-the-art methods on different benchmarks, with excellent scene adaptability. A series of ablation experiments proved the importance of introducing the temporal shift module for better utilization of temporal information. 
	
\end{itemize}

%Admittedly, there are some limitations to our work. For example, in a real-world heavy rain scenario, some rain streaks may remain using our method, but the performance is still superior compared to other SOTA methods. 
%Moving forward, we plan to continue improving the rain removal effectiveness on heavy rain data and enhance the generalization ability to adapt to a wider range of real-world scenarios. 

\section{Related Work}
Designing an effective architecture by stacking different network layers is a widespread manner in multimedia and vision field, there is no exception in the field of video deraining. Existing works can be roughly divided into designing the network architecture and constructing the learning strategy. 

{\bf{Designing the network architecture}}. Learning-based methods have made great progress in video deraining~\cite{ma2021video, fan2022video, chen2018robust, liu2018erase, liu2018d3r, liu2019removing, yang2019frame, yang2020self, xue2021temporal, xue2021gta, yan2021self, yue2021semi}. They commonly follow the paradigm of ``alignment + fusion" to integrate the temporal information by the former and derive the deraining results by the latter. 

%The utilization of optical flow~\cite{dosovitskiy2015flownet,2017FlowNet,2019Video,2019Optical} is common to extract the temporal information between adjacent frames of video efficiently. 

Kim~\emph{et al.}~\cite{2014Single} firstly introduced an optical flow estimation algorithm to achieve the alignment and fusion of inter-frame information to process the temporal information for video deraining. Yang~\emph{et al.} in~\cite{yang2019frame} constructed a two-stage recurrent network, initially estimating the rain-free results of a single frame based on a novel rain synthesis model in the first stage, and then leveraging the motion information of adjacent frames in the second stage through an ``alignment + fusion" paradigm. 
%The estimated rain-free results in the first stage are used as guidance for optical flow alignment and motion modeling. 
Moreover, this schema is also used in~\cite{xue2021temporal}, where the temporal consistency is maintained by optical flow alignment after single-frame enhancement. 

In addition, to achieve better alignment results, researchers have come up with other methods. 
Yan \emph{et al.}~\cite{yan2021self} introduced an end-to-end rain removal network that employs deformable convolution to achieve feature-level alignment. Such a technique holds promise in video deraining research by mitigating the impact of rain streaks on motion estimation and enabling more effective temporal exploitation of adjacent frames.  
Chen \emph{et al.}~\cite{chen2018robust} proposed a spatial-temporal content alignment algorithm at super-pixel levels to improve the performance in high-speed moving scenes. 

Unfortunately, the above methods acquire the aligned results by alignment network, which perhaps leads to the insufficient exploitation of temporal information. 
%For the optical flow-based approaches~\cite{dosovitskiy2015flownet,2017FlowNet,2019Video,2019Optical}, the brightness of rain streaks in the video is often higher than the background, leading to serious error of the optical flow of background area. And the optical flow dataset contains more motion and no rain streaks occlusion, so directly migrating the optical flow model to the rain removal model can be harmful for deraining. 
%As for the super-pixel-based methods, the super-pixel module and the deep network behind cannot achieve the end-to-end training. For the direct method with deformable convolutional alignment, the learning of offsets during training is highly unstable~\cite{wang2019edvr}. 
Therefore, in our work, we create a new paradigm for temporal exploitation, i.e., ``alignment-shift-fusion'' to fully utilize the information exchanged among neighbor frames on channel-level in the feature space. 

{\bf{Constructing the learning strategy}}. 
After establishing the network architecture, an effective training scheme must be designed to optimize the network to improve generalization in various scenarios. A commonly employed approach is training in an end-to-end manner, as has been observed in several prior works~\cite{liu2018erase,liu2018d3r,yang2019frame}. 
%The end-to-end approach helps train the model to transform the input data into the desired output, thus avoiding the complexity that may arise due to the deployment of multiple modules. By minimizing the reliance on complex network modules, this approach helps to simplify the training process and reduce complexity.
%However, the performance of this approach depends heavily on the amount and distribution of training data and can easily lead to undesirable gains or even loss of efficacy in some real-world scenarios.

%Some works attempt to construct an effective connection between synthetic and real scenarios to provide a better solution for practical environments. 
To improve the performance of learning-based models in real-world scenarios, some studies focused on bridging the gap between synthetic and actual scenarios. 
Wei \emph{et al.}~\cite{wei2019semi} proposed a semi-supervised transfer learning framework that adds unlabeled real rain images to the training dataset, and considered the residual between rain and non-rain images as a parameterized rain distribution.
%and the network can be adapted to real unsupervised multiple rain types by supervised synthetic rain data. 
This method significantly alleviates the discrepancy problem between real and synthetic rain images. 
%Zhu \emph{et al.}~\cite{zhu2019singe} proposed an unsupervised rain removal model and multi-scale depth supervised discriminator. The unsupervised rain removal model is based on a multi-scale attention memory generator that fuses contextual information from densely connected networks at different scales and learns rain streaks using unpaired training data, thus recovering rain-free images. The multi-scale depth supervised discriminator further recovers different details of the images. 
Guo~\emph{et al.}~\cite{guo2022derainattentiongan} constructed an unsupervised rainwater extraction model guided by an attention mechanism. This model extracts rainwater from the rain data set and uses the CycleGAN cycle structure to remove rain. Yue \emph{et al.}~\cite{yue2021semi} proposed a semi-supervised method and employed a dynamic rain generator to fit the rain layer. %This model uses the labeled synthetic signed data as a strong constraint on the prior distribution, and encodes the temporal consistency of the background by Markov random fields for the unlabeled real data. 

Despite the significant progress made in learning-based video deraining methods, there are still critical challenges that need to be addressed. Notably, these methods tend to focus on generating high-quality synthetic rainy data, limiting the adaptability to real-world rainy scenarios. 
%Additionally, even with the use of semi-supervision, these methods do not effectively extract useful information from real-world data. 
Furthermore, the quality of unlabeled data is not always fully considered, and many unlabeled samples are not utilized efficiently.
% In unsupervised algorithms, there is a high sensitivity to feature selection, and lacking diverse features in the dataset can significantly impact the performance of the final model. These limitations highlight the inefficiency in selecting and utilizing the vast amounts of real-world scene information. 

%Different from them, we bridge the gap between the synthetic and real scenarios by building a novel online redegraded learning strategy to successfully realize the high-quality deraining performance in unknown real-world scenarios. 
In contrast to existing approaches, our method addresses the disparity between synthetic and real-world scenarios through building a novel online re-degraded learning strategy. This approach allows the model to adapt to real-world conditions, thus bridging the gap between synthetic and real-world scenes.

\begin{figure*}[!t]
	\centering
	\begin{tabular}{c}
		\includegraphics[width=0.968\linewidth]{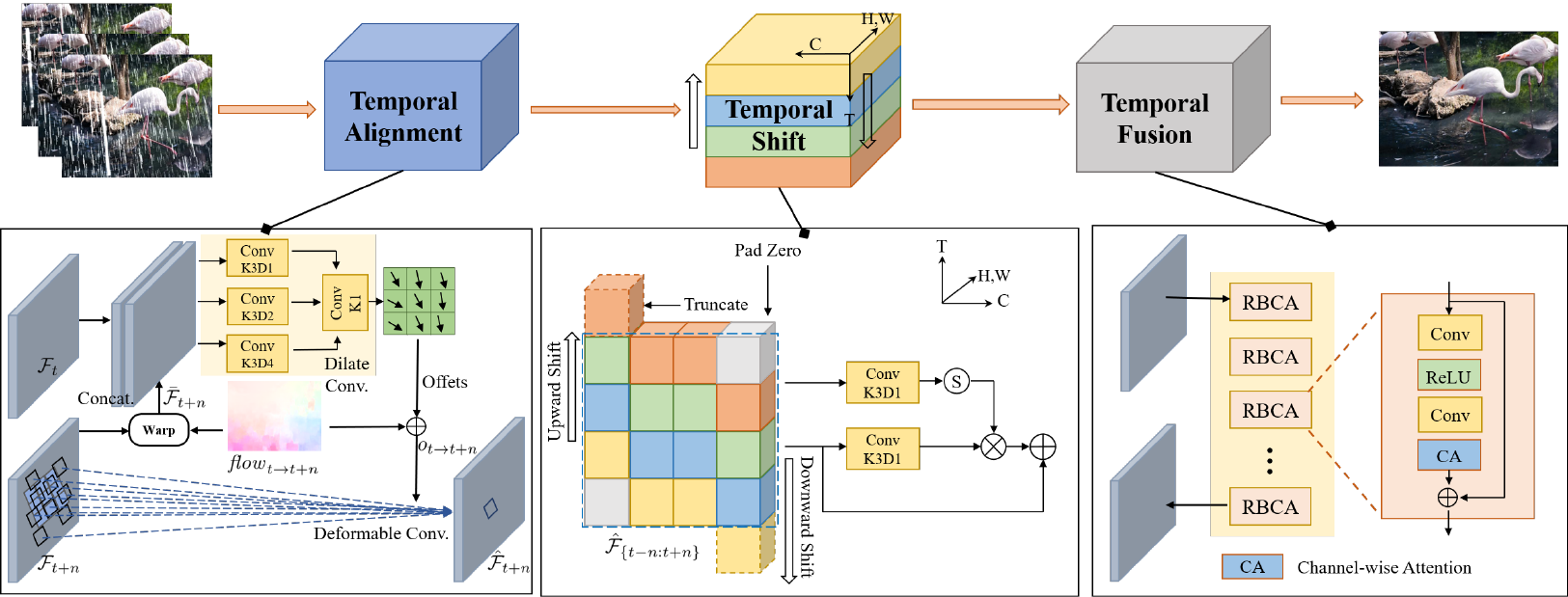}\\
	\end{tabular}
	\caption{The overview of \textbf{ASF-Net}. \textbf{ASF-Net} contains Temporal Alignment, Temporal Shift, and Temporal Fusion modules. Temporal Alignment performs the motion estimation-guided multi-sensory field feature alignment, combining optical flow with deformable convolution while expanding the receptive field with dilated convolution. Temporal shift is introduced in video deraining for the first time, where the temporal information of adjacent frames is deeply exploited by movement channels at the feature level, and the role of shift is enhanced by feature adaptive identification of rain streaks regions. The temporal fusion part contains some residual blocks with attention mechanism, which are fused after the accurate alignment and the full utilization of temporal information.}
	\label{fignet}
	\vspace{-0.5cm}
\end{figure*}

{\bf{Creating the rain dataset}}. 
%In the process of training and testing, the selection of datasets plays a significant part in the degree of model fit, the improvement of algorithm accuracy, the enhancement of robustness, and the improvement of network generalization ability. 
%During the process of both training and testing, the selection of datasets plays a critical role in facilitating model fit, enhancing algorithm accuracy, bolstering algorithmic robustness, and ultimately improving network generalization ability. 
%The current commonly used rain datasets include both synthetic rain datasets and real-world rain datasets. Synthetic rain datasets are formed by overlaying non-rain sequences and rendered rain streaks, with ground truth as the label. 
At present, widely used rain datasets contain both synthetic and real-world data. %Synthetic samples are obtained by overlaying non-rain sequences with rendered rain streaks, and the real-world ones are acquired from website or outdoor shooting. 
With these two kinds of data, robustness of the algorithm can be improved and the performance of rain removal can be better in real life.

To better analyze the characteristics of rain and manually create a rain streak dataset, Garg \emph{et al.}~\cite{garg2006photorealistic} proposed a new rain streak model that captures the complex interactions between the lighting direction, viewing direction, and the oscillating shape of the droplets. 
%The rain rendering algorithm obtained by this model can add rain streaks to images or videos of different scenes, and also lays the foundation of the method for the construction of synthetic rain datasets. 
%The rain rendering algorithm can add rain streaks to images or videos in various scenes. Additionally, it serves as the basis for creating synthetic rain datasets. 
On this basis, \emph{RainSynLight25, RainSynComplex25}, and \emph{RainPractical10} were constructed by Liu \emph{et al.}~\cite{liu2018erase}.  
\emph{RainSynLight25} and \emph{RainSynComplex25} are a light rain dataset and a heavy rain one respectively, synthesized by non-rain sequences with the rain streaks.
% generated by the probabilistic model~\cite{garg2006photorealistic}. 
\emph{RainPractical10} is a real-rain dataset obtained by website cropping. 
%In addition to the rain rendering model, the rain synthesis model can also be used for constructing the dataset. 
Based on a synthetic rain model with four degradation factors~\cite{yang2019frame}, a synthetic rain dataset \emph{RainSynAll100} with 1000 non-rain sequences was constructed. 
%Wang~\cite{wang2022rethinking} developed a video rain synthesis model that incorporates the concept of rain streak movements. This approach ensures consistency in the rain layers across video frames, resulting in more realistic rainy video. 
In addition, Chen \emph{et al.}~\cite{chen2018robust} proposed the \emph{NTURain} dataset, using a high-speed moving camera and a steady stationary camera to shoot real rain videos respectively.%, and adjusting the parameters of raindrop size, scene depth and wind direction through Adobe After Effects to build a more realistic synthetic rain by editing. 

%In general, high-quality datasets have three characteristics, first, large scale. Experiments on large-scale datasets yield more accurate conclusions. Second, abundant diversity, to include more scenarios and variations. Third, more realistic, not entirely synthetic data in laboratory scenarios. 
%High-quality datasets usually have three key characteristics. Firstly, they are large scale, meaning that experiments conducted on such datasets tend to produce more accurate conclusions. Secondly, they exhibit rich diversity by including a wide range of scenarios and variations. Finally, high-quality datasets are inherently more realistic than fully synthetic data generated in a laboratory setting.
%On the synthetic rain dataset, \emph{RainSynLight25, RainSynComplex25} and \emph{RainSynAll100} have achieved great results, but their synthetic rain types are relatively simple and still have huge difference compared with real-world rain streaks, and they lose some continuous temporal information of the real rainfall process, which has great constraints. 
%The synthetic rain data produced by \emph{RainSynLight25, RainSynComplex25} and \emph{RainSynAll100}, is relatively simple and differs significantly from the real-world rain streaks. Additionally, it fails to capture the continuous temporal information of actual rainfall processes, which poses significant limitations. 
In the real rain dataset, it still faces the problems of small scale, single scene and lack of rain diversity. 
%Although the \emph{NTURain} dataset makes use of real scenes, the training and testing datasets are still small in number and cover only light rain intensity. 
%Despite the utilization of real motion scenes in the \emph{NTURain} dataset, both the training and testing datasets remain severely constrained in terms of size and scope, only encompassing light rain intensity.  
These scenarios are also specifically summarized in~\cref{tabdataset}. 
Therefore, we constructed a dataset named LARA, containing abundant rain types of different intensities with temporal continuity, with greater scale and diversity of scenarios on both synthetic and real rain datasets, in accord with the characteristics of a high-quality dataset. 
%Keeping in view the characteristics of a high-standard dataset, we have formulated LARA dataset, which comprises of diverse rain types with varying intensities and distribution patterns, spanning both synthetic and real rain data, while maintaining the temporal consistency of rain sequences. Consequently, it encompasses a broader spectrum of scenarios, with an unprecedented scale and variability, elevating it to the rank of a high-quality dataset. 

\section{Temporal Alignment-Shift-Fusion Net}
\label{sec:ASF-Net}
To make full use of the temporal information between adjacent frames and effectively remove rain streaks from videos, we propose a new paradigm to exploit temporal cues for video deraining. As shown in~\cref{fignet}, ASF-Net consists of temporal alignment, temporal shift, and temporal fusion modules. In the following, we will detail these modules.

\begin{table}[!t]
	\centering
	\caption{Property analysis among different datasets.}
	\renewcommand\arraystretch{1.4}
	\begin{tabular}{|c|c@{\extracolsep{0.2em}}c@{\extracolsep{0.2em}}c@{\extracolsep{0.2em}}c|}
		\hline
		\footnotesize Datasets  &\footnotesize \emph{NTURain} &\footnotesize \emph{Light25} &\footnotesize \emph{Complex25} &\footnotesize LARA \\
		\hline
		\footnotesize {{Intensity Level}}  &\footnotesize Light&\footnotesize Medium &\footnotesize Heavy  &\footnotesize {Both} \\
		\hline
		\footnotesize {Train/Test/Real} &\footnotesize 25/8/7 & \footnotesize 190/25/0  &\footnotesize 190/25/0  &\footnotesize 1000/37/120  \\
		\hline 
		\footnotesize {Rain Type} &\footnotesize  {Streak} &\footnotesize {Streak}  &\footnotesize{Line}&\footnotesize  {Occlusion}  \\ 
		\hline
		\footnotesize {Direction}  &\footnotesize  {Similar}&\footnotesize {Similar}  &\footnotesize{Similar} &\footnotesize  {Diverse}  \\ 
		\hline
	\end{tabular}
	\label{tabdataset}
	\vspace{-0.3cm}	
\end{table}

\subsection{Motion Estimation-guided Multi-sensory Field Feature Alignment}
For more accurate temporal alignment, we propose a motion estimation-guided multi-sensory field feature alignment module. 
%Considering the artifacts caused by using only optical flow, on the basis of ~\cite{wang2019edvr,tian2020tdan}, we first use deformable~\cite{dai2017deformable} convolution to align adjacent frames on the feature level. 
Considering the artifacts that can arise from the use of optical flow, we seek to further enhance the results achieved by recent works, such as~\cite{wang2019edvr,tian2020tdan}, applying deformable convolution%~\cite{dai2017deformable}
 to align adjacent frames at the feature level. 
The detail is shown in~\cref{fignet}. To avoid the instability of deformable convolution%~\cite{chan2021understanding}
, we incorporate optical flow into alignment module%~\cite{chan2021basicvsr++}
, in which optical flow can provide explicit guidance. We utilize SPyNet~\cite{ranjan2017optical} initialized by pretrained weights to obtain the  optical flow. So we only need to estimate the residual offset, which can reduce the burden in typical deformable convolution. 

Specifically, we warp the adjacent feature $\mathbf{\mathcal{F}}_{t+n}$ with optical flow $flow_{t \to {t+n}}$ pre-estimated by SPyNet. Different from~\cite{chan2021basicvsr++}, we design a multi-scale dilated block to estimate residual offsets based on warped feature $\bar{\mathbf{\mathcal{F}}}_{t+n}$ and $\mathbf{\mathcal{F}}_{t}$. Since the dilated convolution can enlarge the receptive field to sense large motions between adjacent frames. The final offsets $o_{t \to {t+n}}$ are obtained by adding the residual offsets to the optical flow. In this way, the obtained offsets can stay stable during training without overflow. 
%At the same time, the expanded receptive field of the dilated convolution makes it possible to achieve better alignment results for backgrounds moving at high or low speeds, and to be more sensitive to motion, further preventing the appearance of artifacts. 
Finally, the aligned feature $\hat{ \mathbf{\mathcal{F}}}_{t+n}$ is obtained by applying deformable convolution to the adjacent feature $\mathbf{\mathcal{F}}_{t+n}$. 

%This temporal alignment module will be more sensitive to motion. Whether the scene is moving at high speed or vulgar movement, better alignment can be achieved through multi-sensory fields and make full use of the temporal information between adjacent frames. 
The proposed temporal alignment module can capture the temporal correlations among adjacent frames more sensitively. This module allows for better alignment using multi-sensory official fields, regardless of the degree of motion occurring in the real scene.

\subsection{Feature-Adaptive Temporal Shift}
\label{ssec:shift}
Traditional video deraining methods %~\cite{xue2021temporal,yang2019frame,yang2020self} 
(``alignment + fusion'') directly fuse the aligned results with the fusion network, which cannot efficiently extract redundant temporal information from adjacent frames. To address these issues, we propose a feature-adaptive temporal shift module, which can efficiently exploit sequence-level information by exchanging information among adjacent frames on the feature space. 

After acquiring the aligned features derived by the temporal alignment, we adopt temporal shift%~\cite{lin2019tsm}
 to borrow as much temporal information as possible from the  adjacent frames.  
%exploit temporal redundant information. 
As shown in~\cref{fignet}, we shift $[0:a]$ channels upwards and $[c-a:c]$ channels downwards. In this way, the feature of $t$-th frame also contains information from $t-n$ and $t+n$ frames. 
Mathematically, the shift process can be described as
\begin{equation}
\begin{array}{lcl}
\tilde{\mathcal{{F}}}_{t}[0:a]=\hat{\mathcal{{F}}}_{t-n}[0:a], \\
\tilde{\mathcal{{F}}}_{t}[c-a:c]=\hat{\mathcal{{F}}}_{t+n}[c-a:c],
\end{array}
\end{equation}
where $c$ denotes the number of feature channels, $a$ denotes the number of shifted feature channels. Empirically, we set $a=c/8$.

Moreover, considering the property of rain streaks, we further utilize the feature-adaptive mechanism and residual learning %~\cite{he2016deep}  
to enforce our shift module to identify rain streaks area from the shifted features. Specifically, we adopt one convolution layer with 3$\times$3 kernel size, and a Sigmoid layer to obtain adaptive weights. And the shifted feature is multiplied by adaptive weights using the element-wise product. 

The introduction of the feature adaptive temporal shift module builds a new paradigm for video decompression networks. The shift module utilizes temporal information to the feature channel level compared to coarse post-alignment fusion and further refines the temporal information based on motion estimation guided alignment.
%The exploitation of adjacent frames appears in both alignment and shift modules, and the unused temporal information from the alignment modules is extracted more deeply in the shift phase. 
The proposed paradigm involves the exploitation of adjacent frames in both alignment and shift modules. The alignment module captures temporal correlations among adjacent frames, while the shift module delves deeper into the temporal information by extracting unused information from the alignment stage. Through this integrated use of temporal information, our approach aims to improve the accuracy and effectiveness of learning-based video decoding methods.
To demonstrate the superiority of the temporal shift module, we further argue this point through experiments, as shown in~\cref{tab:net_study}. In contrast to only using the alignment module, the application of feature-adaptive temporal shift makes a significant improvement in the evaluation metrics. The quantitative evaluation of this temporal shift module is detailed in section~\ref{ssec:net_study}.

\subsection{Temporal Fusion}

To obtain a rain-free video, we adopt several cascaded residual blocks as the basic component of the temporal fusion module to fuse shifted features $\tilde{\mathcal{{F}}}_{\{t-n:t+n\}}$. In this work, we empirically cascade 8 residual blocks for the best fusion performance. And each residual block consists of two convolution layers and a channel-wise attention.%~\cite{hu2018squeeze}. 
The detailed structure is shown in the right side of~\cref{fignet}.
% The local skip connection of residual block is beneficial for propagating gradients and accelerating convergence. 
In addition, less useful information can be bypassed through multiple skip connections. Moreover, on the basis of the temporal shift module for rain area identification, channel-wise attention can assign more weight to important feature maps, which can improve the deraining performance~\cite{li2018recurrent}. 

\section{Synthetic-to-Real Learning}
\label{sec:mda}

\begin{figure}[!t]
	\centering
	\begin{tabular}{c}
		\includegraphics[width=0.95\linewidth]{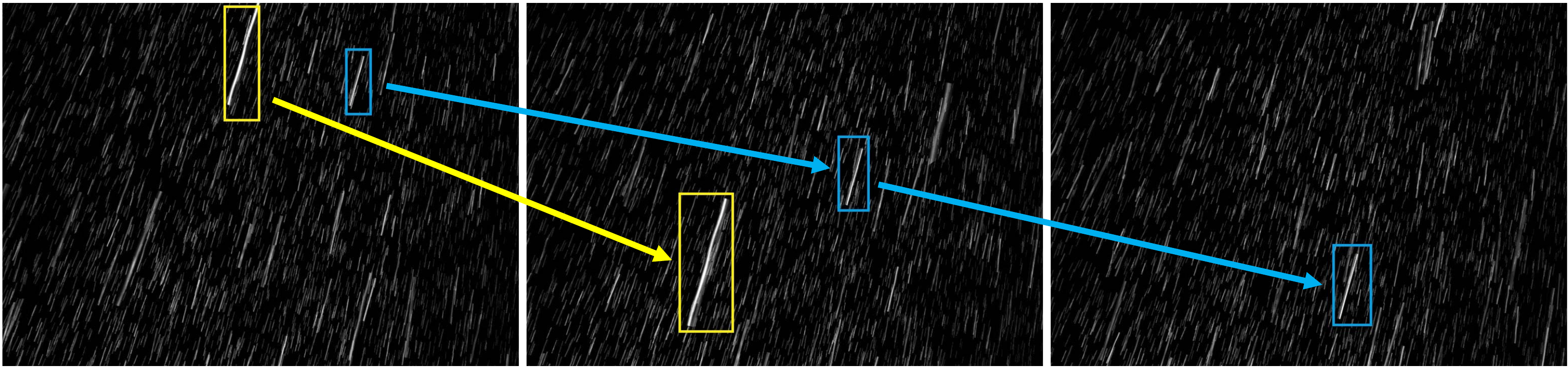}\\
		\footnotesize (a) The consecutive rain streaks in the video sequence \\
		\label{fig:Continuous}
		\includegraphics[width=0.95\linewidth]{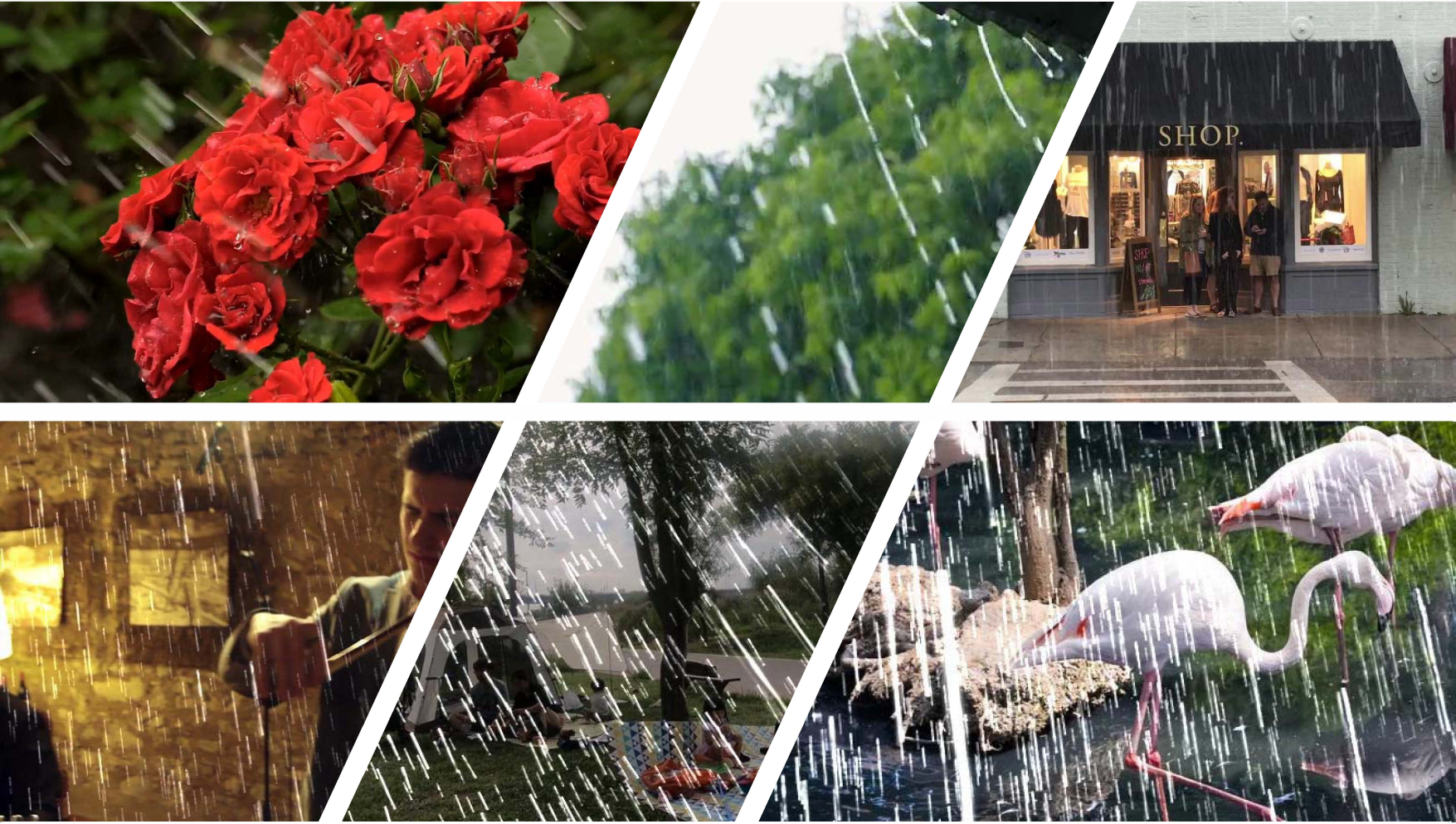}\\
		\footnotesize (b) Examples of rainy inputs (top: real-world, bottom: synthetic)\\		
	\end{tabular}
	\caption{Examples of LARA dataset. (a) shows the continuity of rain streaks in adjacent frames. (b) shows the real-world and synthetic rainy images. }
	\label{fig:Sample}
	\vspace{-0.5cm}
\end{figure}

\subsection{Large-Scale Rainy Video Dataset}
To help remove the rain from videos, we propose a \textbf{LA}rge-scale \textbf{RA}iny video dataset (\textbf{LARA}). It concludes large-scale synthetic data with temporal continuous rain streaks and plentiful real-world rain videos.

\textbf{Synthetic dataset with continuous rain streaks.}
There are some limitations in the common benchmarks~\cite{liu2018erase, chen2018robust} for video deraining. The type and direction of synthetic streaks lack diversity and the synthetic streaks are not realistic enough. And the worst, these rain streaks are not temporal continuous. To solve aforementioned problems, we propose a large-scale diverse video deraining dataset. Since the rain streak is not related to a specific scene, we mainly focus on the synthesis of the rain streak to ease the synthesis process.

We utilize a rain particle system\footnote{https://developer.download.nvidia.cn/SDK/10/direct3d/samples.html} based on Direct3D10 for animating and rendering continuous rain streaks over time. The rendering of the rain particles uses some rain textures which were created by Garg\footnote{https://www1.cs.columbia.edu/CAVE/databases/rain\_streak\_db/}~\cite{garg2006photorealistic}. To generate diverse rain streaks, we adjust various properties of the rain particle system, such as the direction, rain response, wind intensity, particle drawn numbers, distance, etc.

The ground-truth video sequences are simply sampled from REDS~\cite{nah2019ntire}, DAVIS~\cite{pont20172017} and Vimeo90K~\cite{xue2019video} which are proposed for video super-resolution and object segmentation. Based on the linear superposition model of rain video~\cite{yang2020single}, we sample one rain streak video clip randomly and add it on a clean video clip to create synthetic dataset.

\begin{figure}[!t]
	\centering
	\begin{tabular}{c@{\extracolsep{0.3em}}c}
		\includegraphics[width=0.499\linewidth]{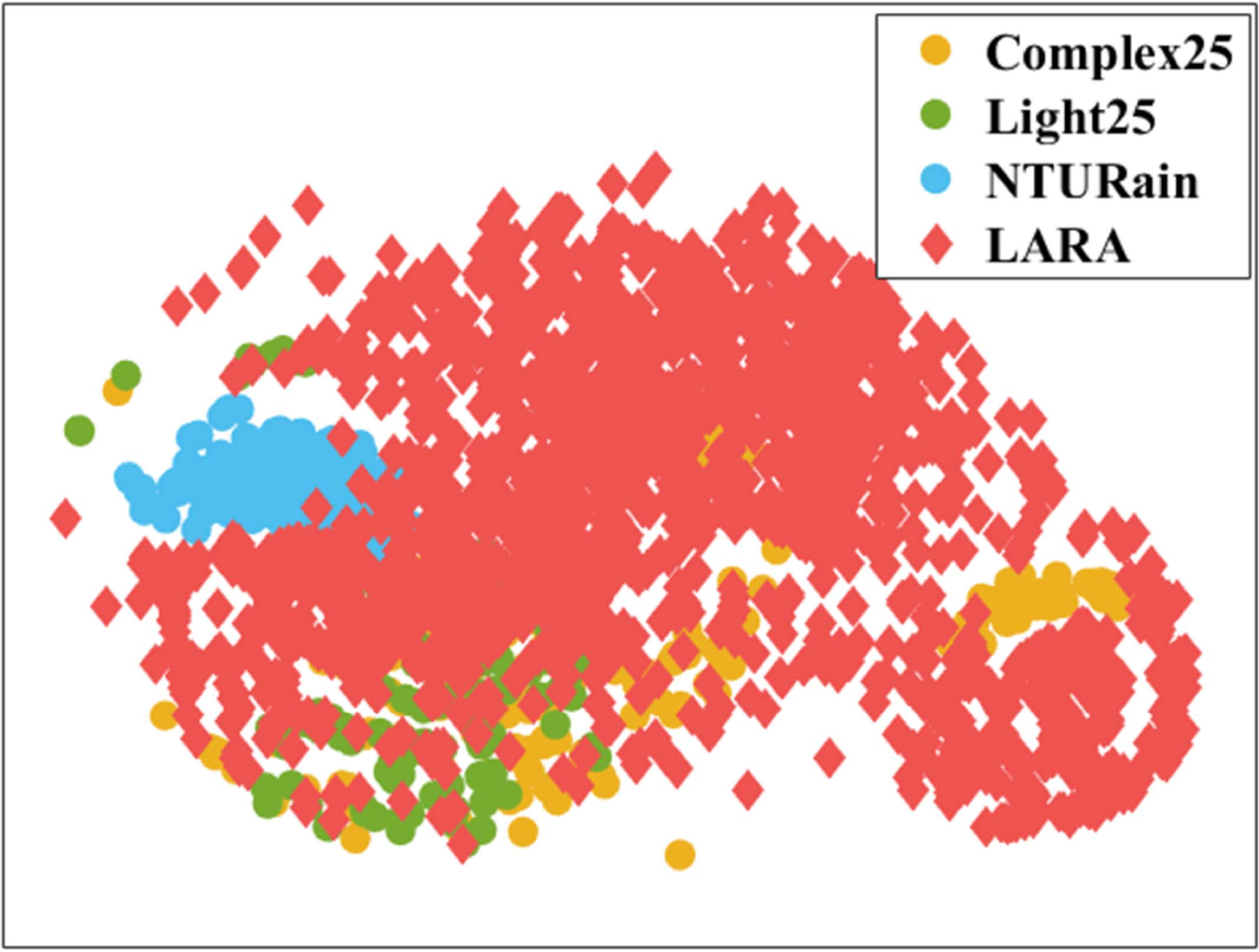}&
		\includegraphics[width=0.456\linewidth]{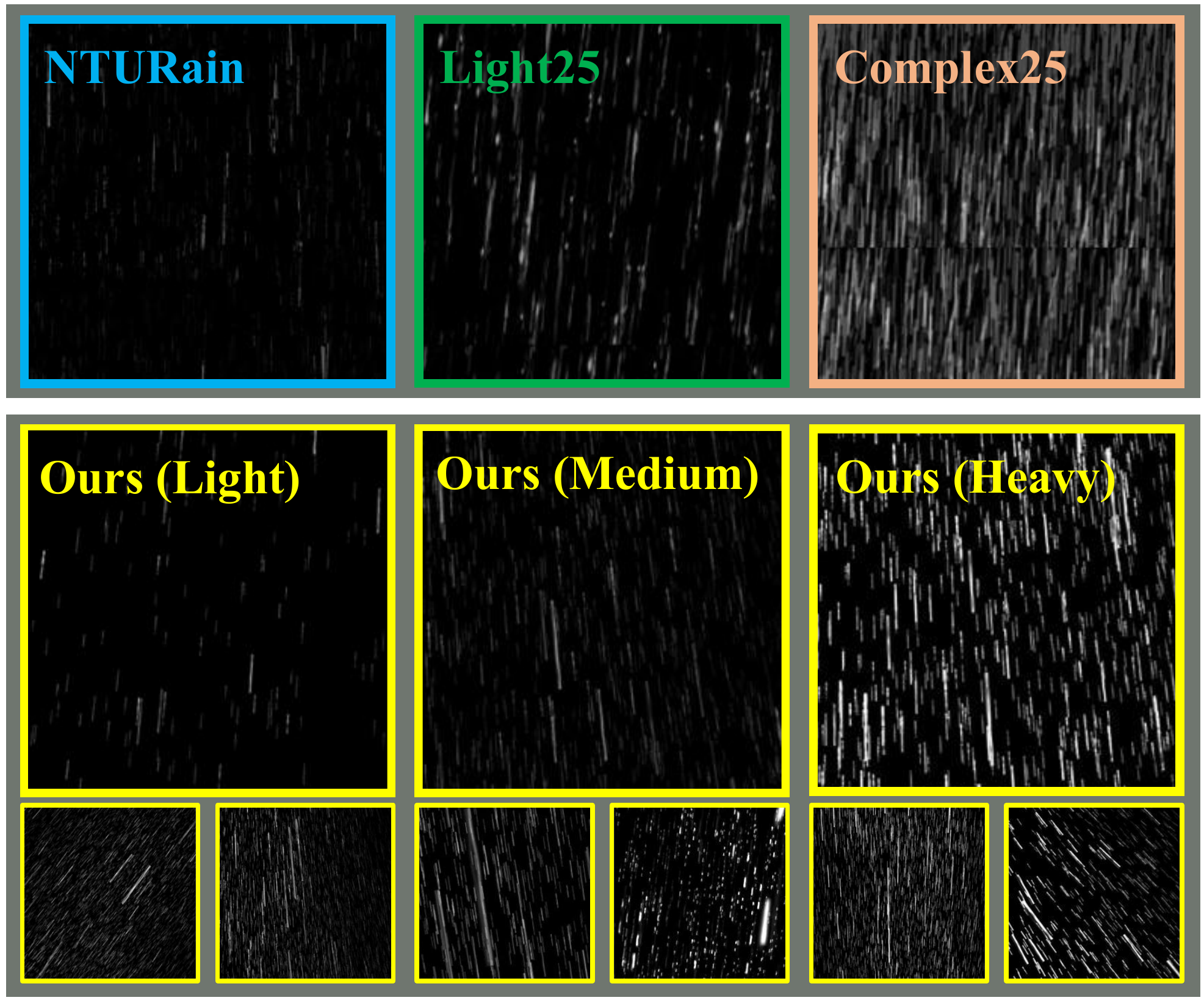}\\
		\footnotesize (a) t-SNE distribution&\footnotesize (b) Comparing rain streaks\\
	\end{tabular}
	\caption{Comparing different datasets from the statistical (a) and visual (b) aspects of rain streaks.}
	\label{fig:tSNE}
	\vspace{-0.5cm}
\end{figure}

\begin{figure}[t]
	\centering
	\begin{tabular}{c}
		\includegraphics[width=0.95\linewidth]{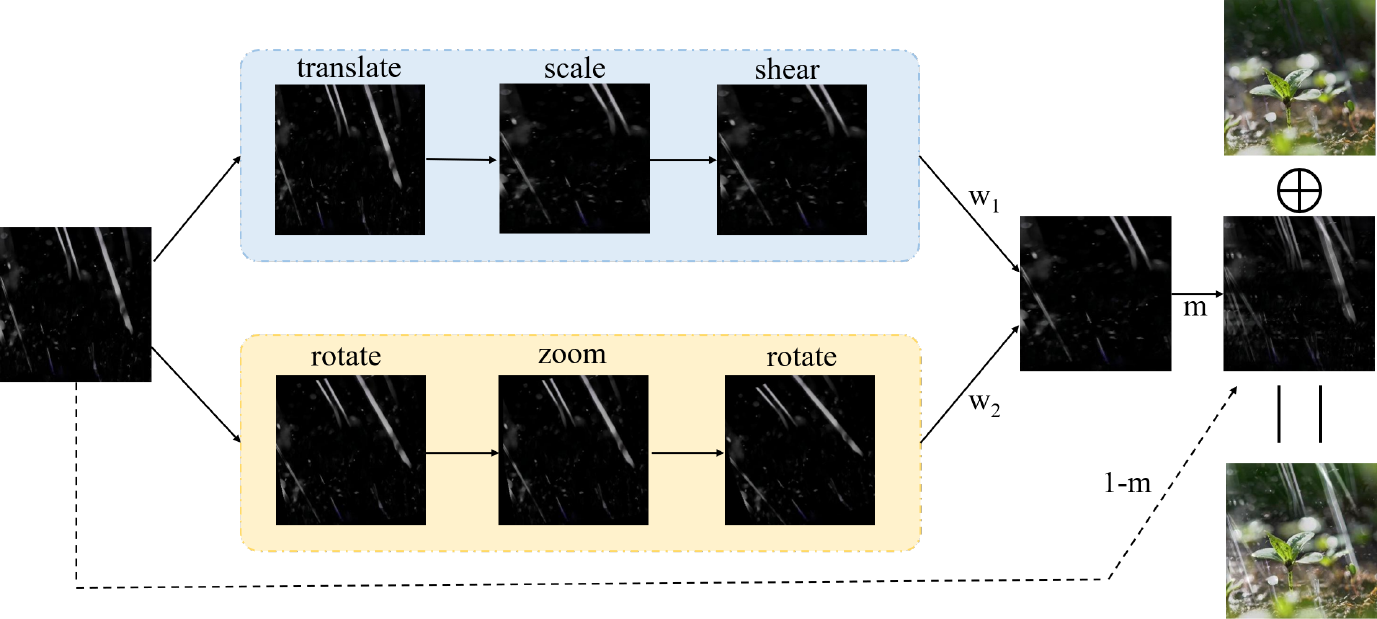} \\	
	\end{tabular}
	\caption{The computational flow of ReDe. ReDe transforms the estimated \textbf{real} streaks with random operations, and further mixes these transformed results with stochastic composition coefficients. %The square frame shows the enumeration of random operation choices.
}
	%\caption{The computational flow of ReDe. }
	\label{fig:rede}
	\vspace{-0.3cm}
\end{figure}

\noindent\textbf{Collecting real dataset from the web.}
In addition to these synthetic data, we also collect some real-world rain videos from online websites or movie clips, such as Youtube\footnote{https://www.youtube.com/}, Pexels\footnote{https://www.pexels.com/}, Pixabay\footnote{https://pixabay.com/}, Videezy\footnote{https://www.videezy.com/} and Mixkit\footnote{https://mixkit.co/}. These collected videos contain various rain streaks, such as different intensities, different directions, different shapes, etc. The scene motion is also different, some videos are taken from fast-moving cameras and other videos have many objects moving at different speeds. Some examples of our synthetic dataset and real dataset are shown in Fig.\ref{fig:Sample} (b).

\noindent\textbf{Comparison with existing datasets.}
In this section, we compare LARA  with other benchmarks in some aspects.~\cref{tabdataset} summarizes the characteristics of different datasets. Generally speaking, the proposed LARA dataset contains different intensities of rain streaks, a large-scale training set, plentiful real rainy data, and more diverse rain streak patterns. More importantly, as shown in~\cref{fig:Sample} (a), our synthesized rain streaks are temporally consistent, which is more suitable for video deraining task. 

As the t-SNE results shown in~\cref{fig:tSNE}. The separation of rain streaks from existing datasets~\cite{liu2018erase,chen2018robust} demonstrates the huge gap among them, which means the deraining models are hard to transfer from one dataset to another. On the contrary, our LARA dataset contains more rain streak patterns and covers almost every rain streak types in these existing datasets. The model trained on our dataset can be transferred to existing datasets more easily. More details are shown in the analysis  section~\ref{ssec:dataset_study}.

\subsection{Online Re-degraded Learning Strategy}
Since there is a huge domain gap between the synthetic and real-world scenes, the model trained purely on synthesized data is hard to obtain satisfactory deraining performance in various real scenarios. To improve the robustness of our ASF-Net on real-world scenes, we propose an Online Re-degraded Learning strategy (ORL) to effectively extract knowledge from real data. The algorithm detail is elaborated in~\cref{alg:algo}.

We first use synthesized data to train ASF-Net, $f_\theta$, the optimization objective is MAE loss, defined as 
\begin{equation}
\label{eq:su}
\mathcal{L}_{SU} = \sum_{t=1}^{T} \|\hat{\mathbf{B}}^t_L - \mathbf{B}^{t}_L\|_1, 
\end{equation}
here $\hat{\mathbf{B}}^t_L$ denotes the estimation of the t-th frame,  $\mathbf{B}^{t}_L$ denotes the ground truth of the t-th frame of a video clip.

\begin{algorithm}[t]
	\caption{ORL: Online Re-degraded Learning}
	\label{alg:algo}
	\KwIn{Labeled synthetic data $\{ \mathbf{O}_{L}, \mathbf{B}_{L} \} \in \Theta_{syn}$, unlabeled real data:  $\mathbf{O}_{U} \in \Theta_{real}$, synthetic rain streak database $\mathbf{S}_{L} \in \Omega_{s}$, model $f_\theta$ pretrained on $\Theta_{syn}$, re-degraded method $\textbf{ReDe}()$.} %Mixing operation set $\mathcal{O} = \{ rotate, zoom, translate, shear \}$.}
	\KwOut{Final deraining model $f_\theta$} 
	\While{$\text{not converged}$}{
		$\mathbf{O}_{U}$ $\gets$ sample from $\Theta_{real}$ \;
		$\mathbf{B}_{U} \gets f_\theta(\mathbf{O}_{U})$ \;
		$\mathbf{S}_{U} = \mathbf{O}_{U} - \mathbf{B}_{U}$ \;
		
		$\mathbf{S}_{L} \gets $  sample from $\Omega_{s}$ \;
		$\mathbf{O}_{P} = \mathbf{B}_{U} + \textbf{ReDe}(\mathbf{S}_{U}, \mathbf{S}_{L})$ \;		
		%	}

		$\hat{\mathbf{B}}_P \gets f_\theta(\mathbf{O}_{P})$ \;
		%		$(\mathbf{O}_{P}, \mathbf{B}_{N}) \gets $ composite a pseudo labeled pair \;
		$\{ \mathbf{O}_{L}, \mathbf{B}_{L} \} \gets$ sample from $\Theta_{syn}$ \;
		
		$\hat{\mathbf{B}}_L \gets f_\theta(\mathbf{O}_{L})$ \;
		
		$\mathcal{L} = \mathcal{L}_{SU}(\hat{\mathbf{B}}_L, \mathbf{B}_{L}) + \lambda_{UN}\mathcal{L}_{UN}(\hat{\mathbf{B}}_P, \mathbf{B}_{U})$ \;
		$\theta \gets \theta - \eta\nabla_\theta\mathbf{\mathit{L}}$ \;
		%		\tcp{Update parameters of model $f_\theta$}\; 
	}
	
	\Return{$f_\theta$} \;
\end{algorithm}

\begin{table*}[t]
	\centering
	\caption{Quantitative results among different video deraining methods on our constructed {LARA} dataset.}
     \renewcommand\arraystretch{1.3}
     \resizebox{\textwidth}{!}{
	\begin{tabular}{|c@{\extracolsep{0.6em}}|c@{\extracolsep{0.6em}} c@{\extracolsep{0.6em}} c@{\extracolsep{0.6em}} c@{\extracolsep{0.6em}} c@{\extracolsep{0.6em}} c@{\extracolsep{0.6em}} c@{\extracolsep{0.6em}} c@{\extracolsep{0.6em}} c@{\extracolsep{0.6em}} c@{\extracolsep{0.6em}} c@{\extracolsep{0.6em}} c@{\extracolsep{0.6em}} c|@{\extracolsep{0.6em}} c|}
		\hline 
		\footnotesize Metrics&\footnotesize SIRR &\footnotesize Syn2Real &\footnotesize MOSS &\footnotesize SE&\footnotesize DIP&\footnotesize MSCSC&\footnotesize FastDerain &\footnotesize J4RNet &\footnotesize SpacCNN&\footnotesize SLDNet&\footnotesize S2VD &\footnotesize ESTI &\footnotesize RDD &\footnotesize Ours\\
		\hline
		\footnotesize PSNR $\uparrow$ &\footnotesize 20.810 &\footnotesize 20.492 &\footnotesize 20.664  &\footnotesize 19.140 &\footnotesize 25.192 &\footnotesize 19.178 &\footnotesize 21.625 &\footnotesize 26.143 &\footnotesize 27.121  &\footnotesize 25.698 &\footnotesize 29.274 &\footnotesize 25.756 &\footnotesize 25.511 &\footnotesize \textbf{33.332}\\
		\hline
		\footnotesize SSIM $\uparrow$ &\footnotesize 0.692 &\footnotesize 0.674 &\footnotesize 0.701 &\footnotesize 0.654 &\footnotesize 0.797 &\footnotesize 0.563 &\footnotesize 0.695 &\footnotesize 0.835 &\footnotesize 0.821  &\footnotesize 0.761 &\footnotesize  0.887 &\footnotesize 0.803 &\footnotesize 0.812 &\footnotesize \textbf{0.945}\\
		\hline
		\footnotesize NIQE $\downarrow$ &\footnotesize 4.337  &\footnotesize 5.026 &\footnotesize 4.436  &\footnotesize 5.462 &\footnotesize 3.087 &\footnotesize 4.653 &\footnotesize 4.813 &\footnotesize 2.789 &\footnotesize 2.777  &\footnotesize 3.823 &\footnotesize 2.901 &\footnotesize 3.786 &\footnotesize 3.254 &\footnotesize \textbf{2.724}\\				
		\hline
		\footnotesize LPIPS $\downarrow$ &\footnotesize 0.331  &\footnotesize 0.357 &\footnotesize 0.326  &\footnotesize 0.379 &\footnotesize 0.226 &\footnotesize 0.404 &\footnotesize 0.335 &\footnotesize 0.186 &\footnotesize 0.178  &\footnotesize 0.233 &\footnotesize 0.110 &\footnotesize 0.243 &\footnotesize 0.206 &\footnotesize \textbf{0.047}\\
		\hline
		\footnotesize tLP $\downarrow$ &\footnotesize 0.131  &\footnotesize 0.129 &\footnotesize 0.129  &\footnotesize 0.119 &\footnotesize 0.066 &\footnotesize 0.146 &\footnotesize 0.101 &\footnotesize 0.077 &\footnotesize 0.057  &\footnotesize 0.037 &\footnotesize 0.015 &\footnotesize 0.046 &\footnotesize 0.075 &\footnotesize \textbf{0.011}\\		
		\hline
	\end{tabular}
}
	\label{tab:value_our}
\end{table*}

\begin{table*}[t]
	\centering
	\caption{Quantitative results among different deraining methods on different existing benchmarks.}
     \renewcommand\arraystretch{1.3}
     \resizebox{\textwidth}{!}{
	\begin{tabular}{|c|c|c@{\extracolsep{0.5em}}c@{\extracolsep{0.8em}}c@{\extracolsep{0.5em}}c@{\extracolsep{0.5em}}c@{\extracolsep{0.8em}}c@{\extracolsep{0.8em}}c@{\extracolsep{0.8em}}c@{\extracolsep{0.8em}}c@{\extracolsep{0.8em}}c@{\extracolsep{0.8em}}c|@{\extracolsep{0.8em}}c|}
		\hline 
		\footnotesize Datasets&\footnotesize Metrics  &\footnotesize SIRR&\footnotesize Syn2Real&\footnotesize MOSS&\footnotesize SE&\footnotesize DIP&\footnotesize MSCSC&\footnotesize FastDerain  &\footnotesize J4RNet &\footnotesize SpacCNN &\footnotesize SLDNet&\footnotesize S2VD &\footnotesize Ours\\
		\hline
		\multirow{2}{*}{\footnotesize {Light25} }&\footnotesize PSNR &\footnotesize 24.590 &\footnotesize 25.465 &\footnotesize 26.774 &\footnotesize 25.443 &\footnotesize 28.070 &\footnotesize 24.487 &\footnotesize 31.571 &\footnotesize 30.542 &\footnotesize 31.577   &\footnotesize 28.734 &\footnotesize 33.843 &\footnotesize \textbf{38.099}\\
		\cline{2-14}
		~&\footnotesize SSIM&\footnotesize 0.823 &\footnotesize 0.866 &\footnotesize 0.868 &\footnotesize 0.750 &\footnotesize 0.853 &\footnotesize 0.734 &\footnotesize 0.905 &\footnotesize 0.906 &\footnotesize 0.899  &\footnotesize 0.843 &\footnotesize 0.925 &\footnotesize \textbf{0.969}\\
		\cline{2-14}
		%			~&\footnotesize LPIPS &\footnotesize \textcolor{blue}{\textbf{-}} &\footnotesize - &\footnotesize - &\footnotesize - &\footnotesize - &\footnotesize - &\footnotesize - &\footnotesize - &\footnotesize - &\footnotesize - &\footnotesize - &\footnotesize \textcolor{red}{\textbf{-}}\\
		
		\hline
		\multirow{2}{*}{\footnotesize {Complex25} }&\footnotesize PSNR &\footnotesize 18.185 &\footnotesize 18.374 &\footnotesize 18.596 &\footnotesize 18.085 &\footnotesize 18.834 &\footnotesize 16.604 &\footnotesize 22.464 &\footnotesize 23.613 &\footnotesize 21.582  &\footnotesize 18.804 &\footnotesize 27.772 &\footnotesize \textbf{30.805}\\
		\cline{2-14}  
		~&\footnotesize SSIM&\footnotesize 0.555 &\footnotesize 0.559 &\footnotesize 0.562 &\footnotesize 0.590 &\footnotesize 0.532 &\footnotesize 0.486 &\footnotesize 0.734 &\footnotesize 0.749 &\footnotesize 0.596  &\footnotesize 0.529 &\footnotesize 0.818 &\footnotesize \textbf{0.904}\\
		\cline{2-14}
		\hline
		\multirow{2}{*}{\footnotesize {NTURain} }&\footnotesize PSNR&\footnotesize 23.683 &\footnotesize 26.302 &\footnotesize 30.292 &\footnotesize 25.423 &\footnotesize 30.752 &\footnotesize 25.634 &\footnotesize 31.743 &\footnotesize 32.976 &\footnotesize 33.348  &\footnotesize 34.265 &\footnotesize 37.372 &\footnotesize \textbf{39.706}\\
		\cline{2-14}  
		~&\footnotesize SSIM&\footnotesize 0.836 &\footnotesize 0.927 &\footnotesize 0.936 &\footnotesize 0.755 &\footnotesize 0.897 &\footnotesize 0.760 &\footnotesize 0.928 &\footnotesize 0.952 &\footnotesize 0.955  &\footnotesize 0.956 &\footnotesize 0.968 &\footnotesize \textbf{0.981}\\
		\cline{2-14}
		\hline
	\end{tabular}}
	\label{tab:value_LH}
	\vspace{-0.4cm}
\end{table*}
As illustrated in~\cref{alg:algo}, we utilize the pre-trained network $f_\theta$ to separate the real rain streak $\mathbf{S}_{U}$ and background $\mathbf{B}_{U}$, which may remain some rain streaks. 
To adequately exploit these real rain streaks, we design a Re-Degrade method (ReDe) to re-degrade $\mathbf{B}_{U}$ to generate new pseudo-paired data. 
Specifically, we use ReDe to transform the estimated real streaks $\mathbf{S}_{U}$ and synthetic streaks $\mathbf{S}_{L}$ sampled from~\cite{garg2006photorealistic} to generate more diverse rain streaks. Then the diverse rain streaks are added to $\mathbf{B}_{U}$ to construct the re-degraded frame $\mathbf{O}_{P}$. Then $\mathbf{O}_{P}$ and $\mathbf{B}_{U}$ can form a new pseudo-paired data. 

As shown in~\cref{fig:rede}, ReDe performs two transformation chains on one streak layer. The transformation chain is constructed by composing several randomly selected operations, such as rotation, zoom, translation, and shear transformation. 
%Based on~\cite{hendrycks2019augmix}, the three transformed streak layers are mixed via the weights sampled from Dirichlet distribution and are further blended with the original streak layer through a second weight sampled from Beta distribution. 
After multiple operations, the transformed streak layers are mixed via randomly generated weights \textit{w} and are further blended with the original streak layer through a second weight \textit{m} sampled by randomization. 
Finally, $\mathbf{O}_{P}$ is obtained by adding the transformed streak layer back to $\mathbf{B}_{U}$. The multiple random processes in ReDe makes it possible to efficiently generate diverse rain streaks without empirical parameters.  
%The algorithm details of ReDe are provided in the supplementary material. 

As illustrated in~\cref{alg:algo}, we use this pseudo paired data $\{ \mathbf{O}_{P}, \mathbf{B}_{U} \}$ and our LARA synthetic data $\{ \mathbf{O}_{L}, \mathbf{B}_{L} \} $ to further fine-tune the network $f_\theta$. Similar to Eq.\ref{eq:su}, we adopt MAE loss for the pseudo labeled data, defined as
\begin{equation}
\mathcal{L}_{UN} = \sum_{t=1}^{T} \|\hat{\mathbf{B}}^t_P - \mathbf{B}^{t}_U\|_1,
\end{equation}
where $\hat{\mathbf{B}}^t_P$ denotes the estimation of the t-th real rainy frame. So the total objective for online re-degraded learning is 
\begin{equation}
\mathcal{L} = \mathcal{L}_{SU} + \lambda_{UN}\mathcal{L}_{UN}, 
\end{equation}
where $\lambda_{UN}$ is a parameter which is set to be 1.0 empirically. 

\section{Experimental Results}
\label{sec:experiment}

\subsection{Experiment Settings}

\noindent\textbf{Datasets}. 
We compare the proposed method with several state-of-the-art methods on our LARA dataset and three commonly-used benchmarks, \emph{Light25}, \emph{Complex25}, and \emph{NTURain}. %\emph{Light25}~\cite{liu2018erase} dataset contains light rain streaks and \emph{Complex25}~\cite{liu2018erase} dataset contains heavy rain streaks, sharp line streaks, and sparkle noises. \emph{NTURain}~\cite{chen2018robust} was conducted from an unstable camera with slow motion and a fast-moving camera.  

\noindent\textbf{Training Details}
In the training process, we employed ADAM optimizer and weight decay is set to $10^{-4}$ to train the whole network. Learning rate is set to $10^{-4}$. Besides, We randomly cropped the video pairs with the spatial size of 256*256 pixels and the temporal length of 5 to train the whole network in each iteration. The number of features extracted from each frame is set to $64$. The code is implemented under PyTorch library, and the model is trained on a NVIDIA TITAN Xp GPU and tested on a NVIDIA GeForce RTX 2060 GPU. 
For training stability, we firstly train ASF-Net on synthesized data using MAE loss, which takes about 20,000 iterations to get converged. It should be mentioned that we employ MAE loss instead of MSE loss, since using MSE loss may produce over-smoothed results as the squared penalty works poorly at the edges and usually results in large errors.%~\cite{fu2019lightweight}. 

\noindent\textbf{Baselines}. We compare the proposed method with many deraining approaches: SIRR~\cite{wei2019semi}, Syn2Real~\cite{yasarla2020syn2real}, MOSS~\cite{huang2021memory}, SE~\cite{wei2017should}, DIP~\cite{jiang2017novel}, MS-CSC~\cite{li2018video}, FastDerain~\cite{jiang2018fastderain}, SpacCNN~\cite{chen2018robust}, J4RNet~\cite{liu2018erase}, SLDNet~\cite{yang2020self}, S2VD~\cite{yue2021semi}, ESTI~\cite{zhang2022enhanced} and RDD~\cite{wang2022rethinking}. SIRR, PReNet, Syn2Real, MOSS, and MPRNet are representative single image deraining methods. SE, DIP, MS-CSC, FastDerain, SpacCNN, J4RNet, SLDNet, S2VD, ESTI and RDD are video deraining methods. SE, DIP, MSCSC, and FastDerain are traditional methods. SIRR, Syn2Real, MOSS, S2VD, SLDNet, ESTI and RDD are semi-supervised or unsupervised methods. 

\noindent\textbf{Evaluation Criterions}. 
We adopt PSNR and SSIM to evaluate the image fidelity quality.  %Since human visual system is more responsive to luminance  than chrominance information, 
We evaluate all the results only in the luminance channel following previous works~\cite{yang2019frame}. We also utilize the Learned Perceptual Image Patch Similarity (LPIPS) %~\cite{zhang2018unreasonable} 
and the no-reference Natural Image Quality Evaluator (NIQE) %~\cite{mittal2012making} 
to evaluate perceptual quality. 
Besides, we utilize the tLP~\cite{chu2020learning} metric to measure the temporal consistency by computing the LPIPS distance for consecutive restored frames and the related ground-truth frames. 

\begin{figure*}[t]
	\centering
	\begin{tabular}{c@{\extracolsep{0.1em}}c@{\extracolsep{0.1em}}c@{\extracolsep{0.1em}}c@{\extracolsep{0.1em}}c@{\extracolsep{0.1em}}c@{\extracolsep{0.1em}}c@{\extracolsep{0.1em}}c}
		\includegraphics[width=0.121\linewidth]{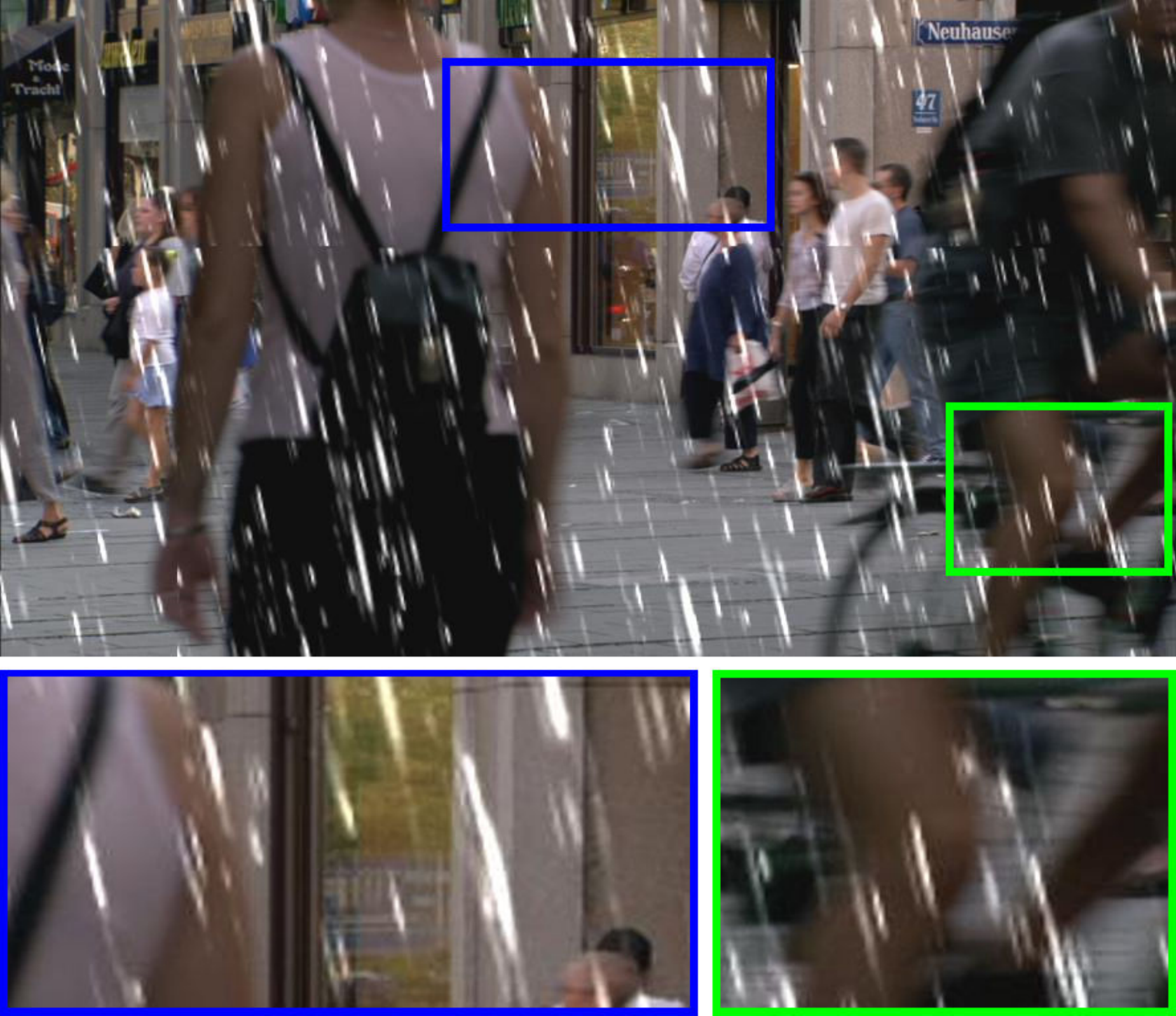}&
		\includegraphics[width=0.121\linewidth]{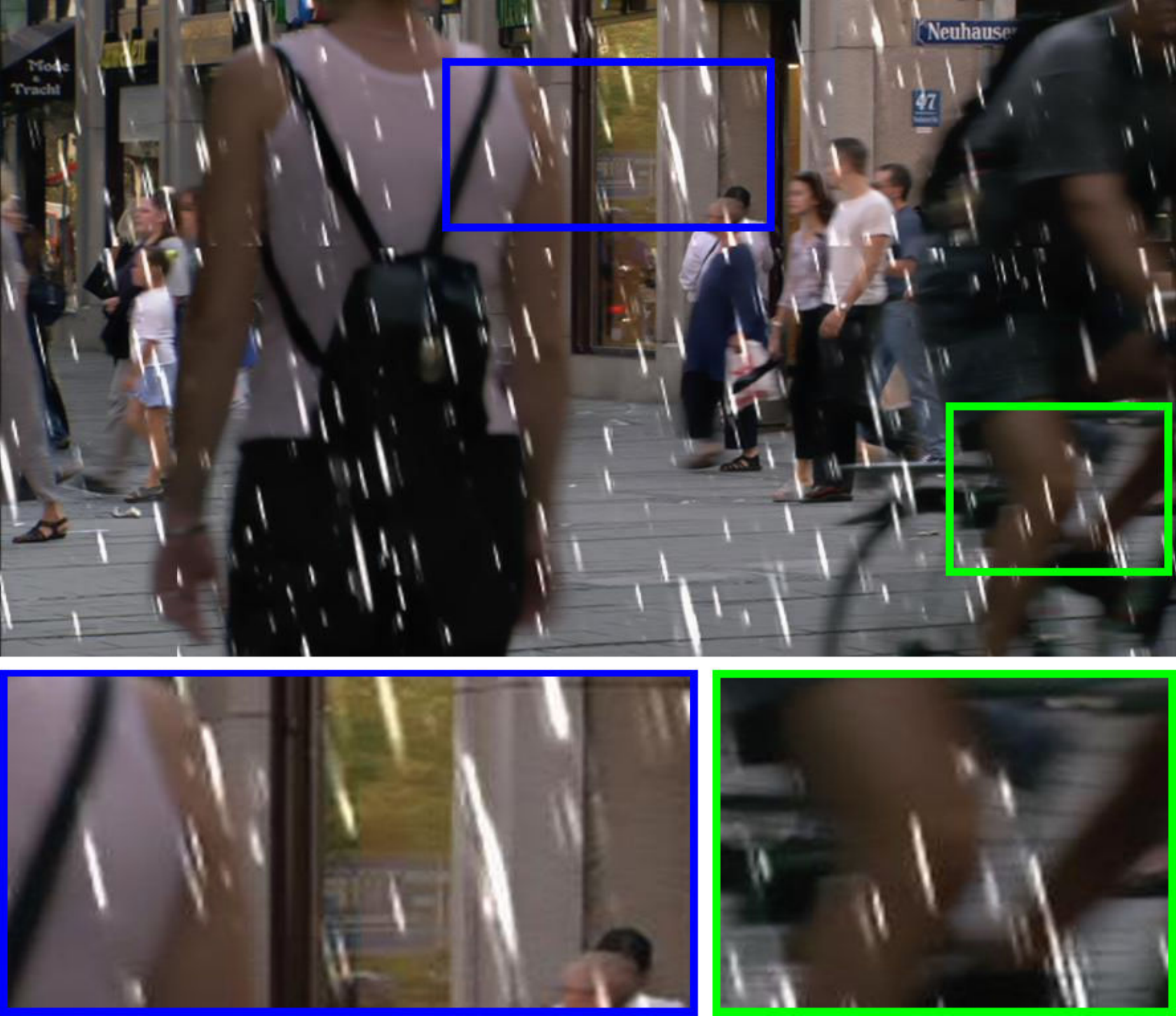}&
		\includegraphics[width=0.121\linewidth]{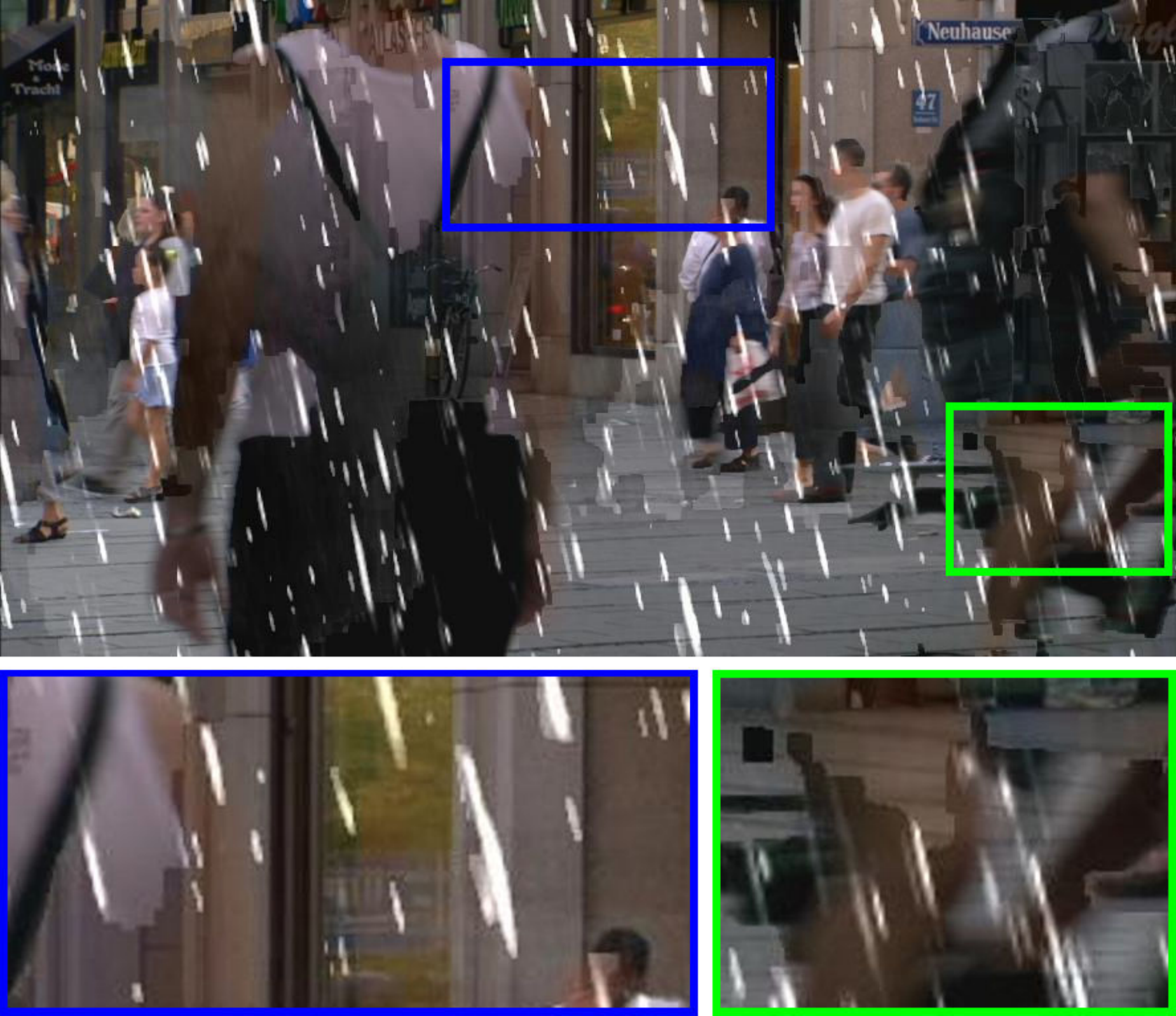}&
		\includegraphics[width=0.121\linewidth]{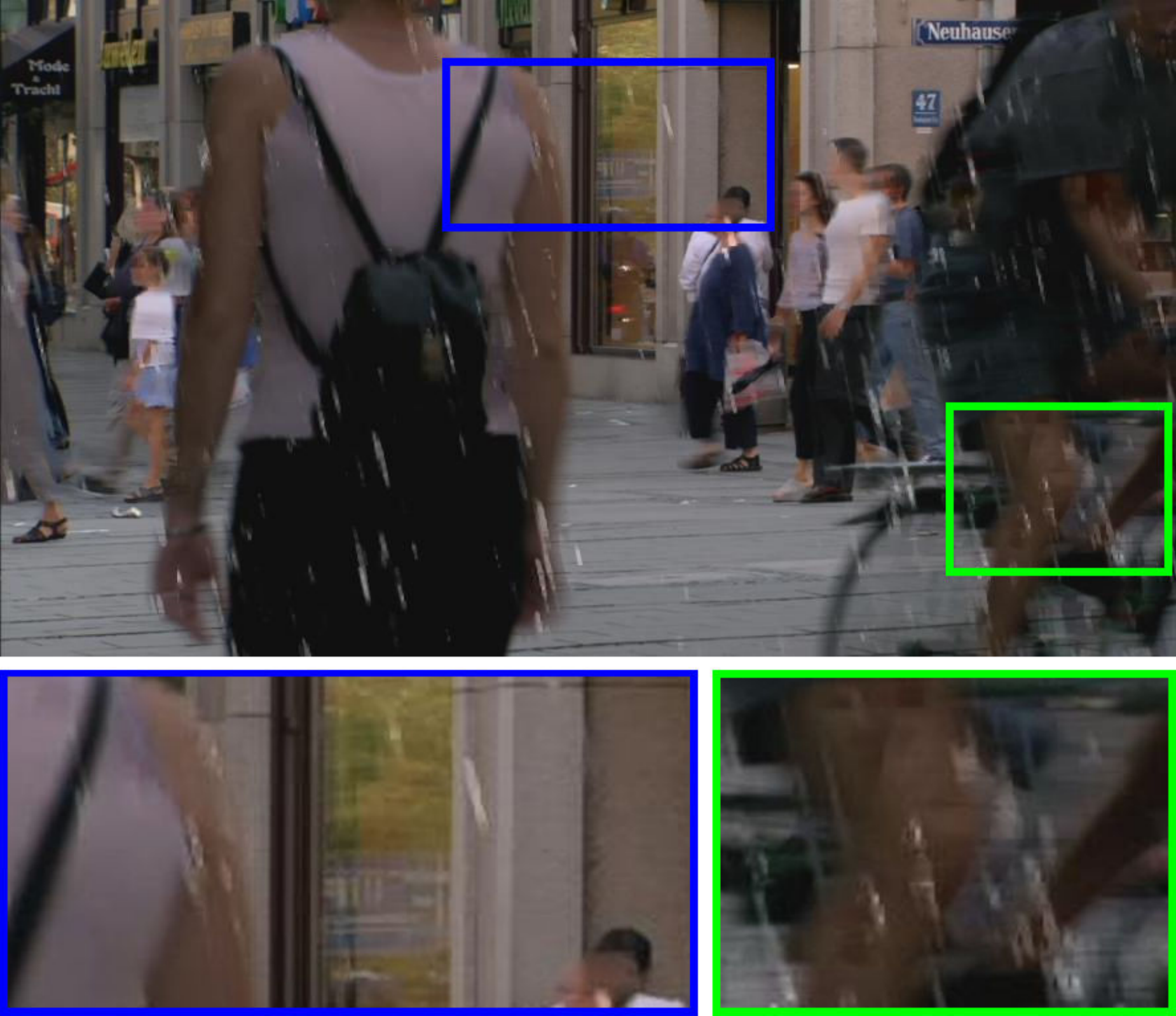}&
		\includegraphics[width=0.121\linewidth]{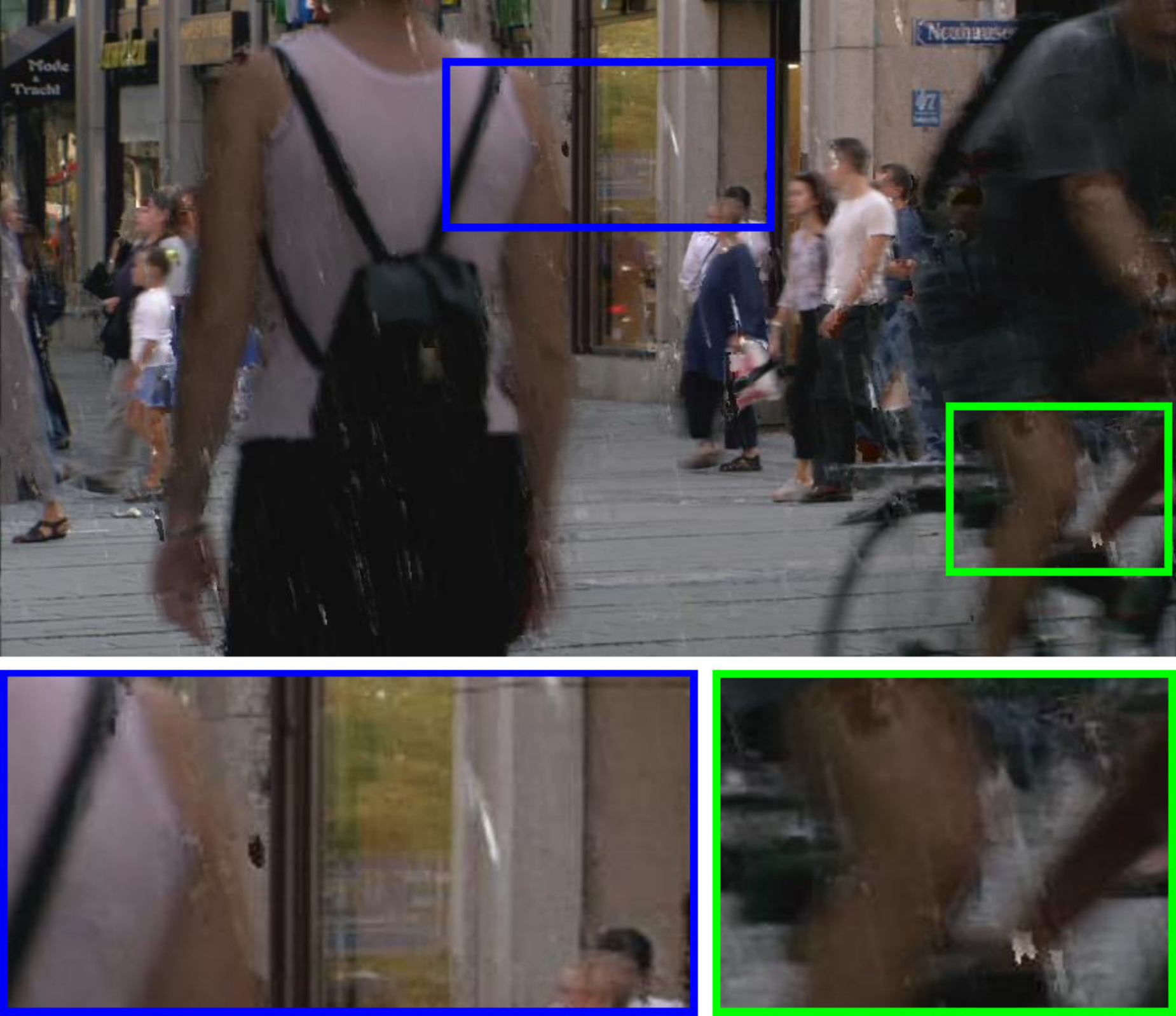}&
		\includegraphics[width=0.121\linewidth]{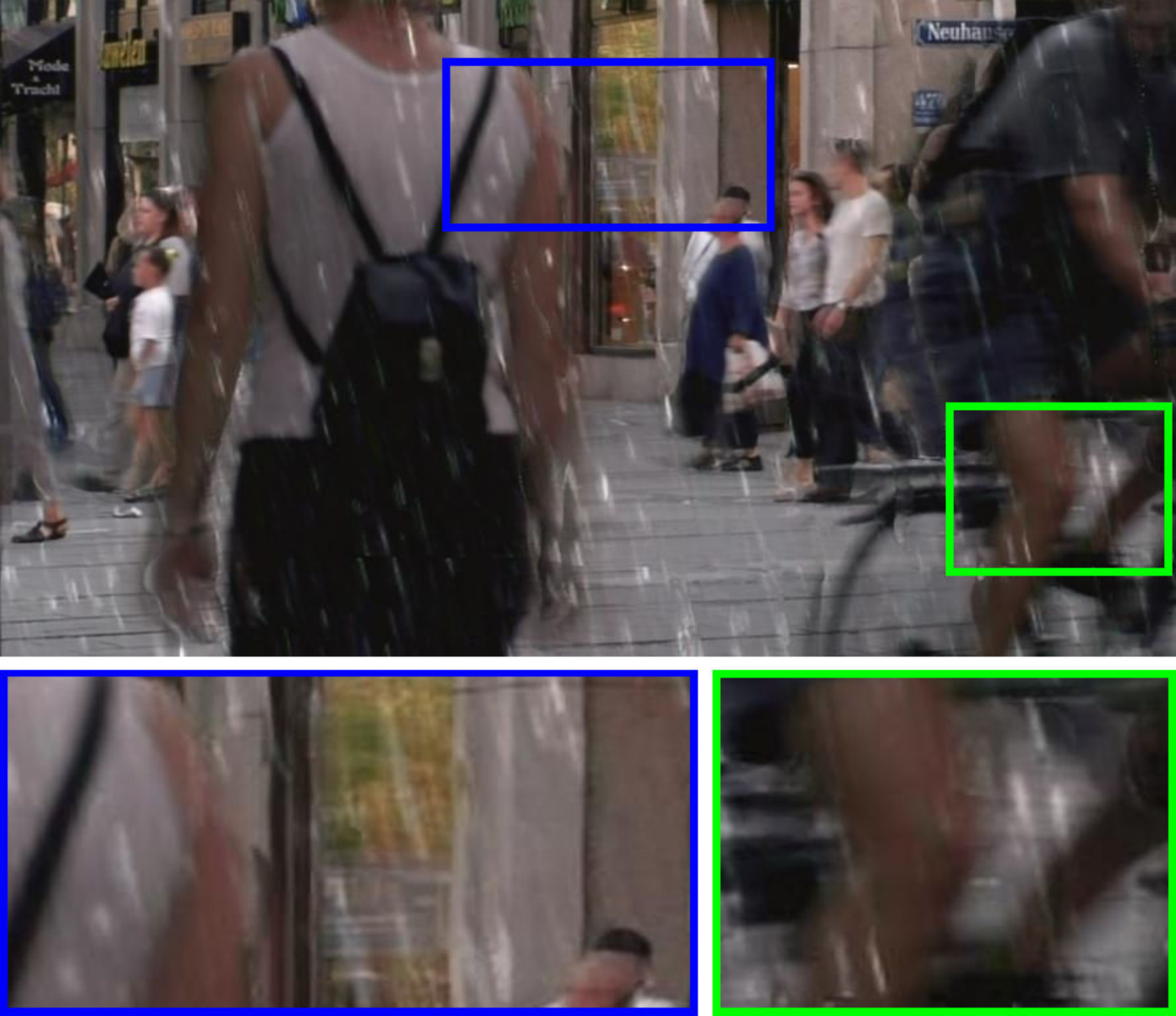}&
		\includegraphics[width=0.121\linewidth]{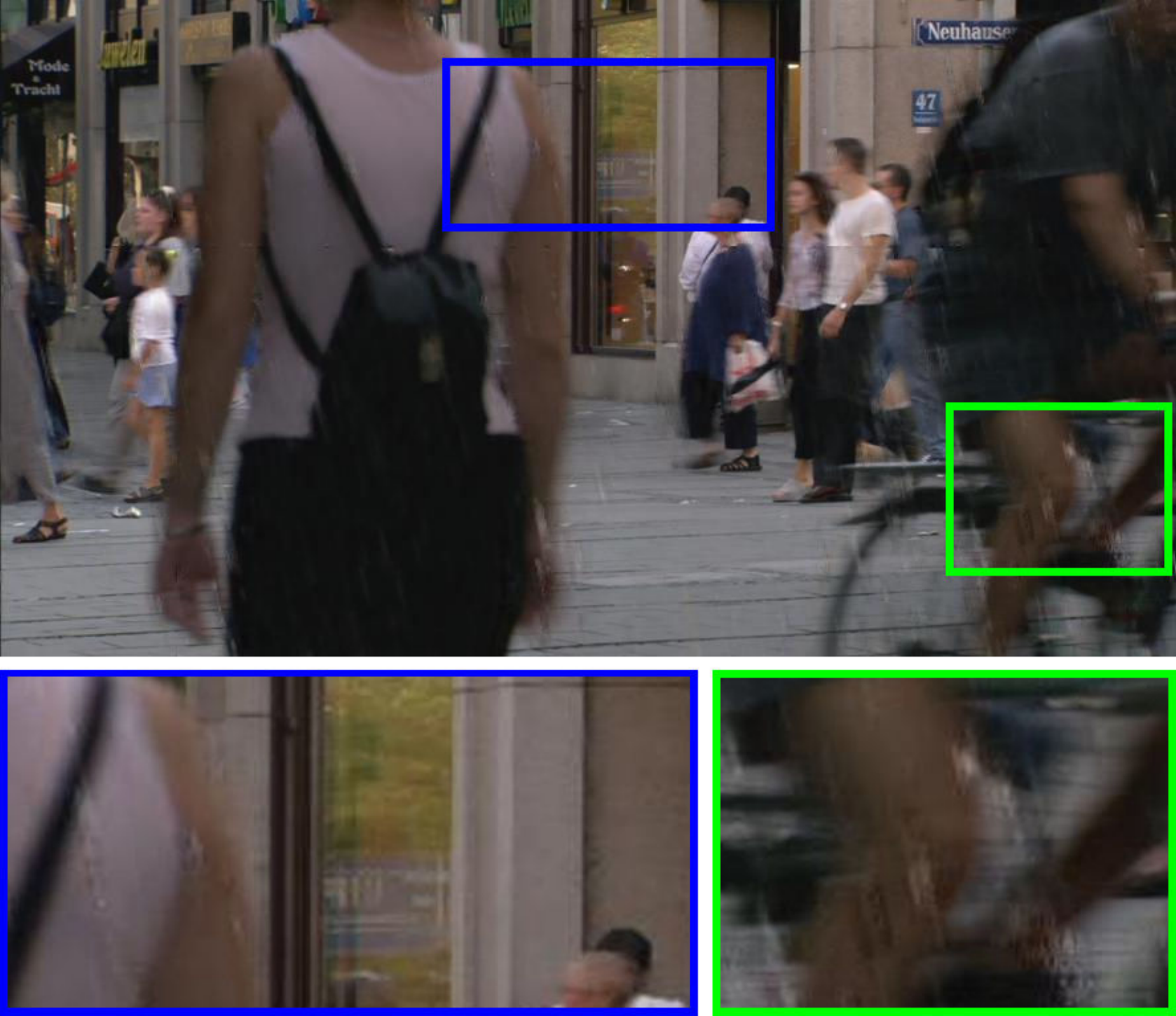}&
		\includegraphics[width=0.121\linewidth]{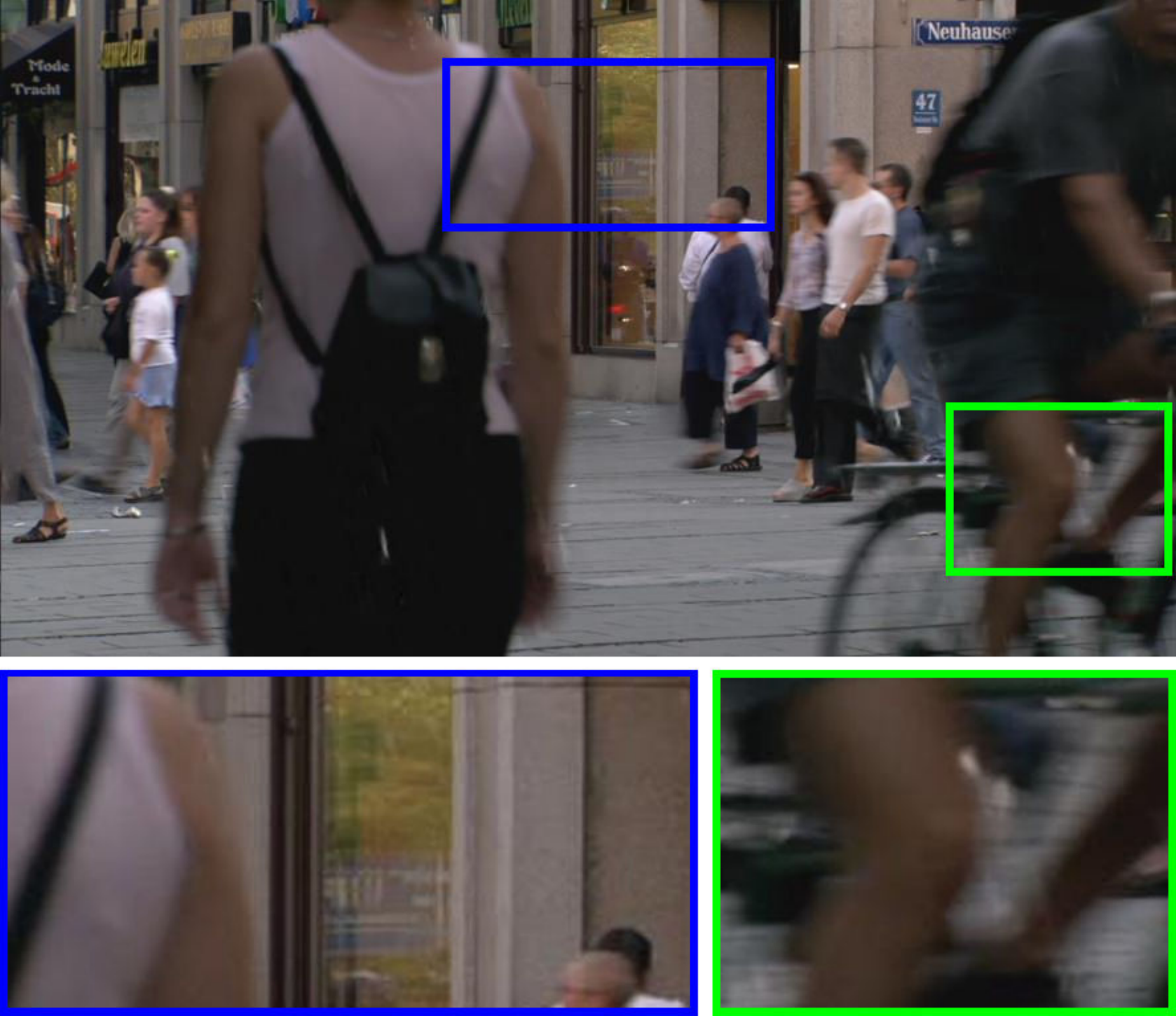}\\
		\includegraphics[width=0.121\linewidth]{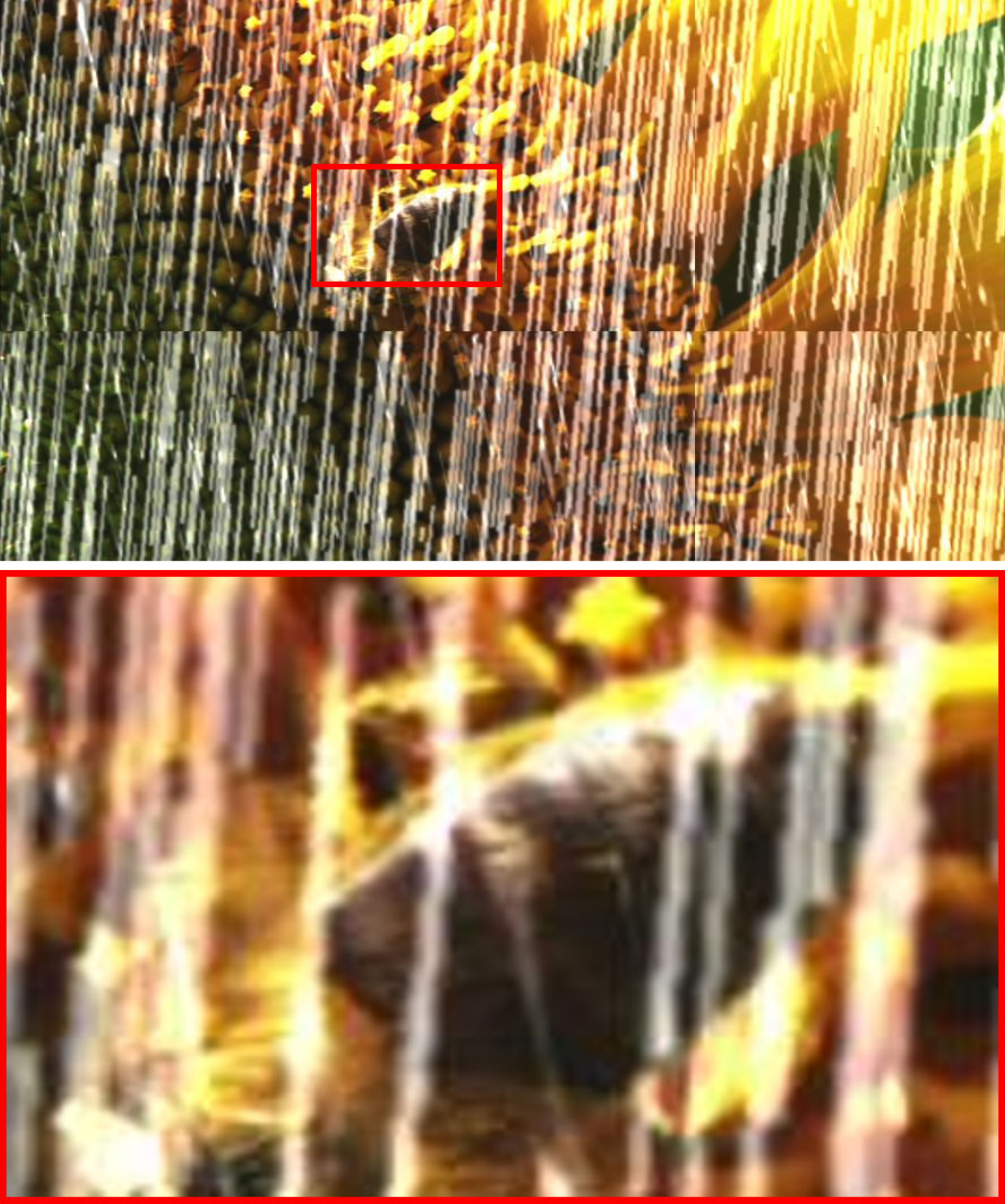}&
		\includegraphics[width=0.121\linewidth]{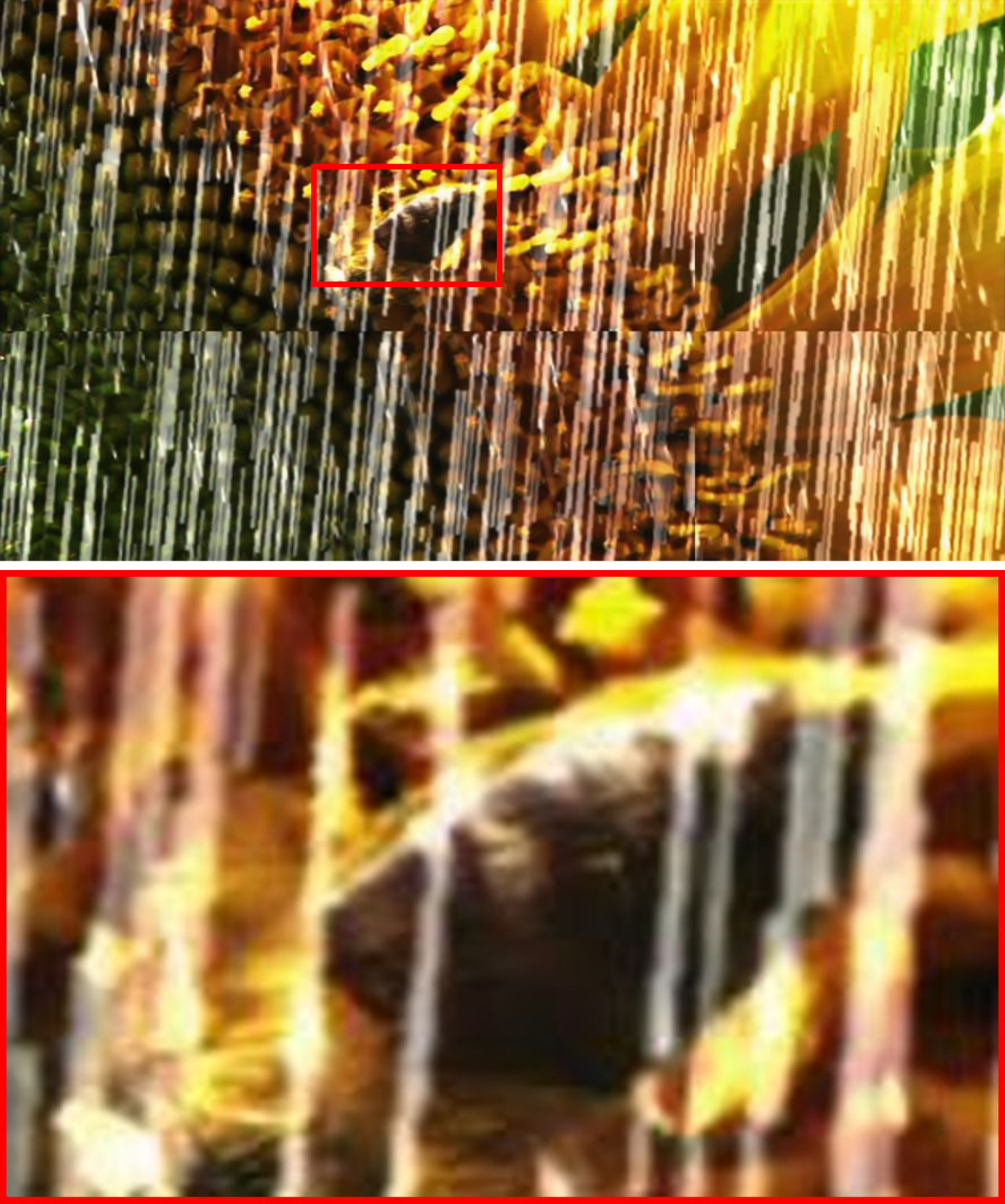}&
		\includegraphics[width=0.121\linewidth]{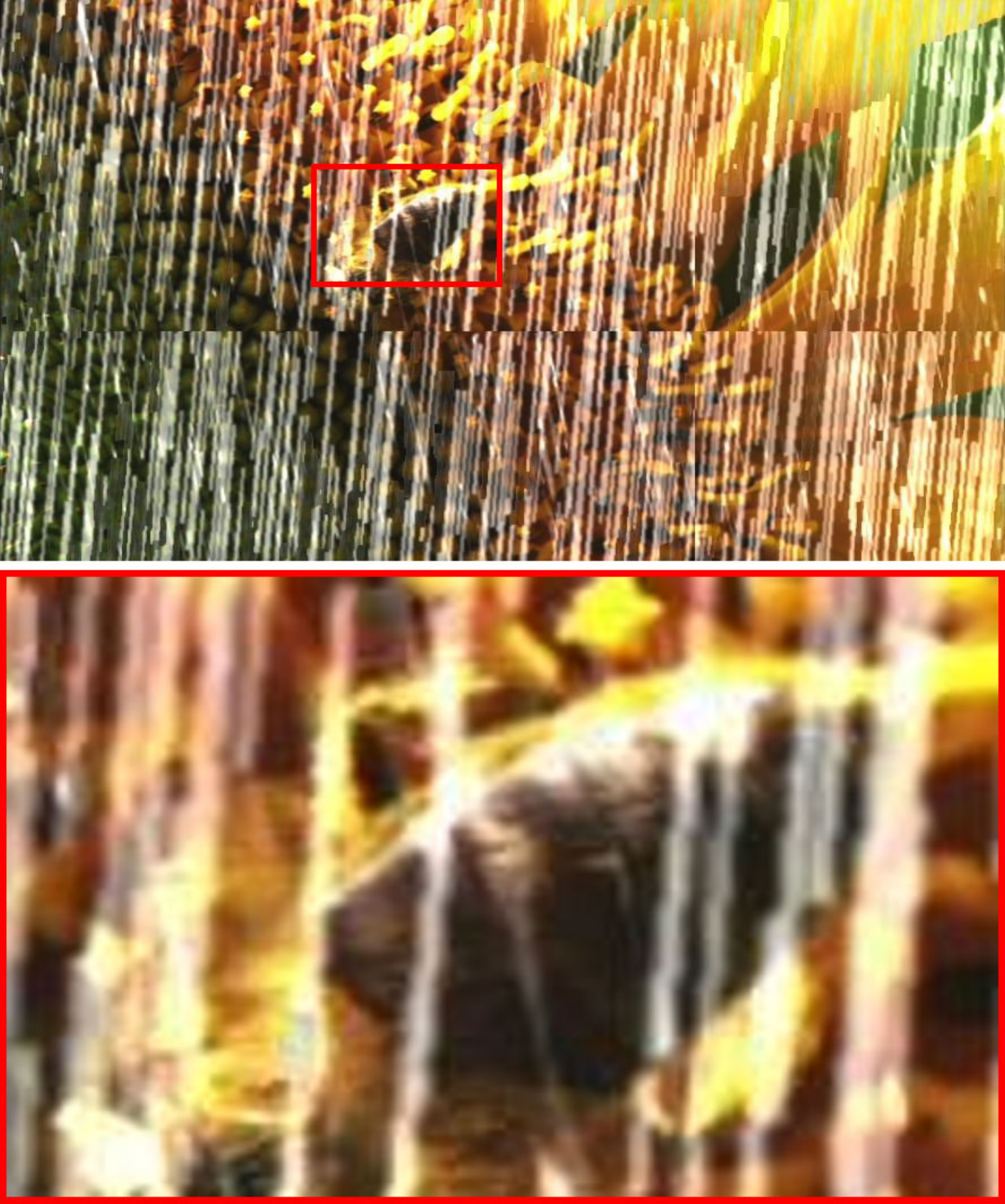}&
		\includegraphics[width=0.121\linewidth]{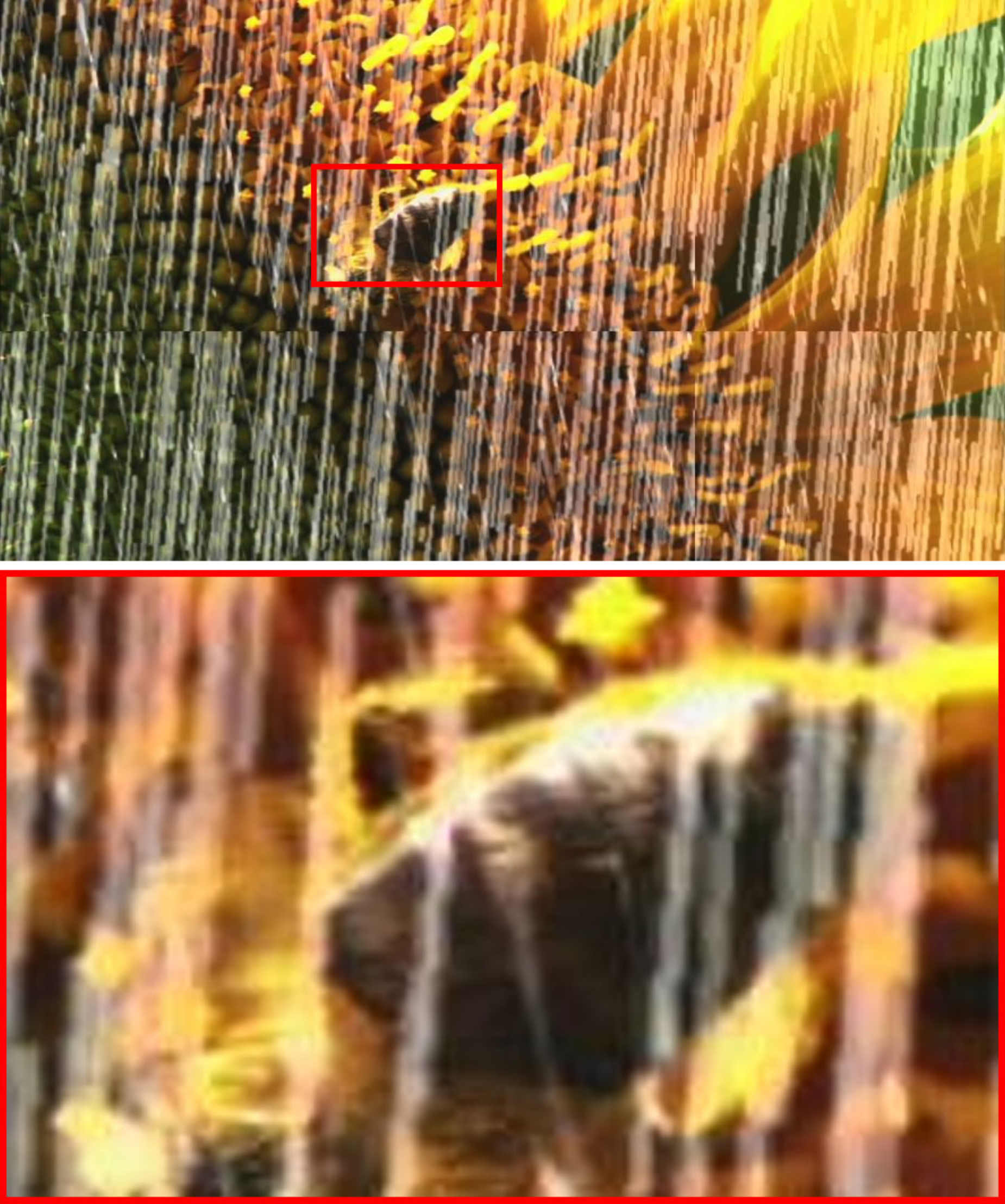}&
		\includegraphics[width=0.121\linewidth]{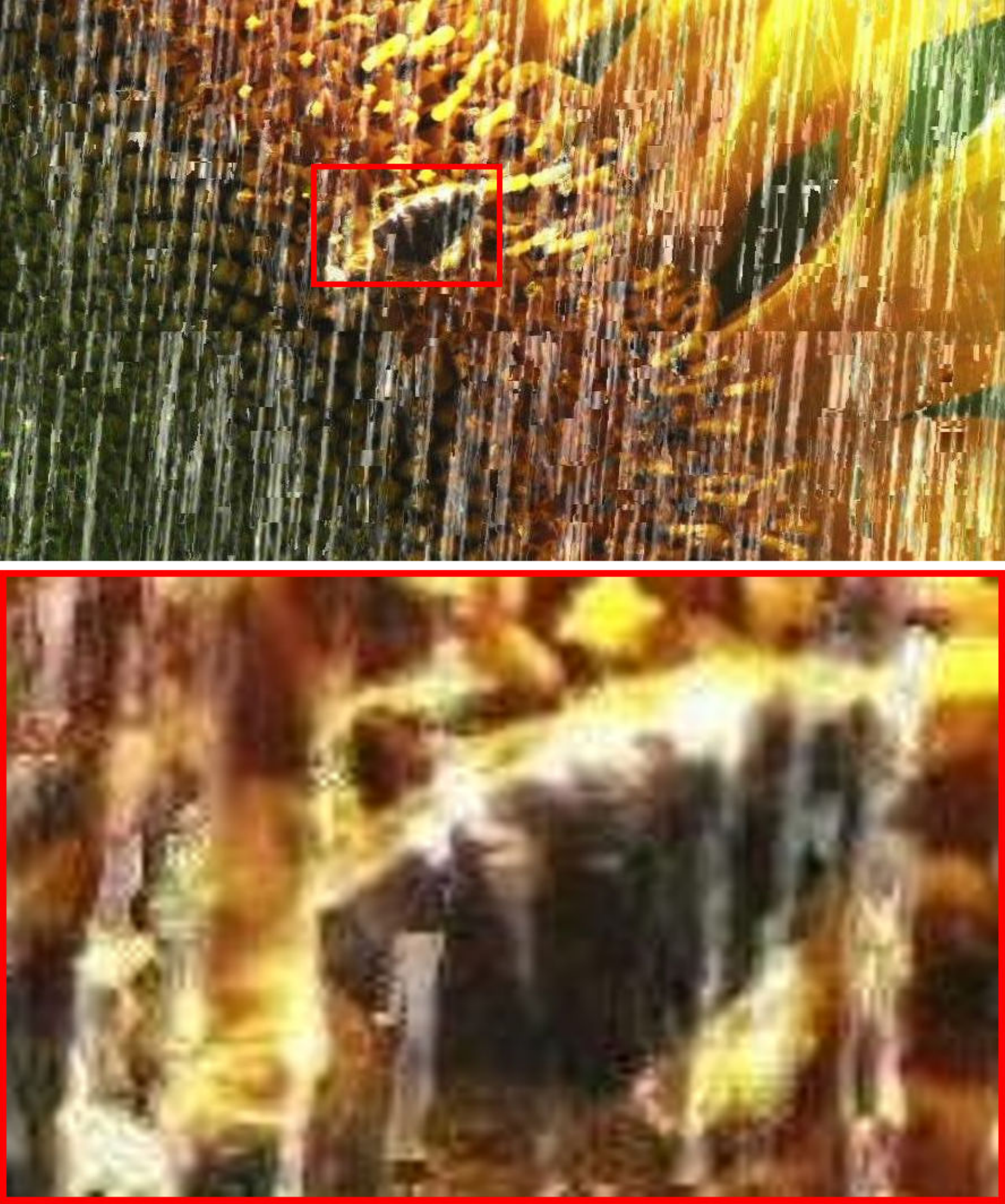}&
		\includegraphics[width=0.121\linewidth]{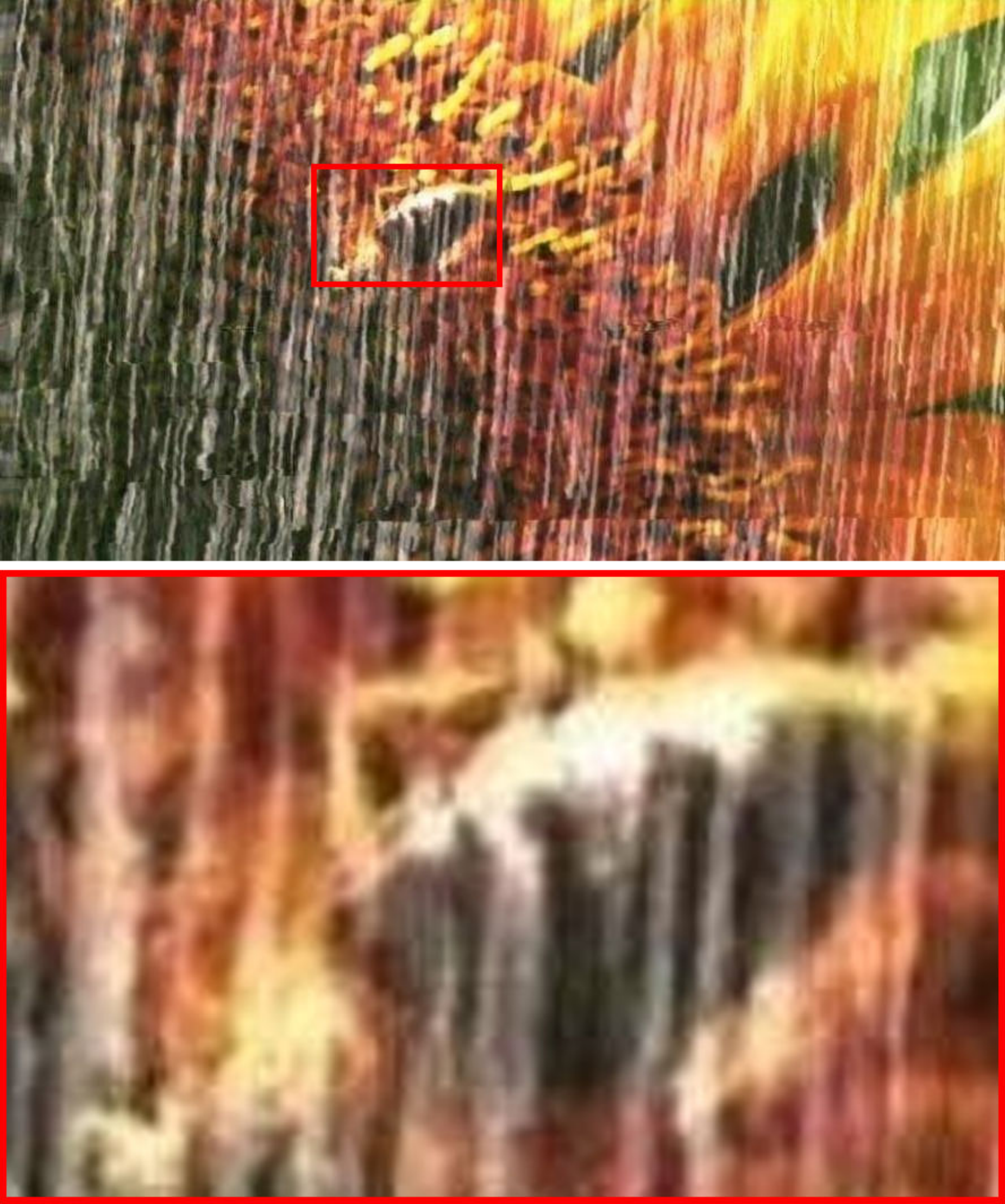}&
		\includegraphics[width=0.121\linewidth]{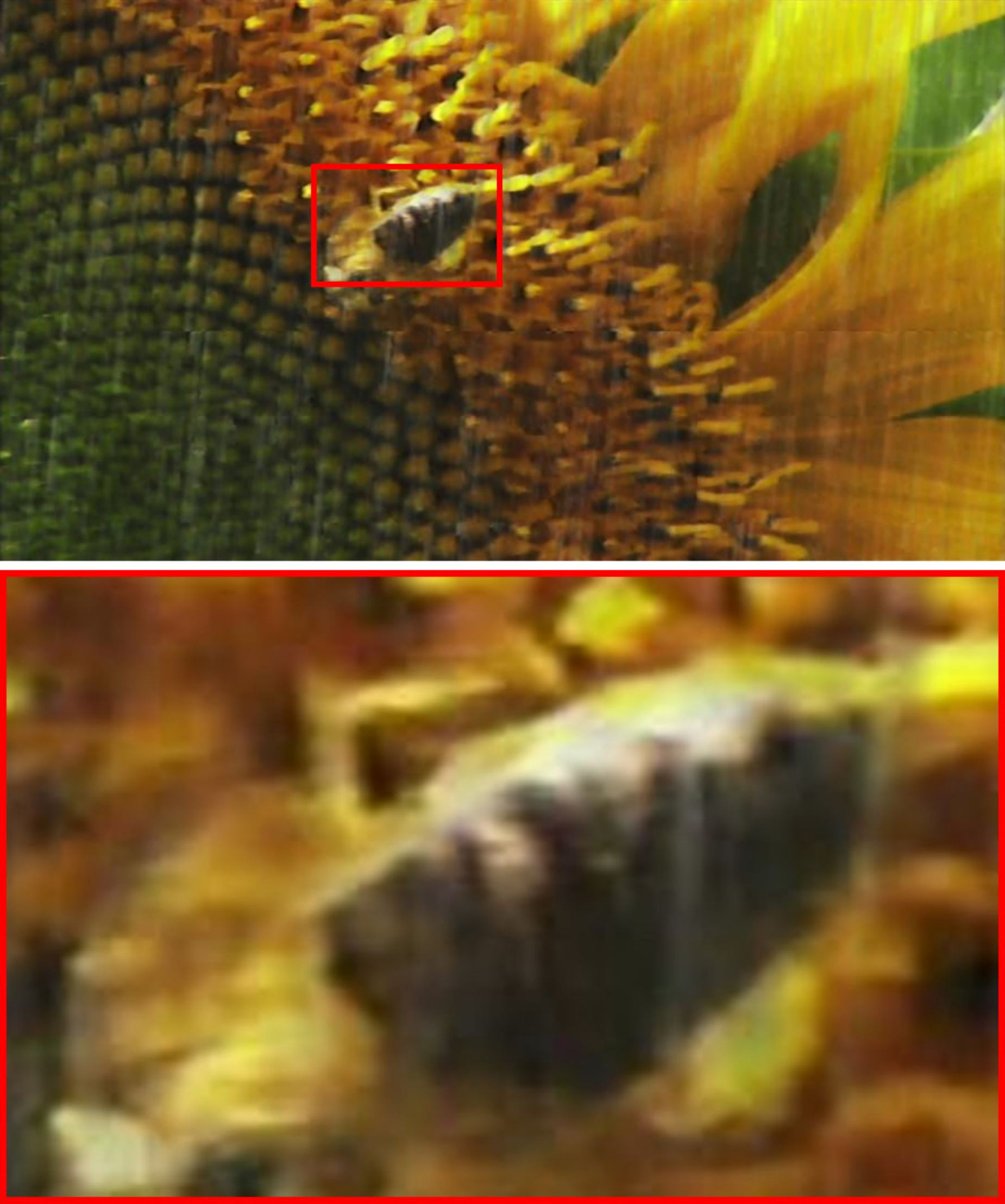}&
		\includegraphics[width=0.121\linewidth]{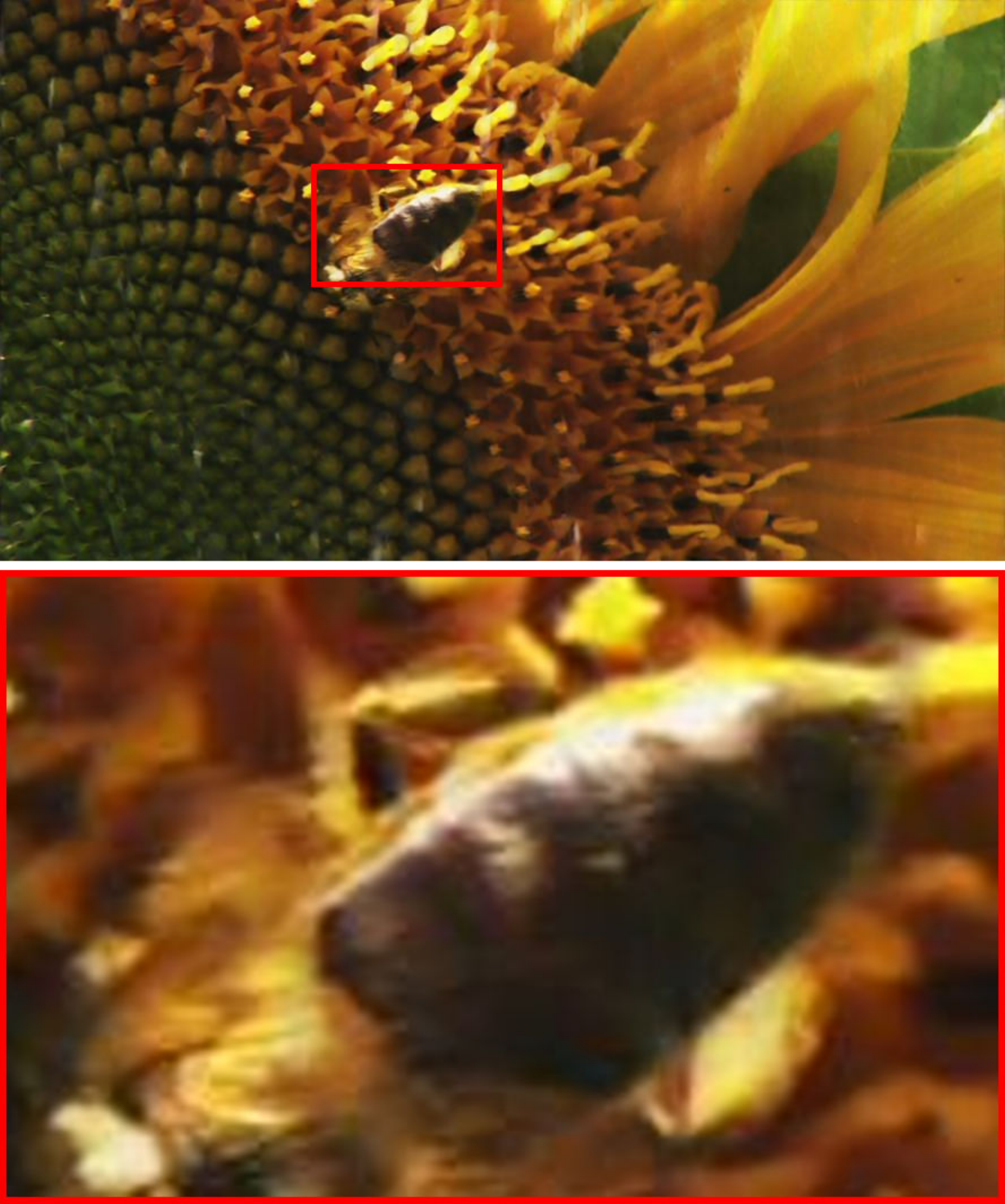}\\		
           \includegraphics[width=0.121\linewidth]{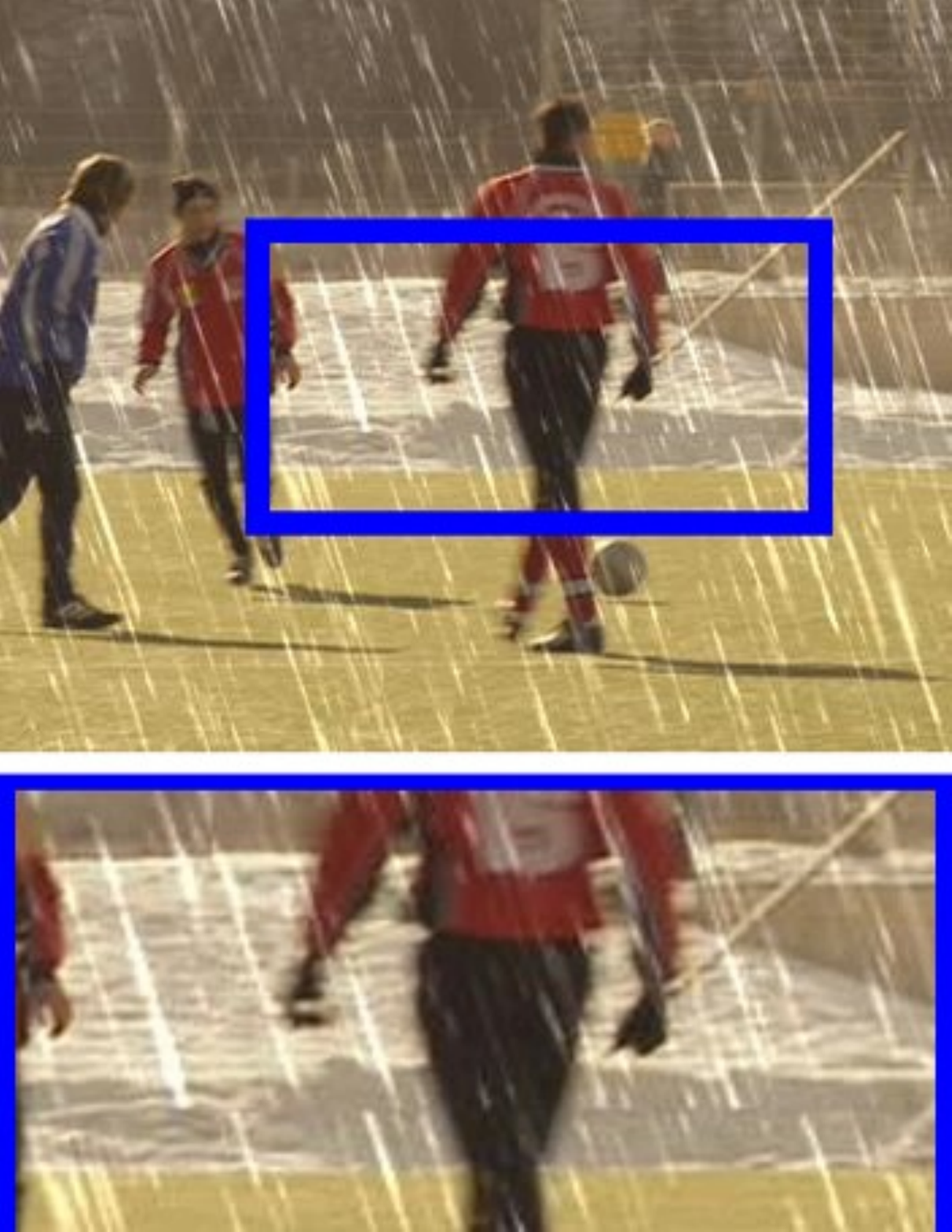}&
		\includegraphics[width=0.121\linewidth]{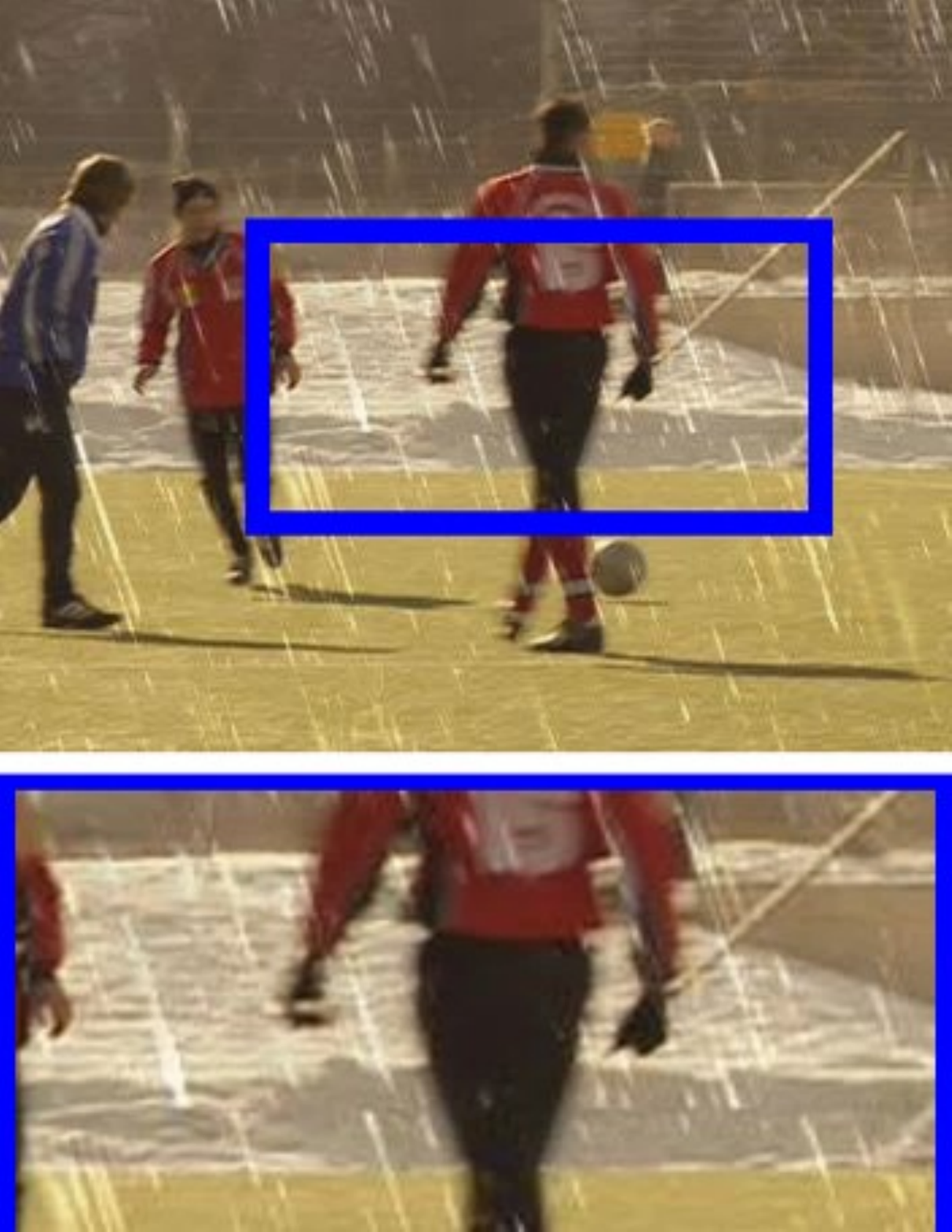}&   
		\includegraphics[width=0.121\linewidth]{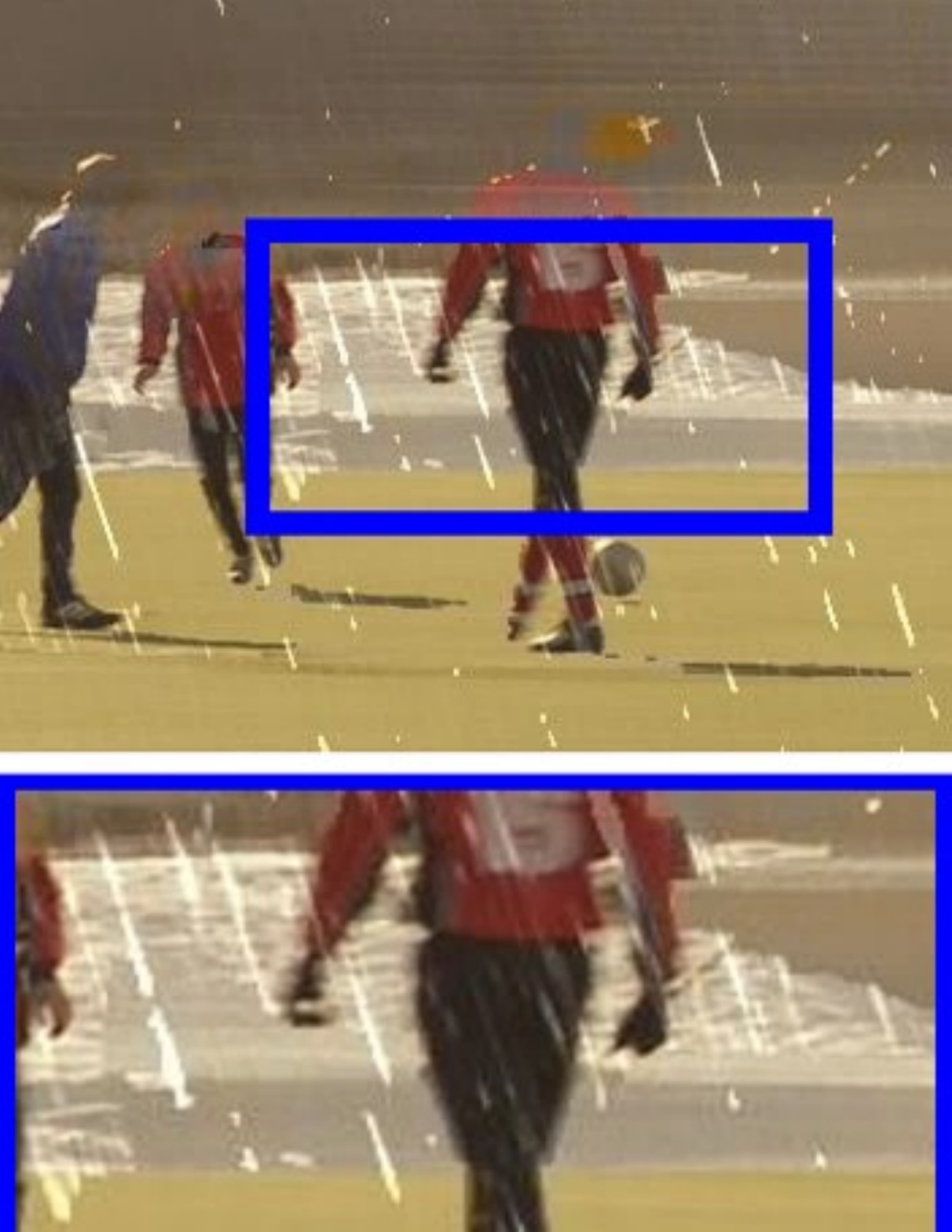}&
		\includegraphics[width=0.121\linewidth]{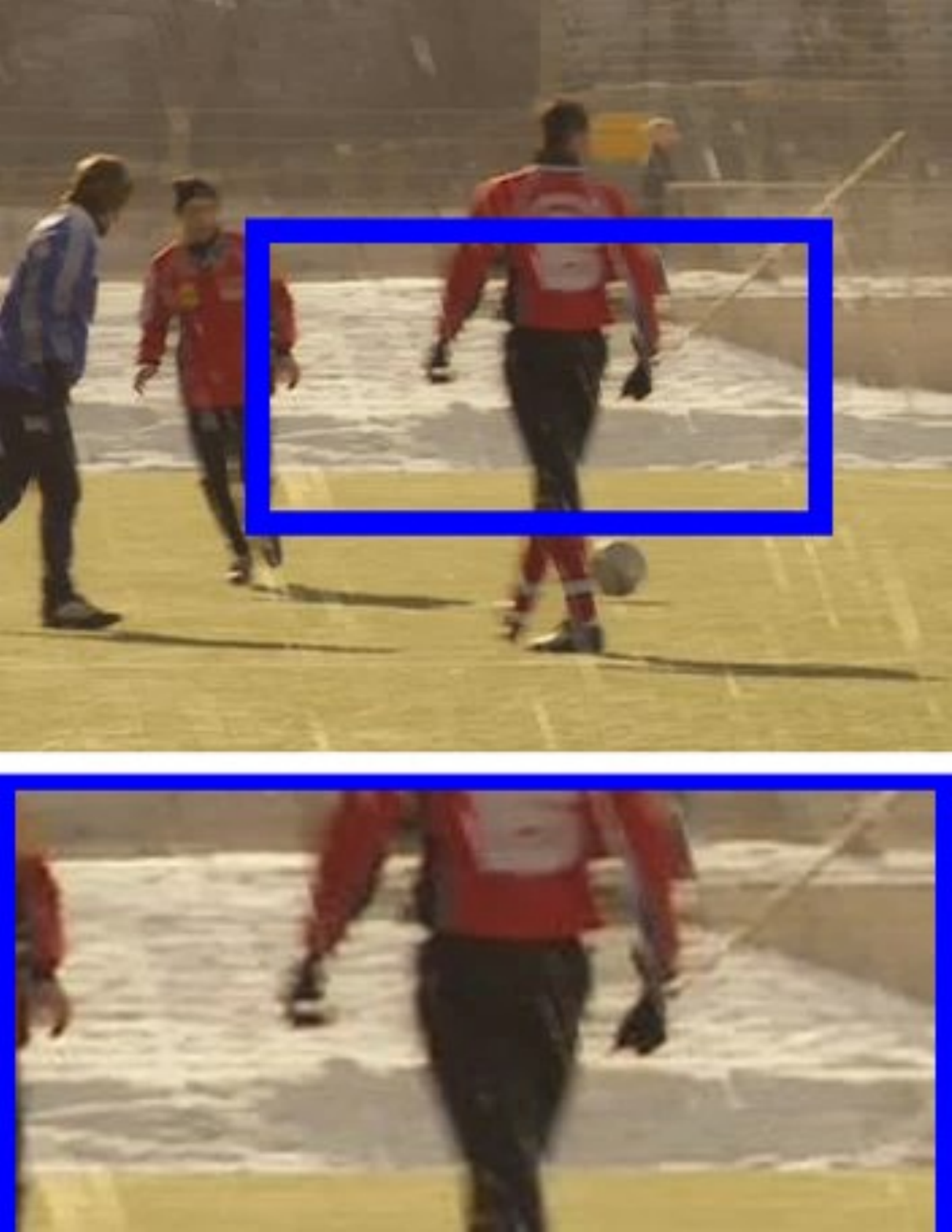}&
		\includegraphics[width=0.121\linewidth]{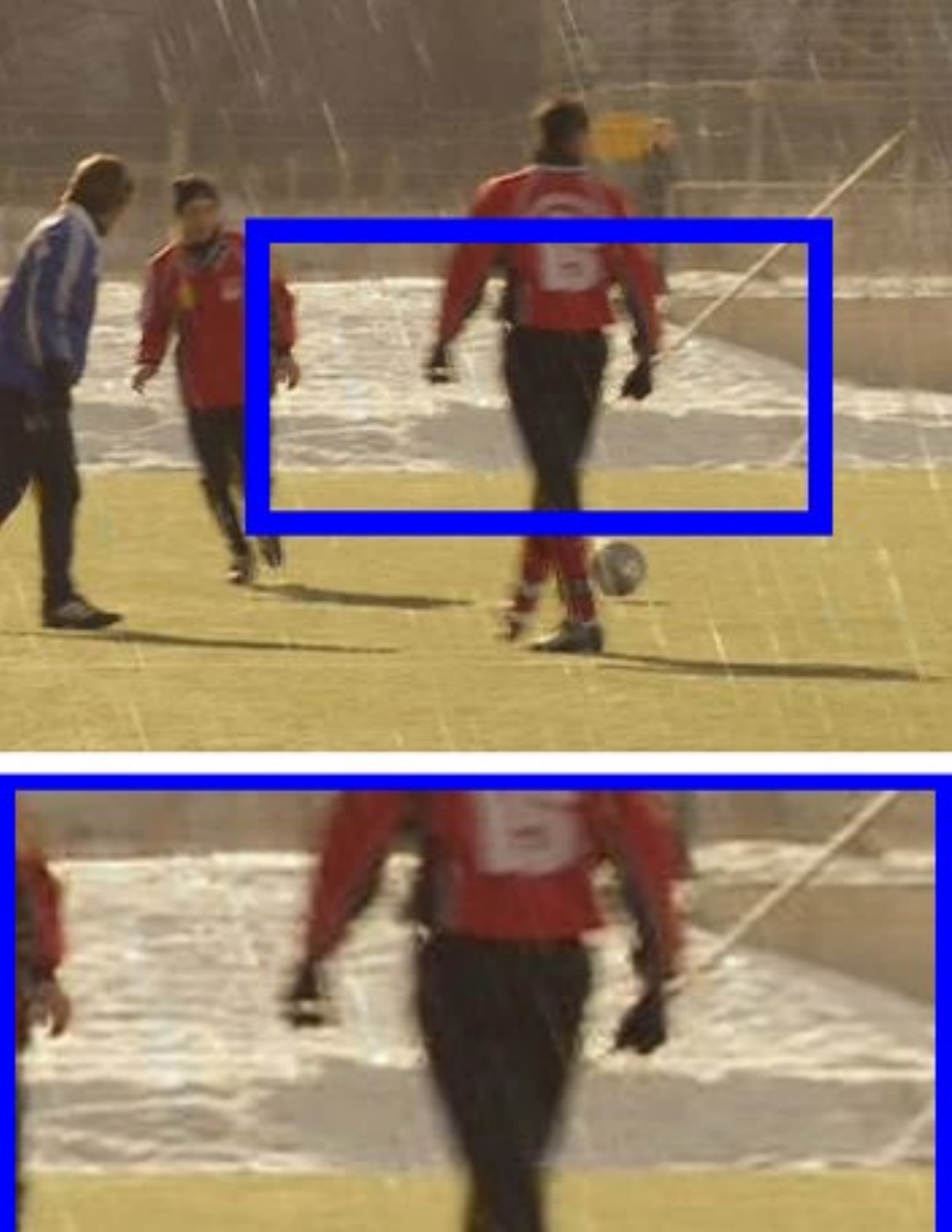}&
		\includegraphics[width=0.121\linewidth]{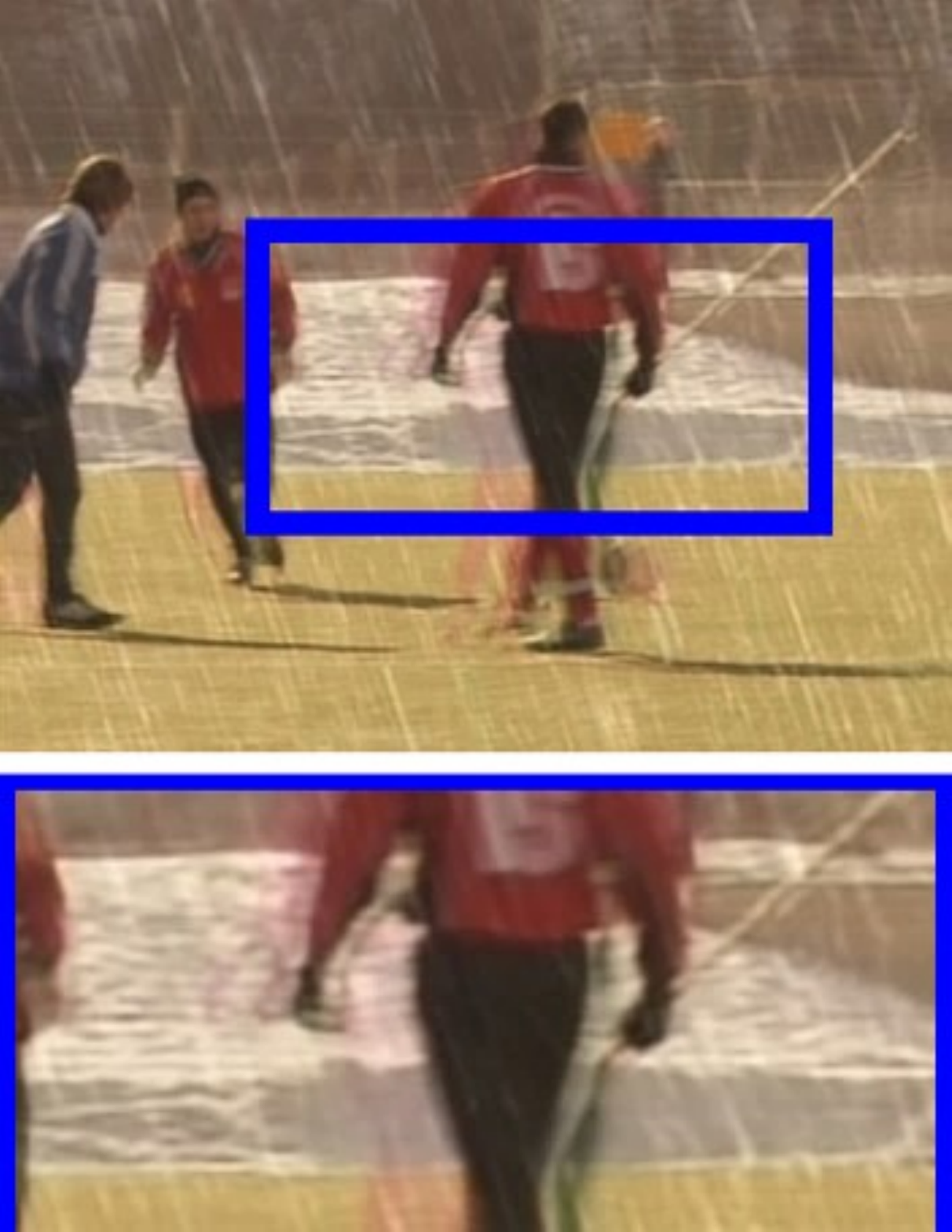}&
		\includegraphics[width=0.121\linewidth]{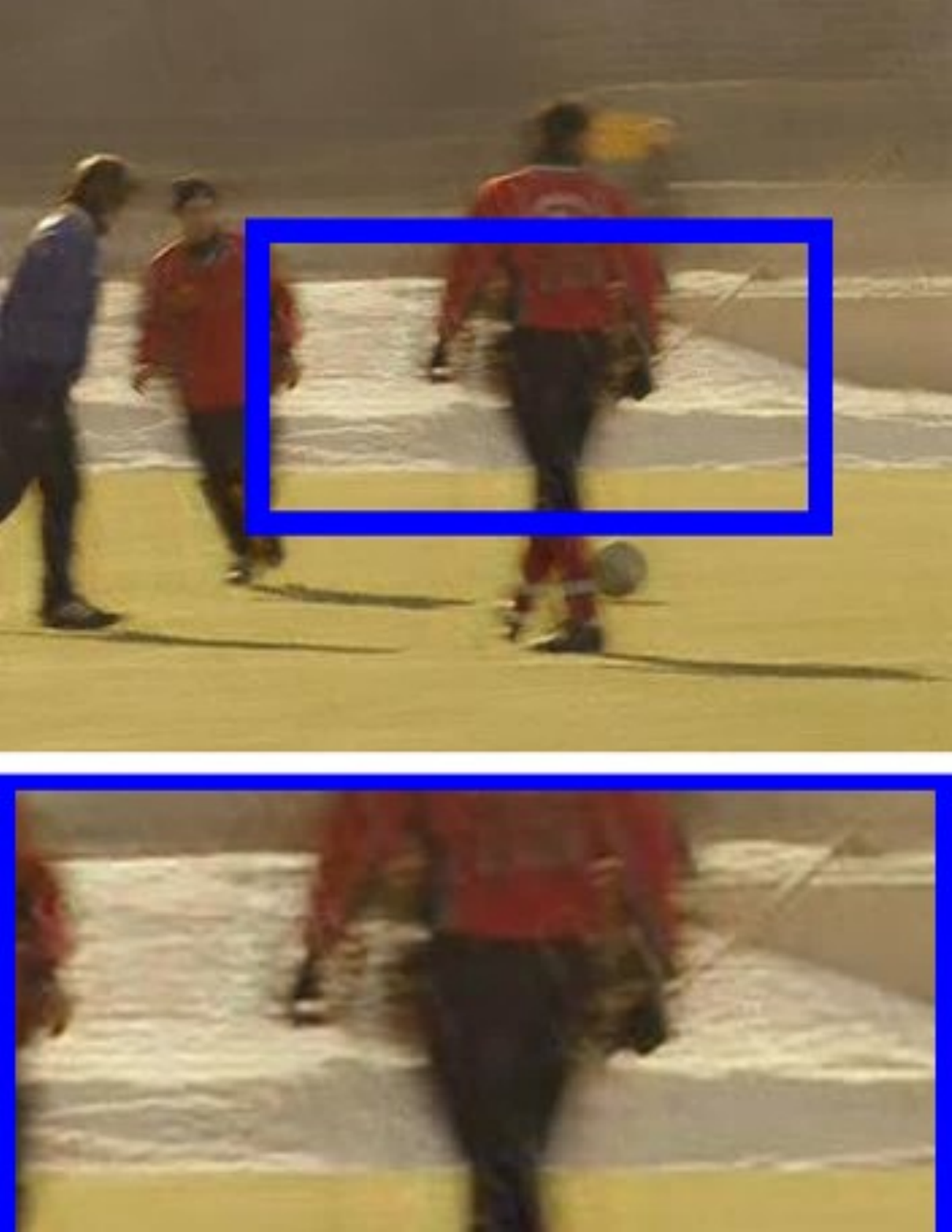}&
		\includegraphics[width=0.121\linewidth]{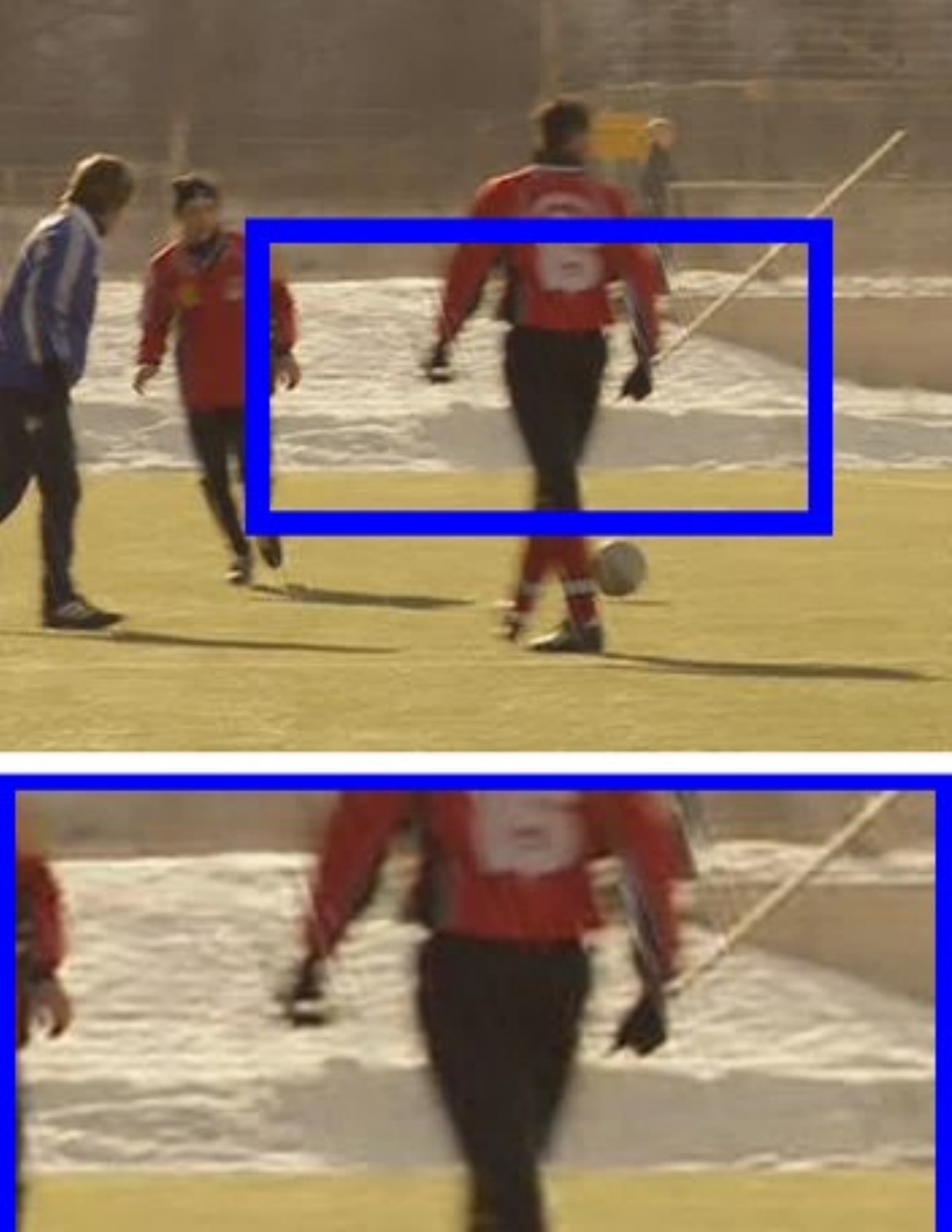}\\		
		
		\footnotesize{Input}&\footnotesize{MOSS}&\footnotesize MSCSC&\footnotesize{FastDerain}&\footnotesize SpacCNN&\footnotesize SLDNet&\footnotesize S2VD&\footnotesize Ours\\
	\end{tabular}	
	\caption{Visual comparisons on different datasets with sate-of-the-art methods. The first row shows the results on Light25. The second and last rows show the results on Complex25. Our method achieves better visual representation on both synthetic rain datasets.}
	\label{fig:fig_other}
	\vspace{-0.5cm}
\end{figure*}	

\begin{figure*}[!t]
	\centering
	\begin{tabular}{c@{\extracolsep{0.1em}}c@{\extracolsep{0.1em}}c@{\extracolsep{0.1em}}c@{\extracolsep{0.1em}}c@{\extracolsep{0.1em}}c@{\extracolsep{0.1em}}c@{\extracolsep{0.1em}}c}
		\includegraphics[width=0.121\linewidth]{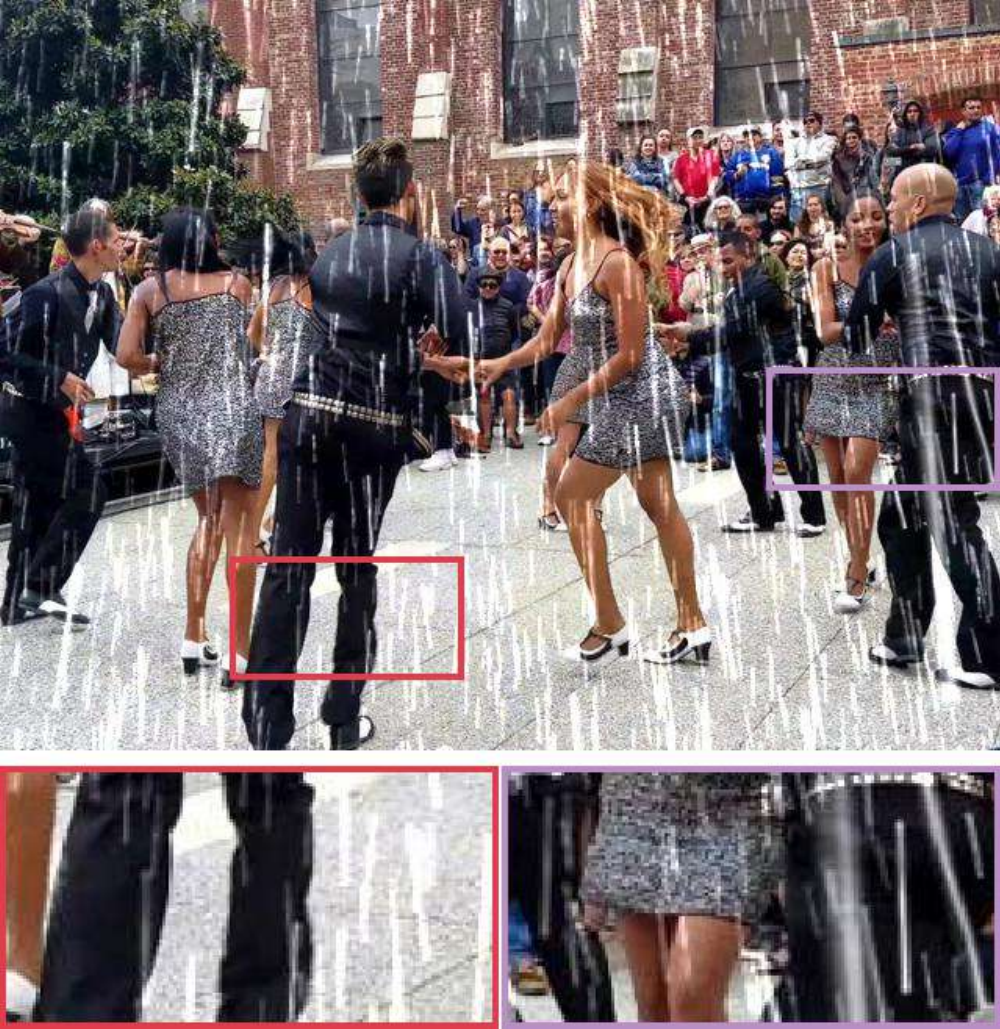}&
		\includegraphics[width=0.121\linewidth]{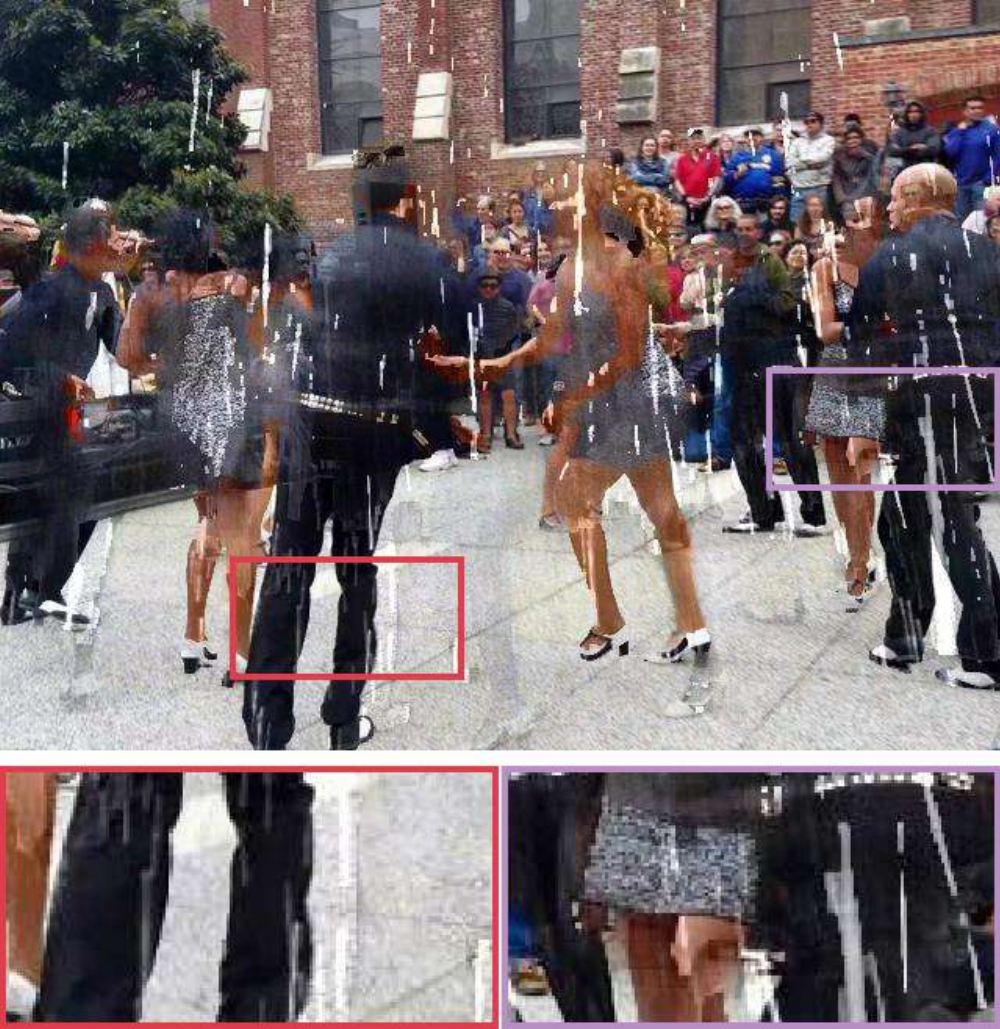}&
		\includegraphics[width=0.121\linewidth]{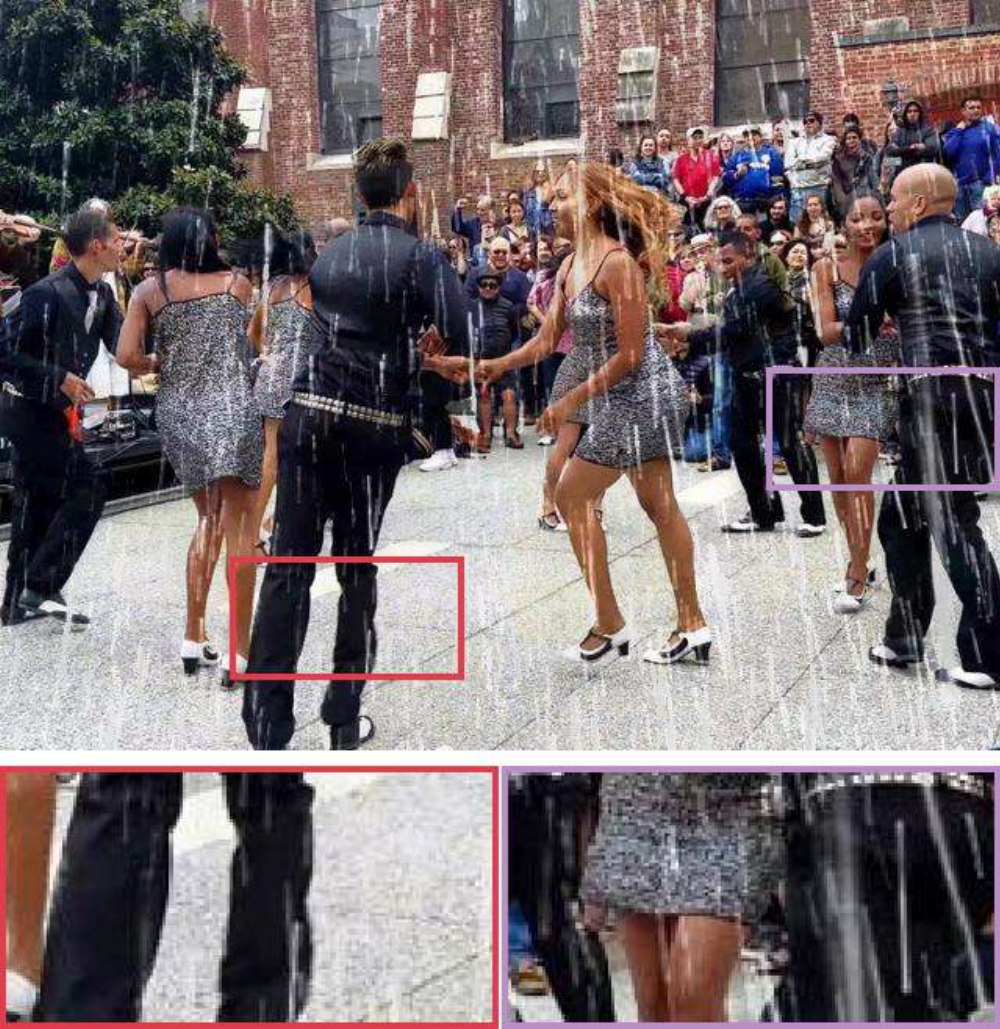}&
		\includegraphics[width=0.121\linewidth]{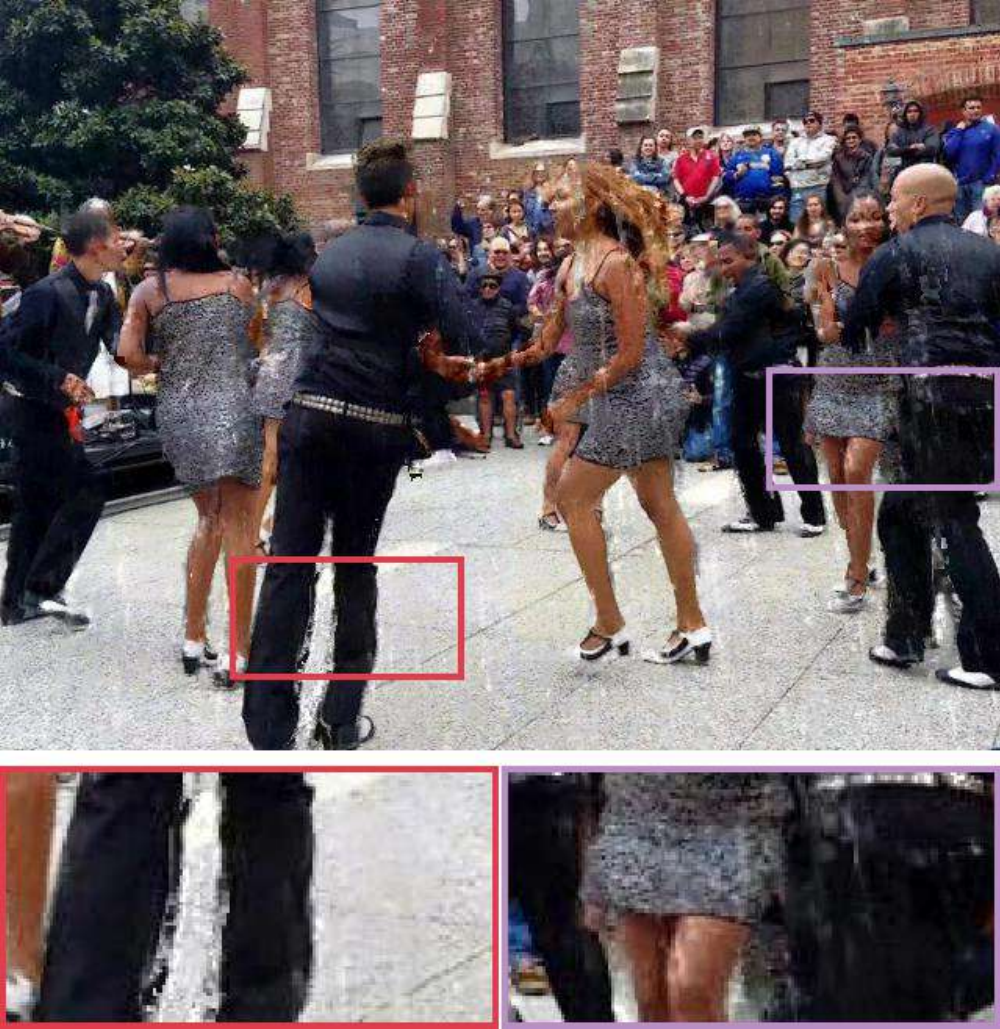}&
		\includegraphics[width=0.121\linewidth]{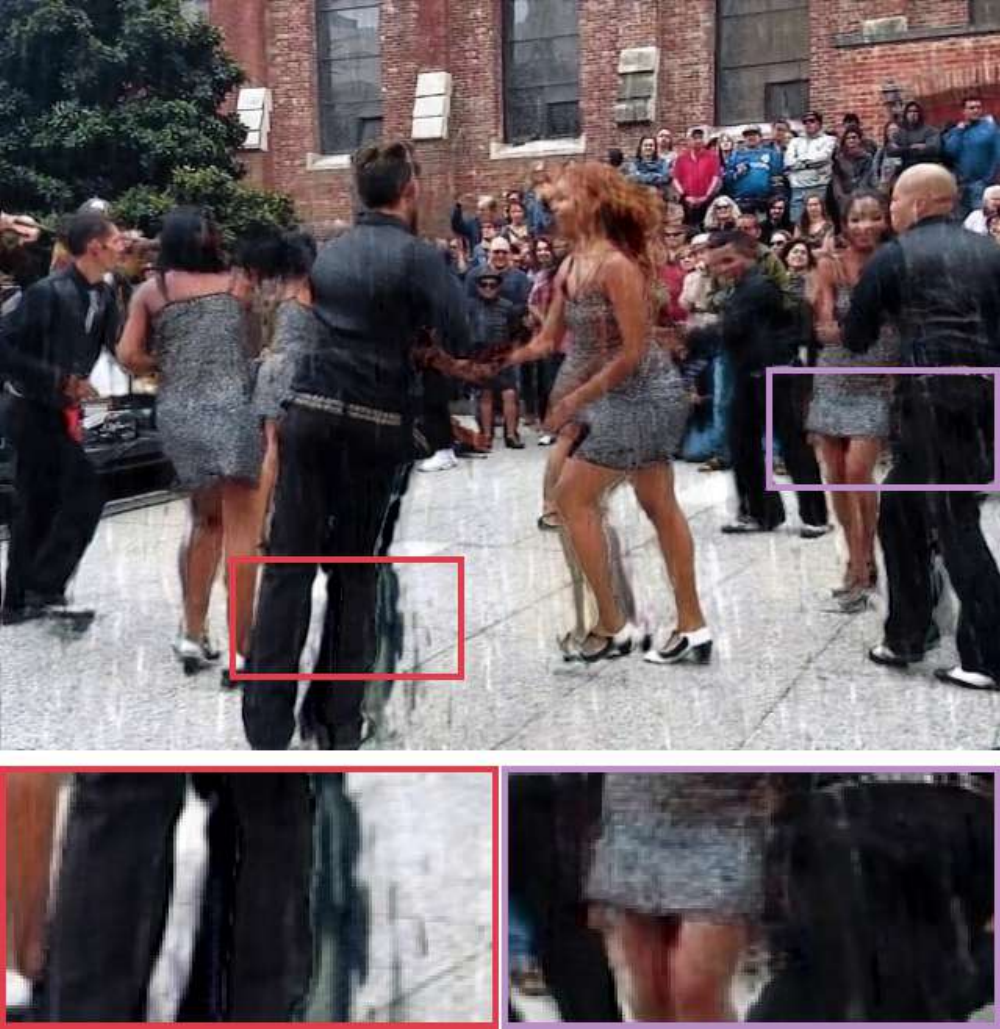}&
		\includegraphics[width=0.121\linewidth]{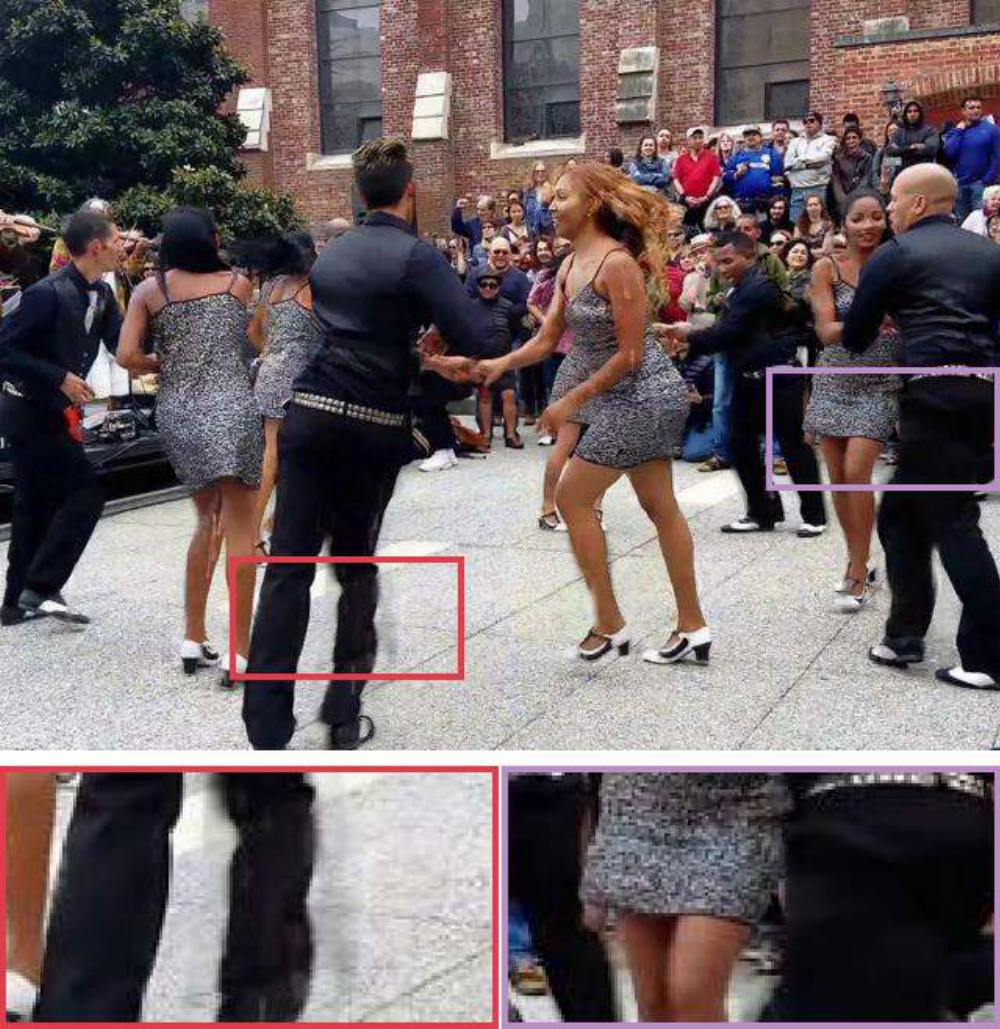}&
		\includegraphics[width=0.121\linewidth]{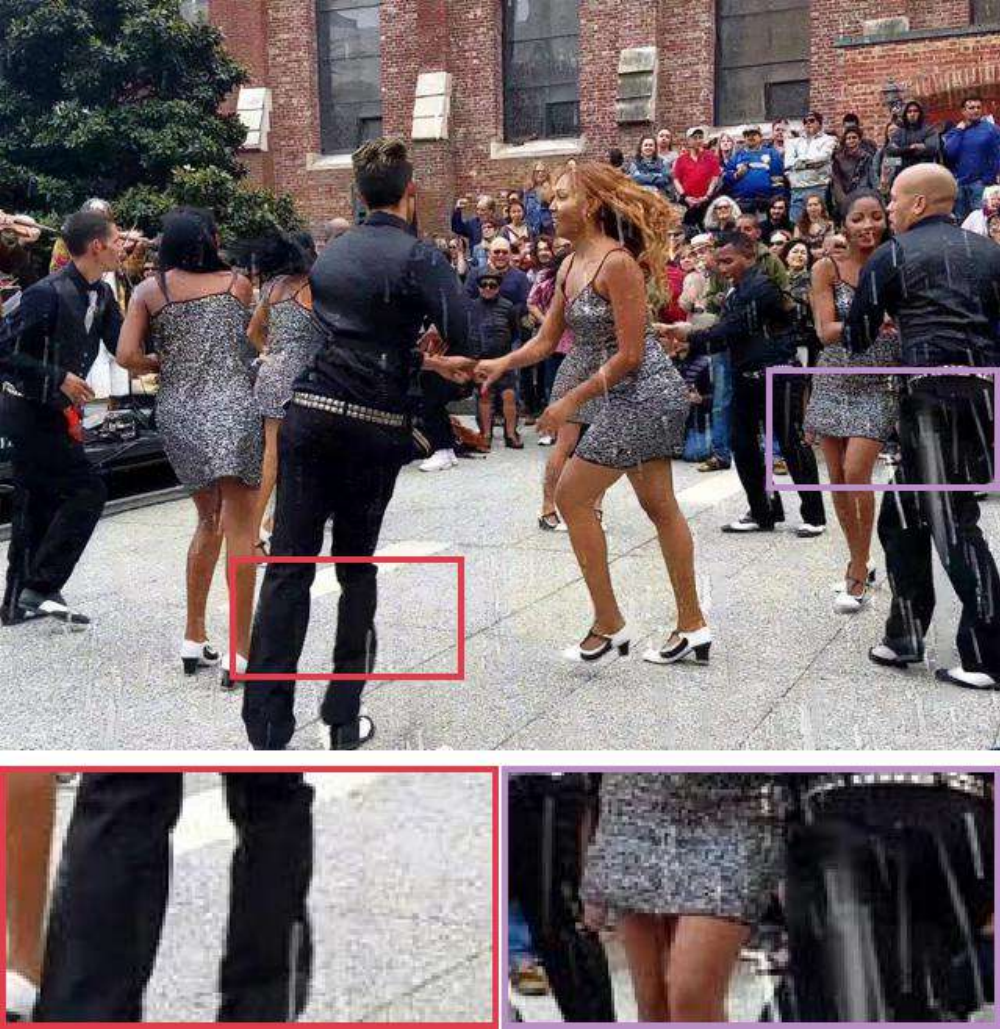}&
		\includegraphics[width=0.121\linewidth]{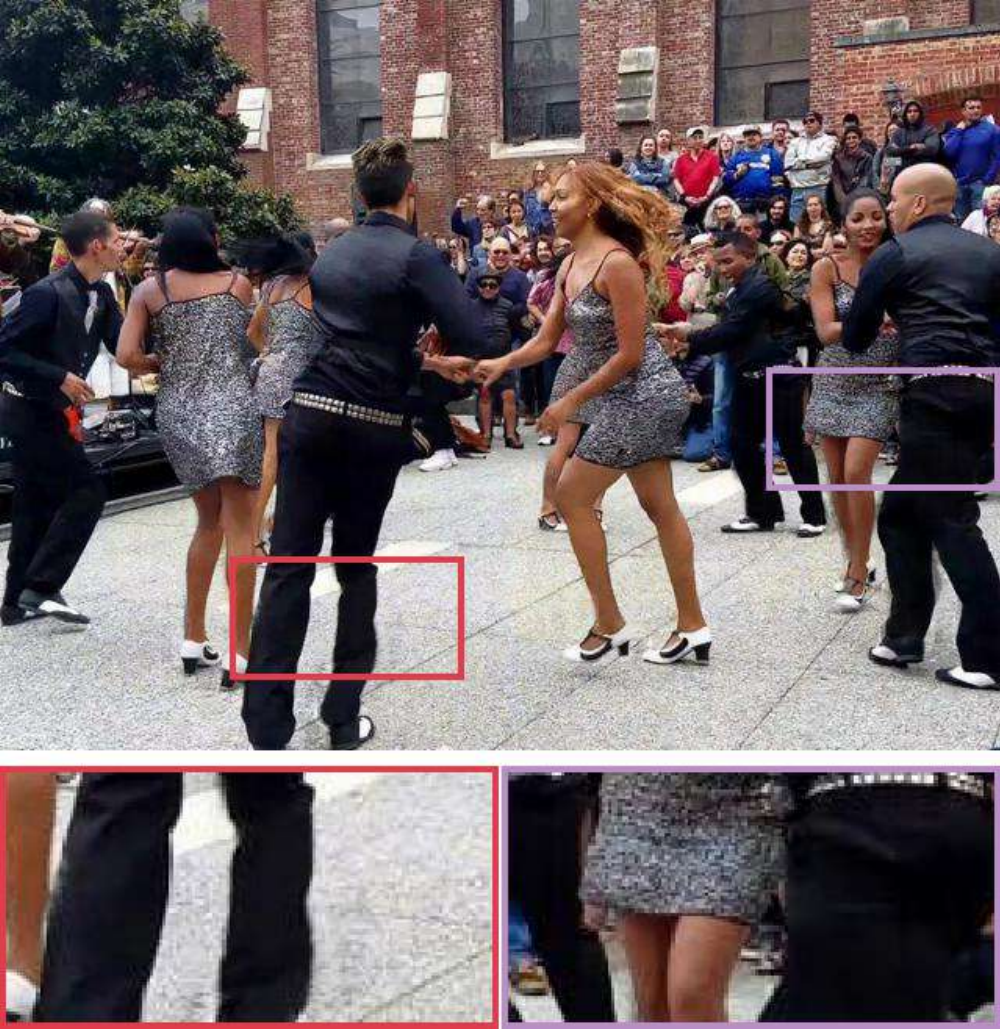}\\
		\includegraphics[width=0.121\linewidth]{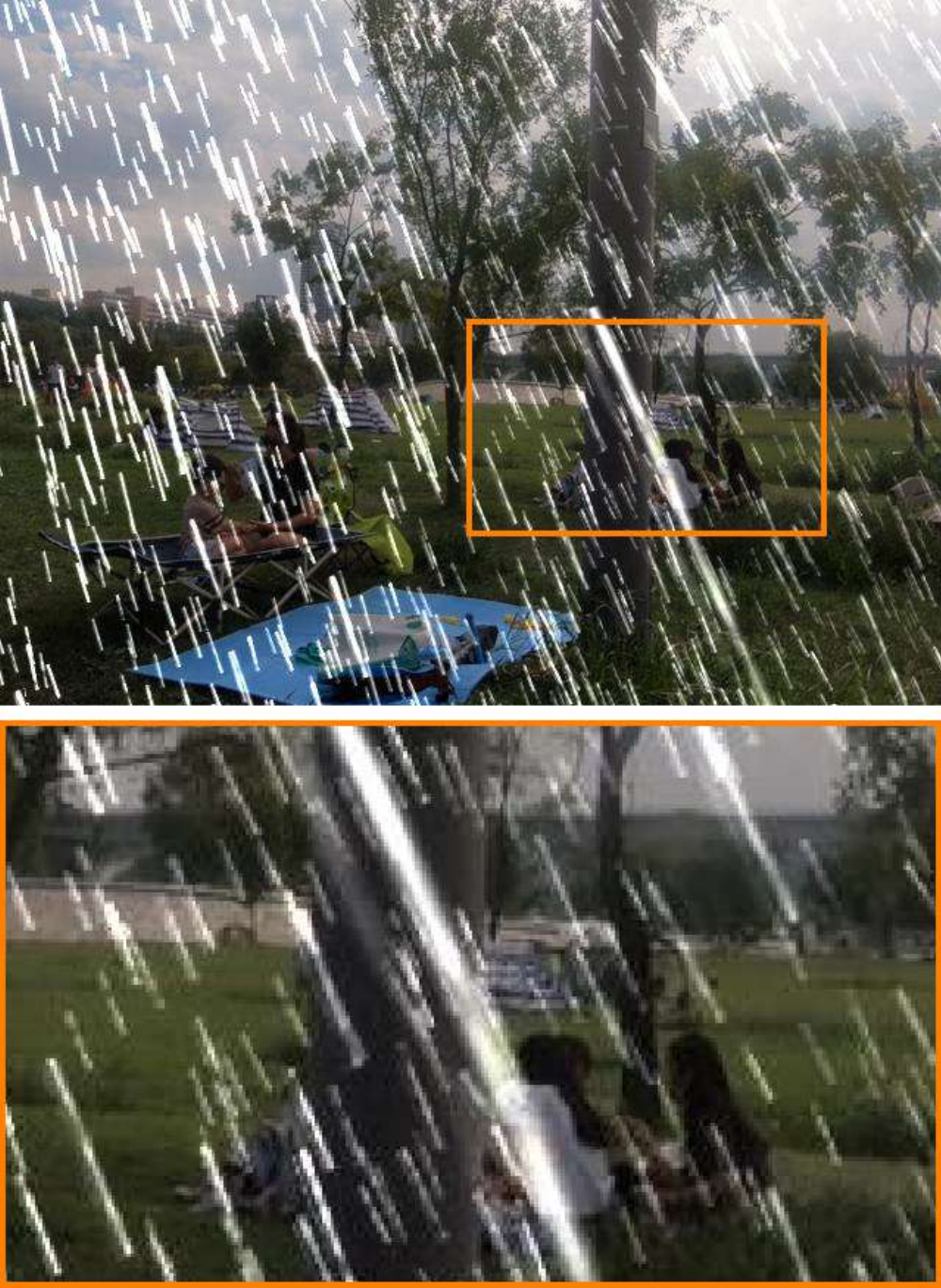}&
		\includegraphics[width=0.121\linewidth]{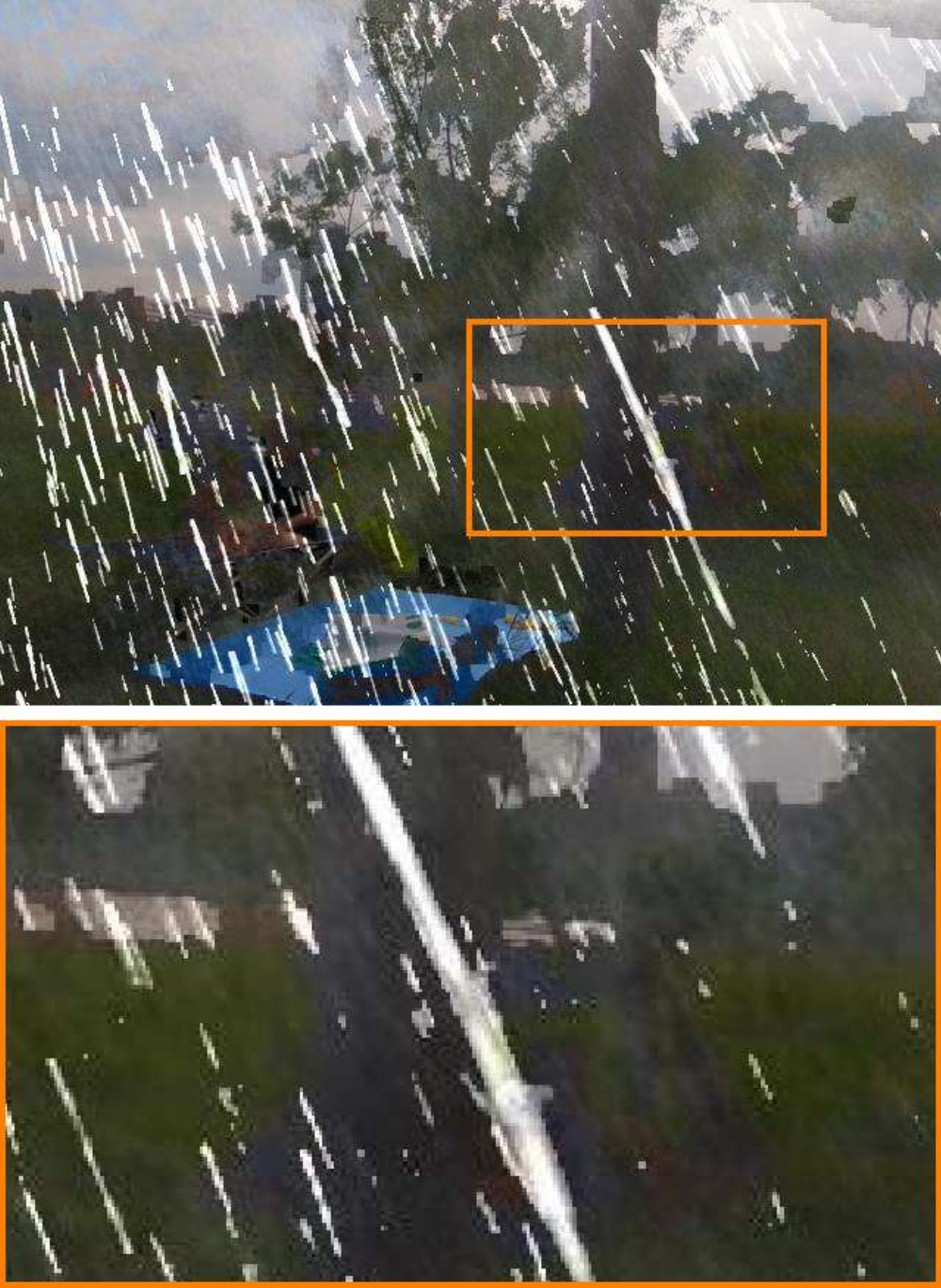}&
		\includegraphics[width=0.121\linewidth]{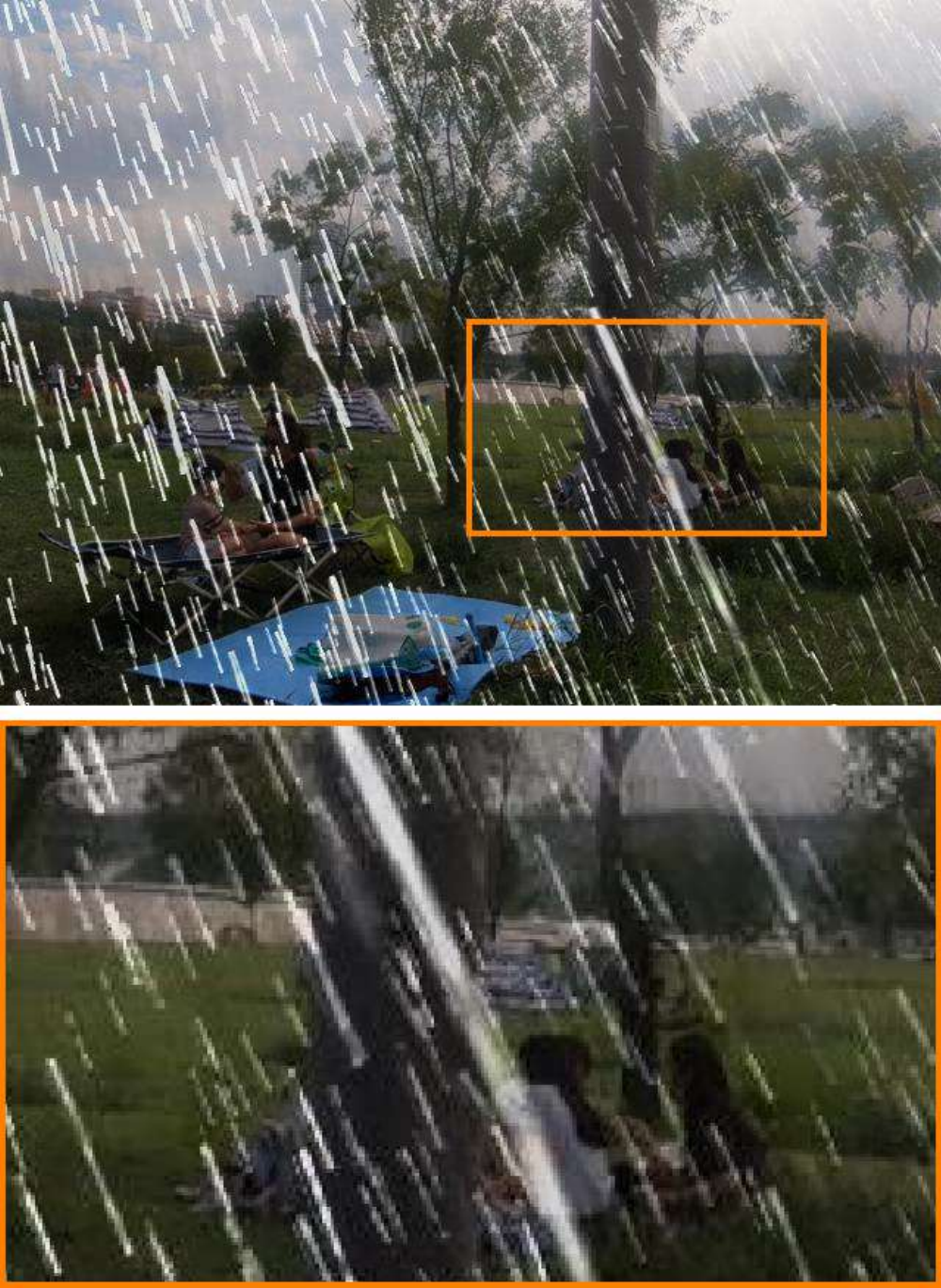}&
		\includegraphics[width=0.121\linewidth]{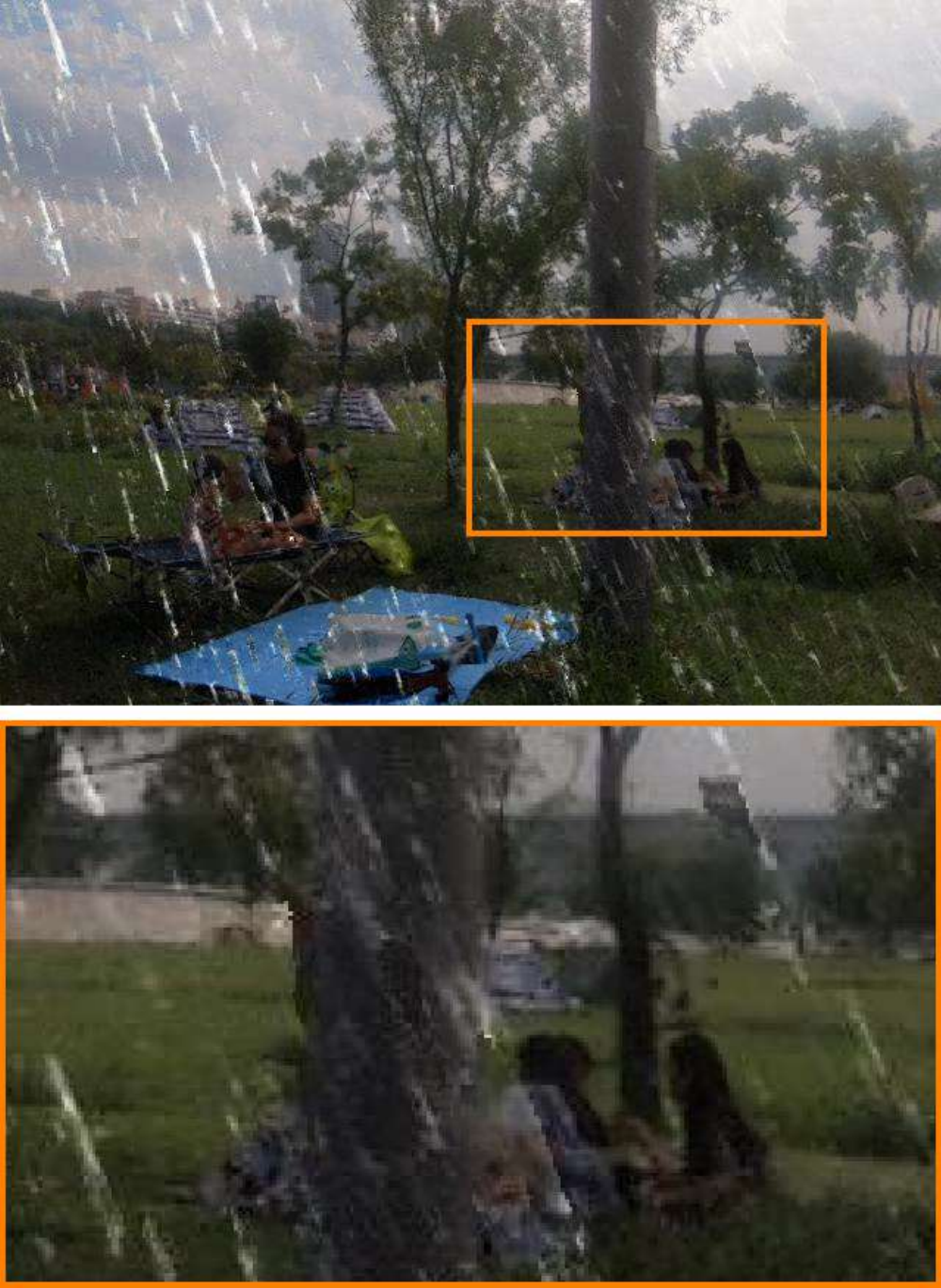}&
		\includegraphics[width=0.121\linewidth]{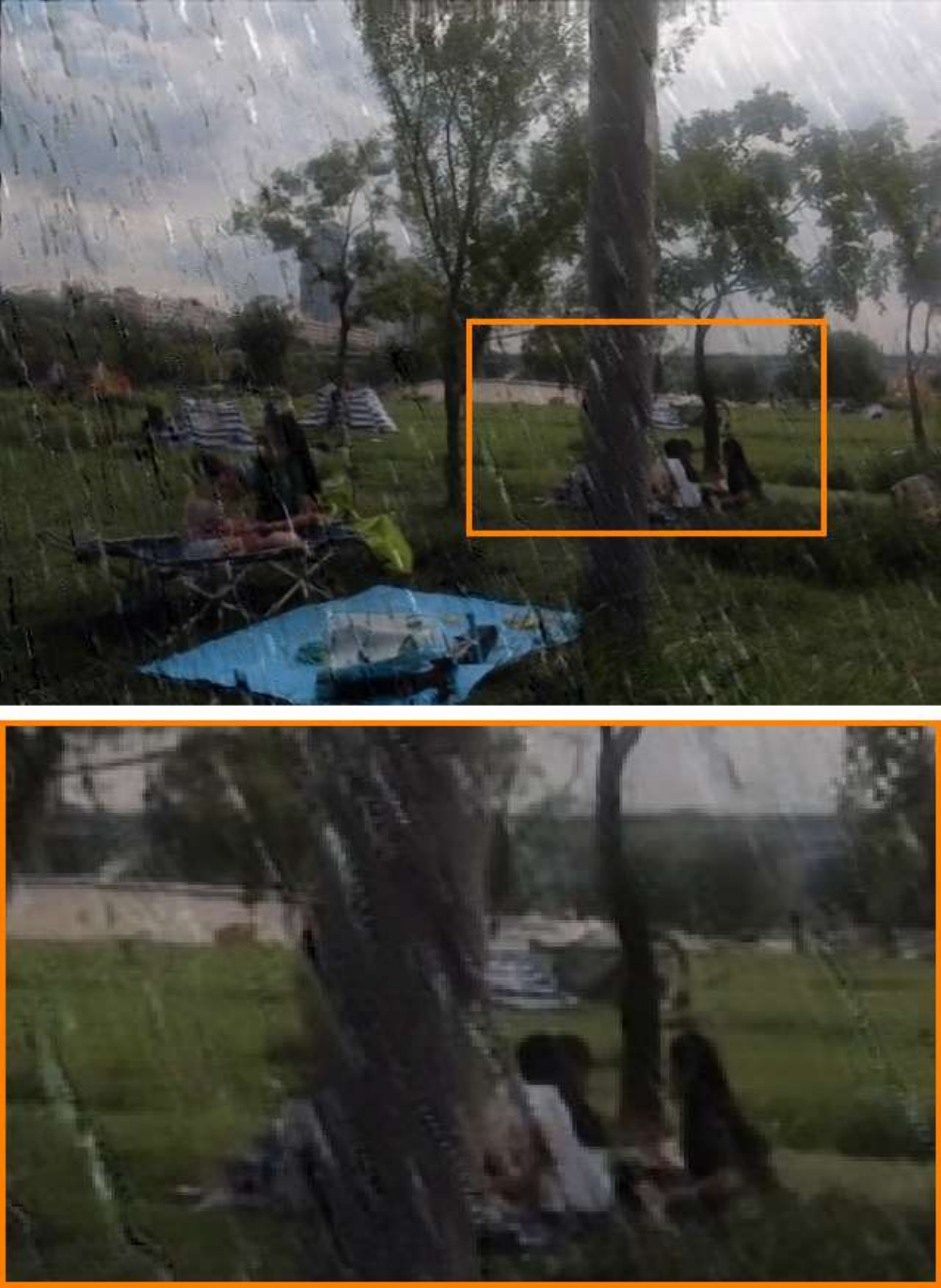}&
		\includegraphics[width=0.121\linewidth]{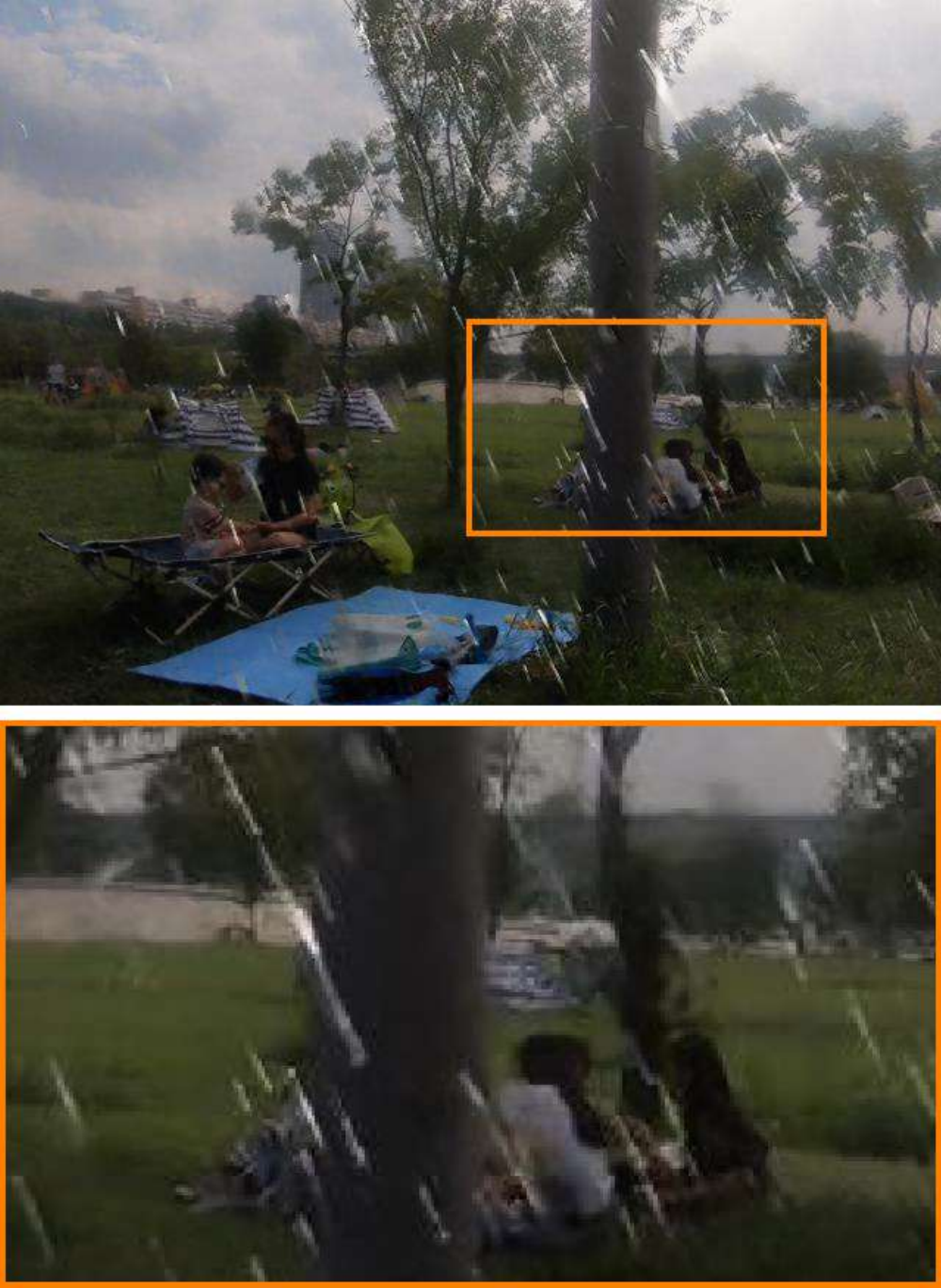}&
		\includegraphics[width=0.121\linewidth]{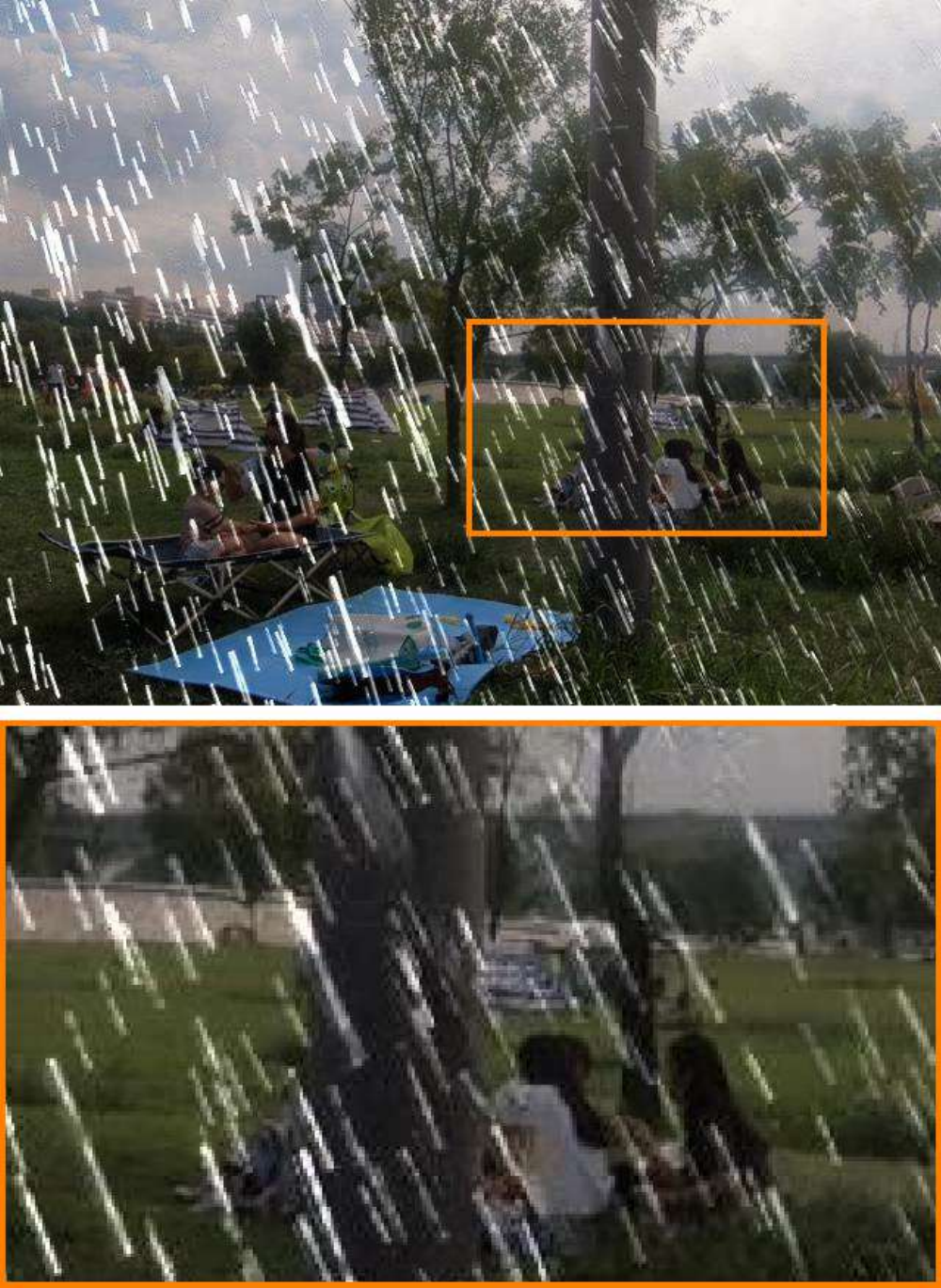}&
		\includegraphics[width=0.121\linewidth]{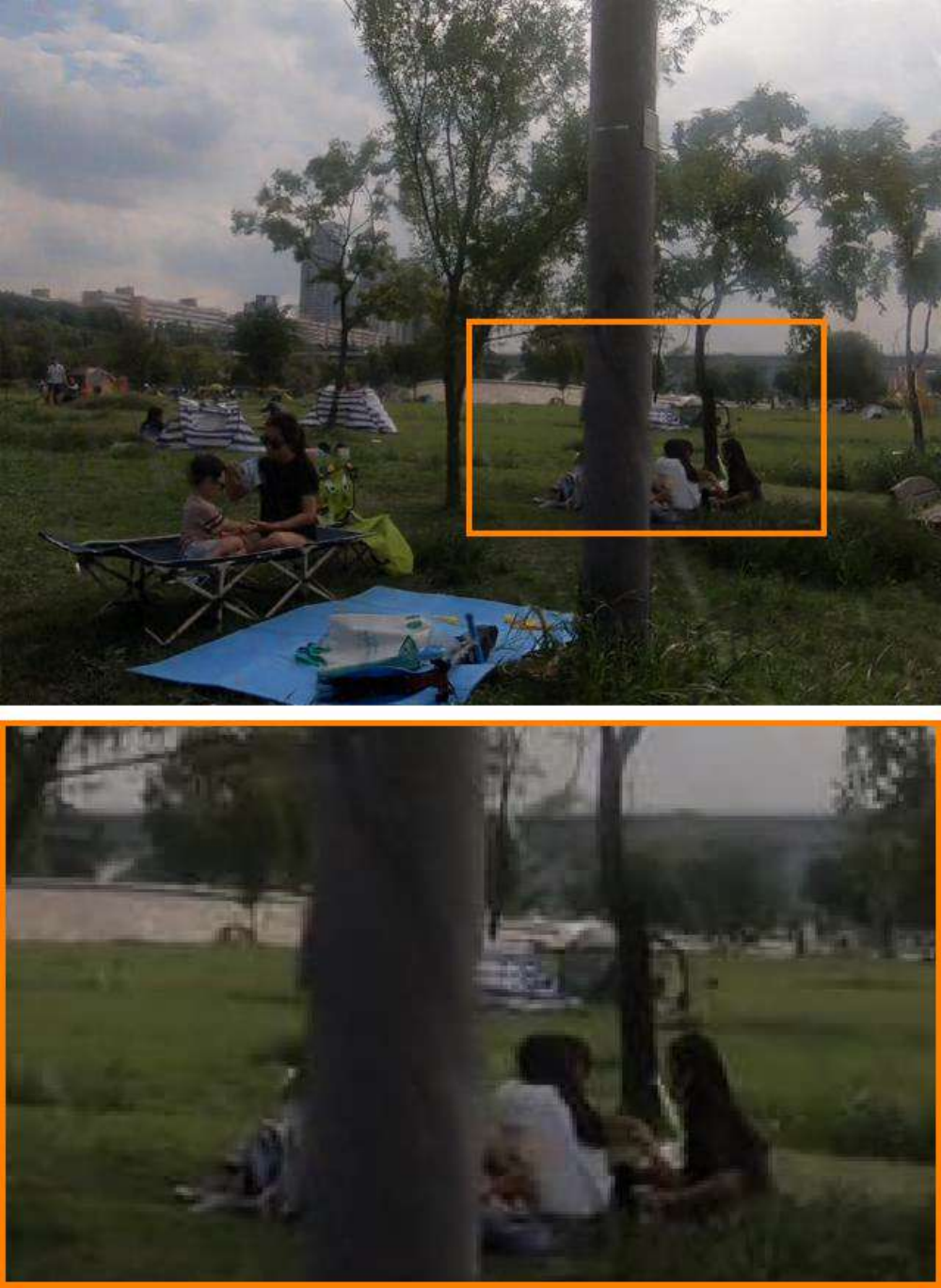}\\
		\includegraphics[width=0.121\linewidth]{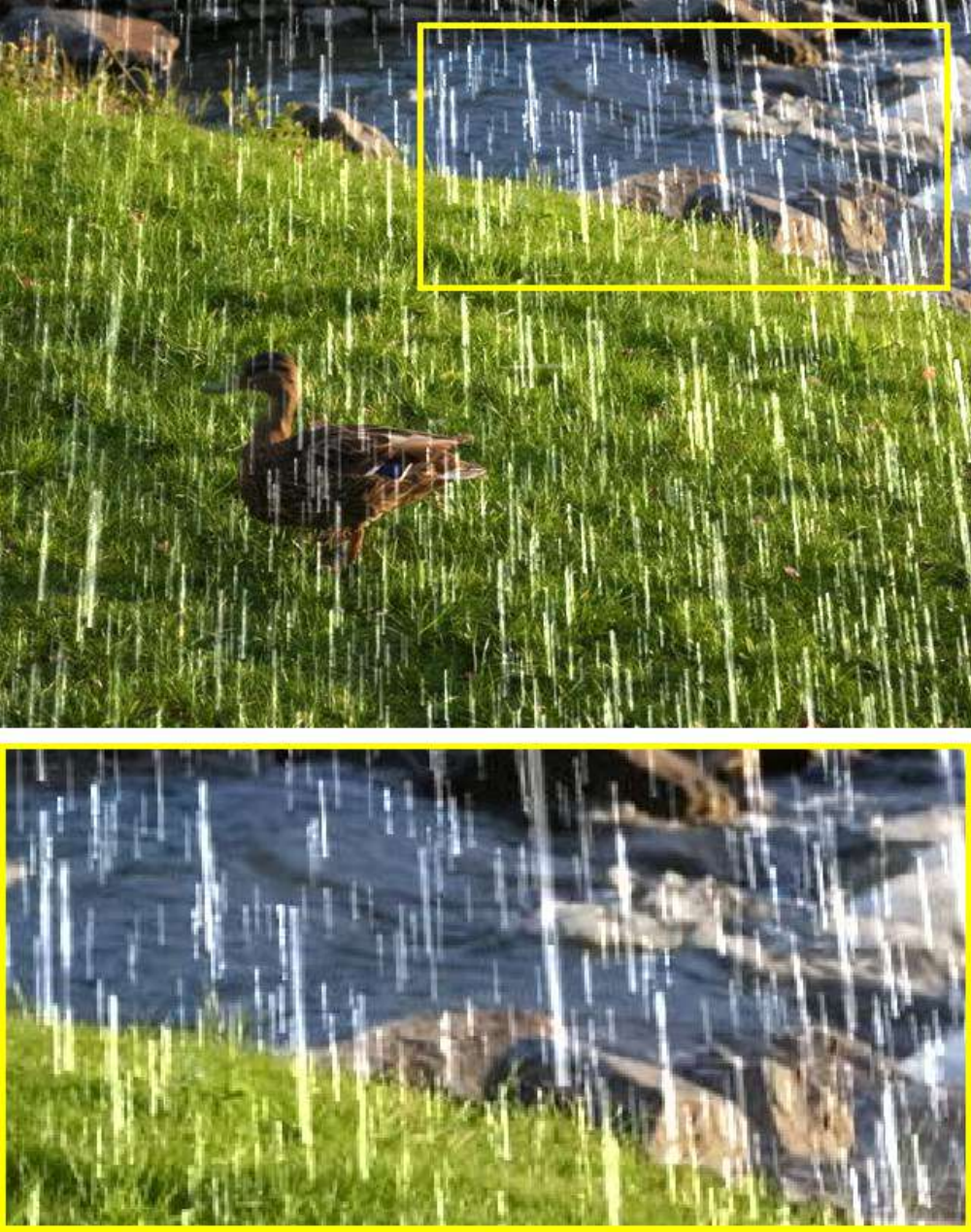}&
		\includegraphics[width=0.121\linewidth]{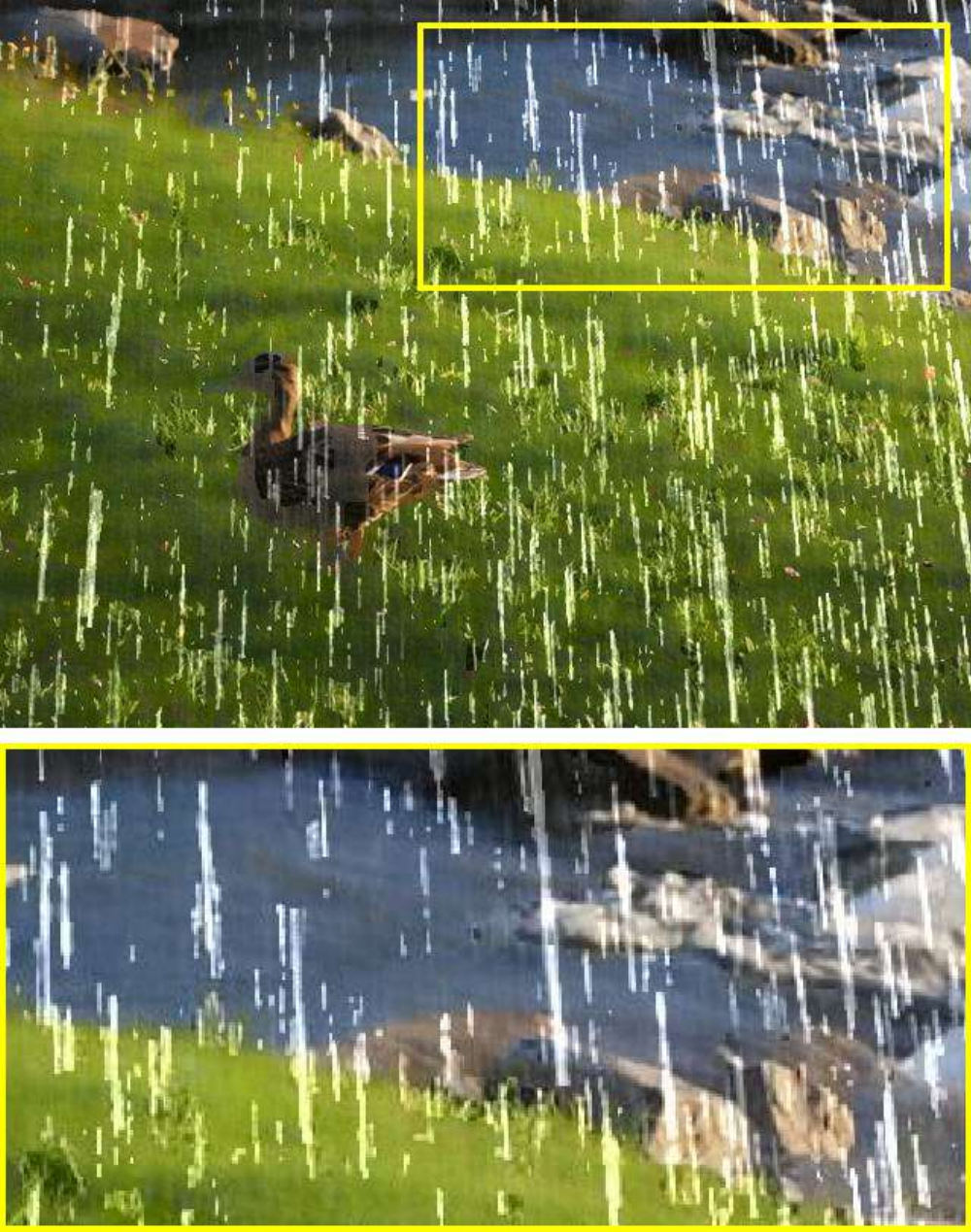}&
		\includegraphics[width=0.121\linewidth]{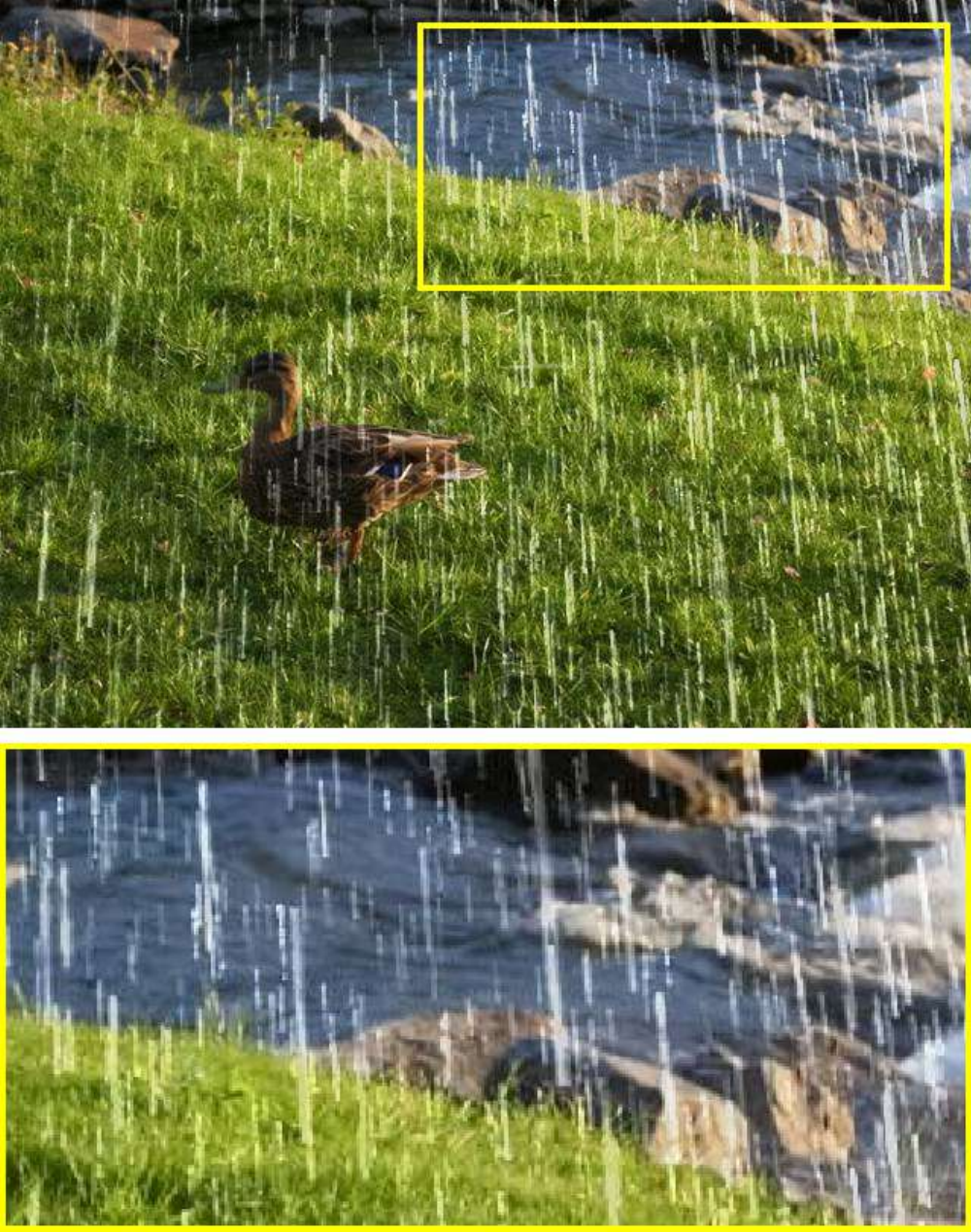}&
		\includegraphics[width=0.121\linewidth]{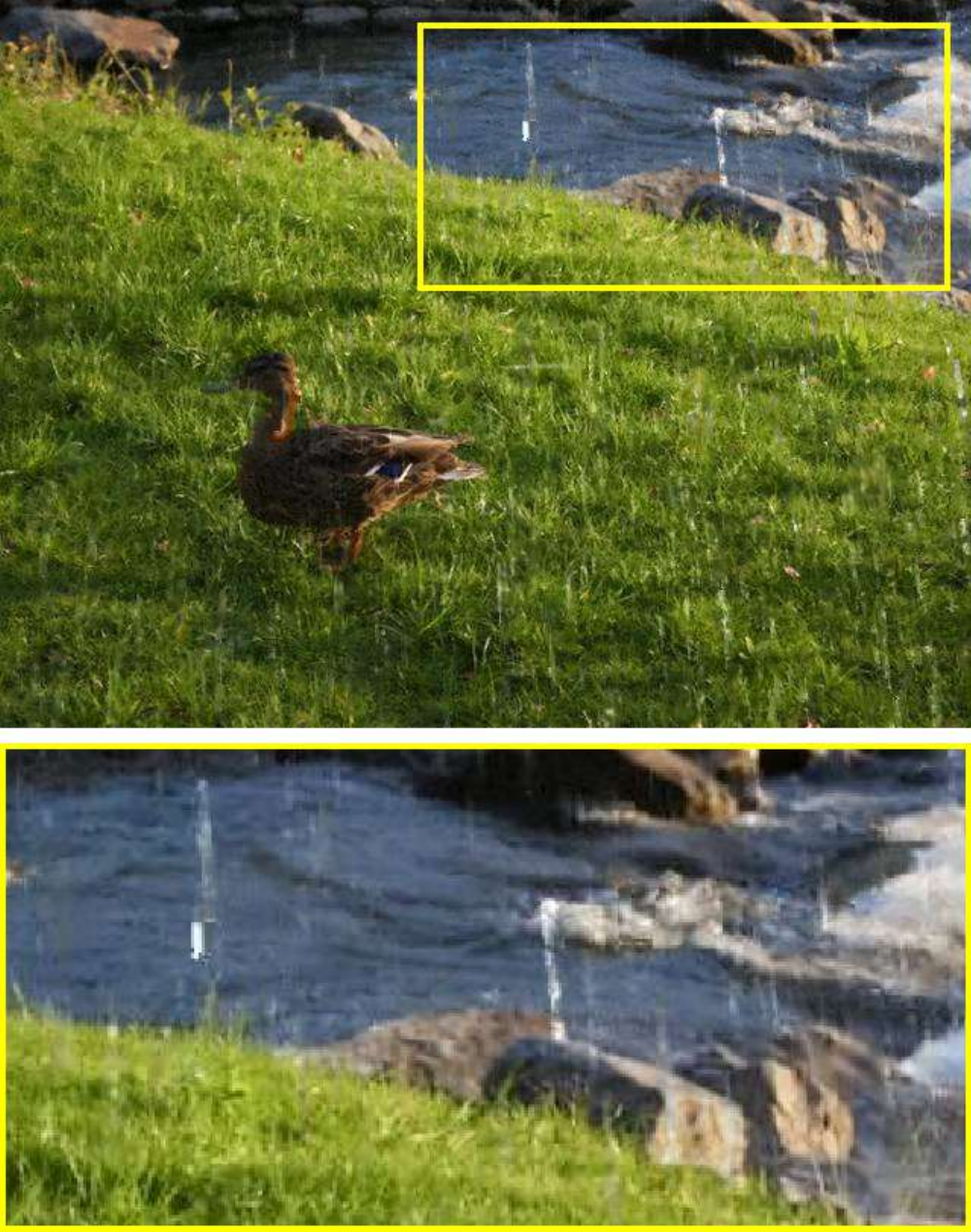}&
		\includegraphics[width=0.121\linewidth]{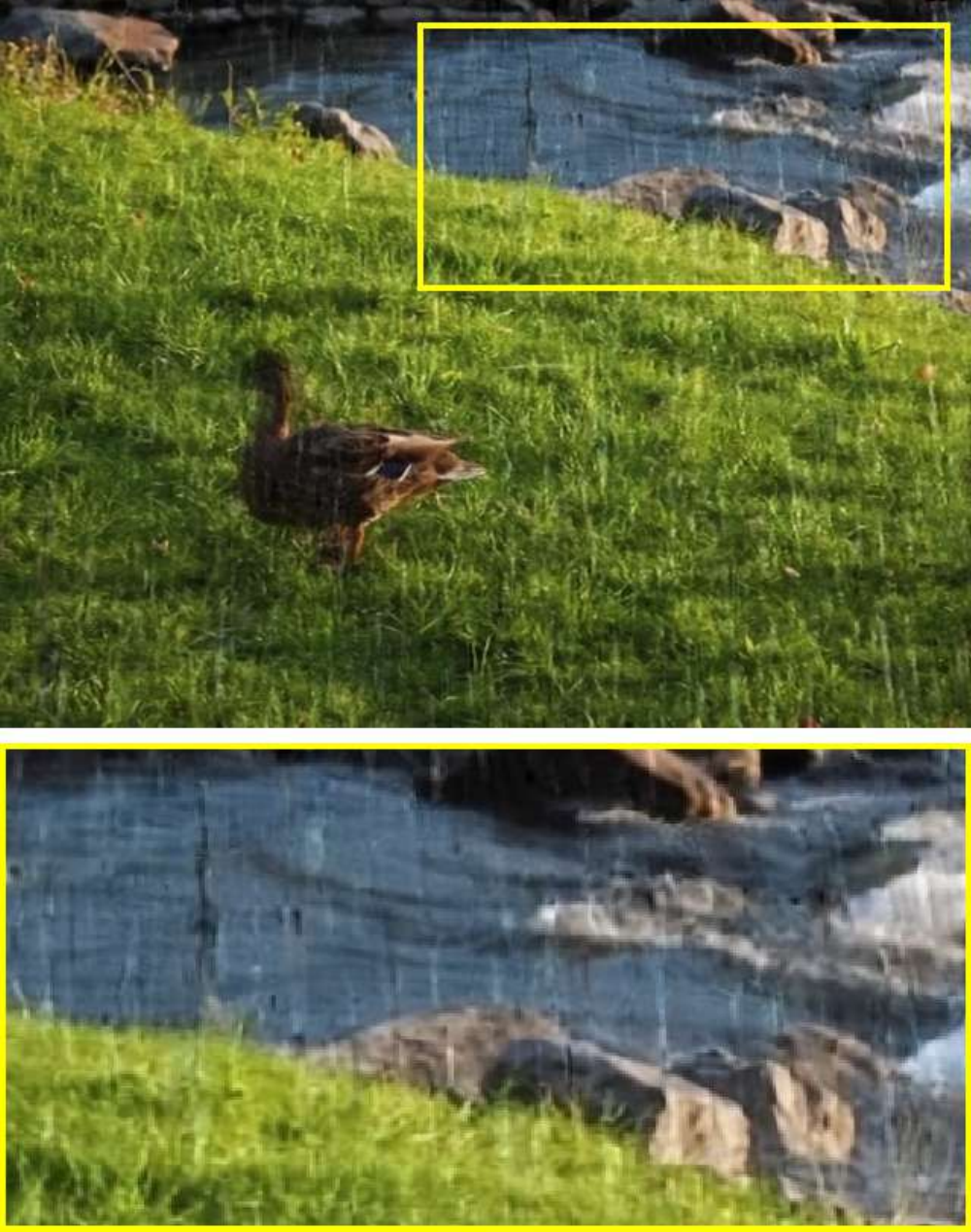}&
		\includegraphics[width=0.121\linewidth]{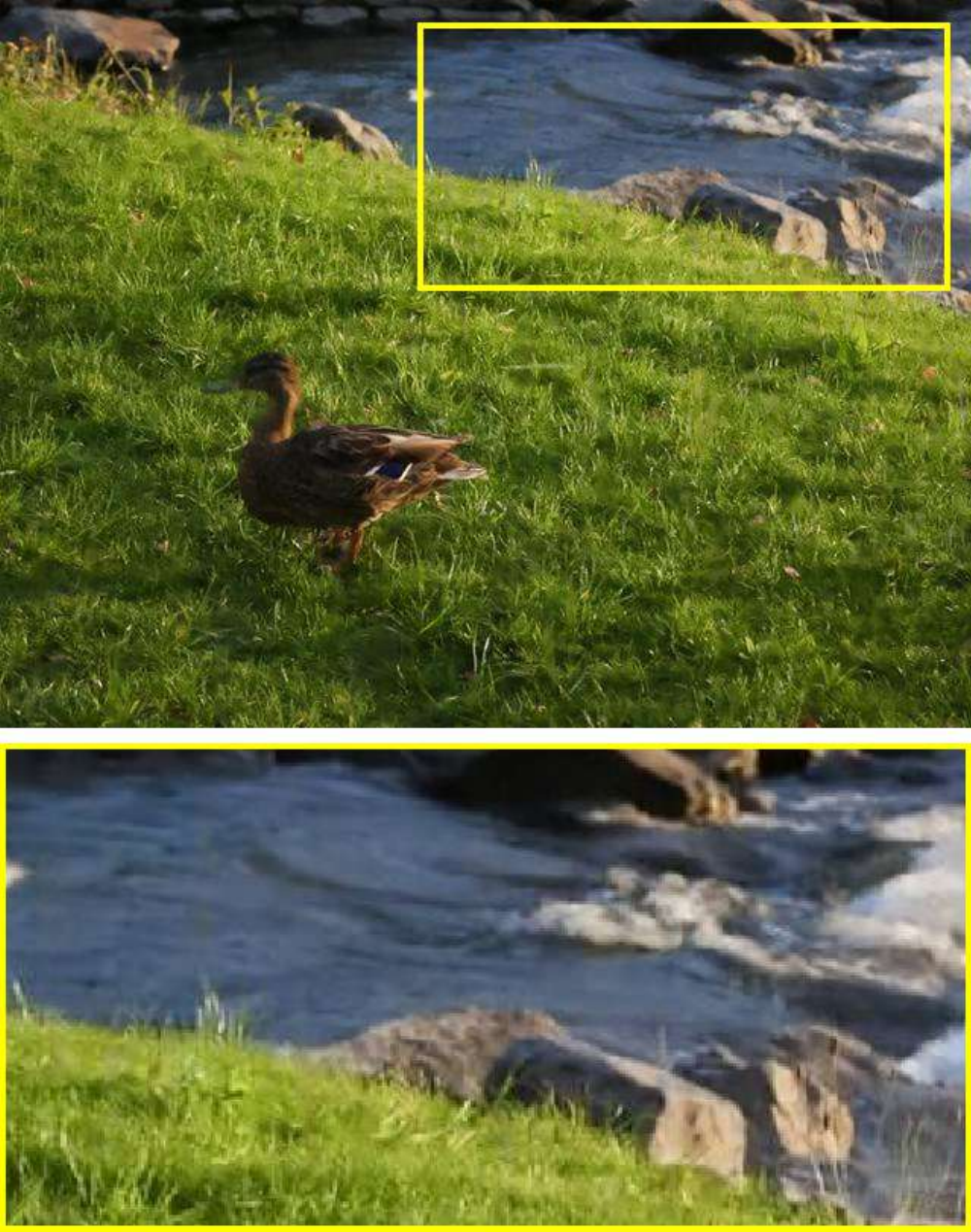}&
		\includegraphics[width=0.121\linewidth]{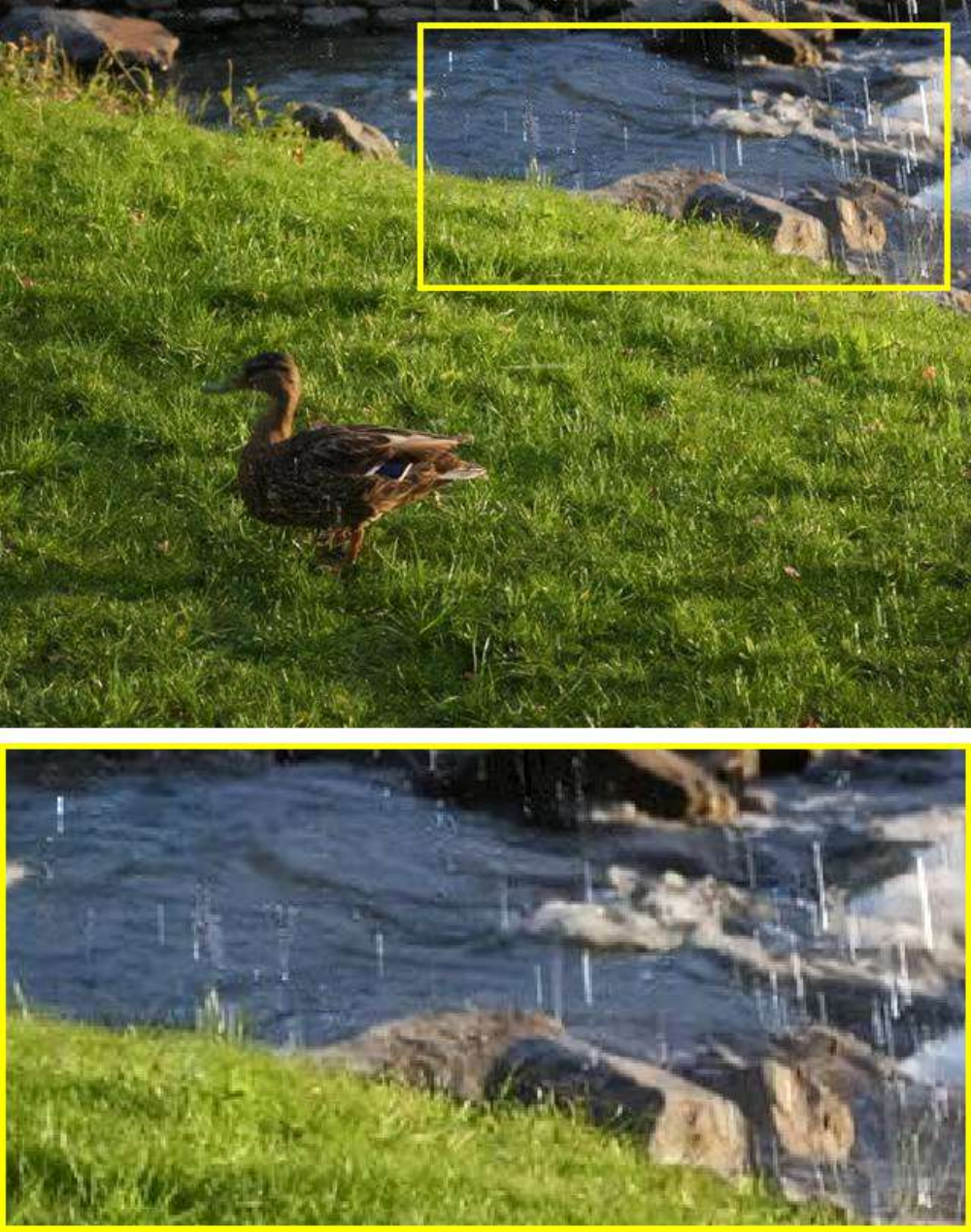}&
		\includegraphics[width=0.121\linewidth]{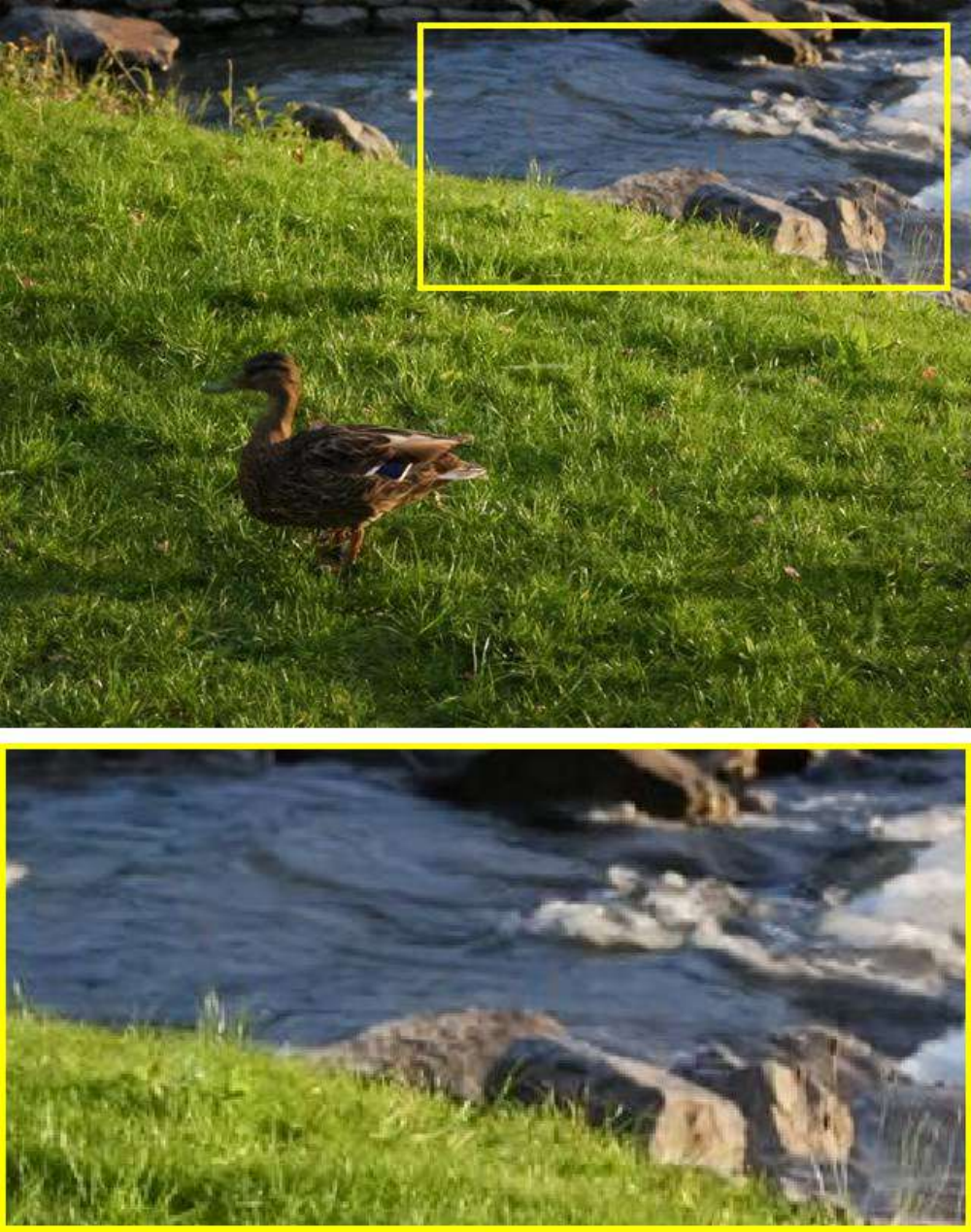}\\	

		\footnotesize{Input}&\footnotesize MSCSC&\footnotesize{FastDerain}&\footnotesize SpacCNN&\footnotesize SLDNet&\footnotesize S2VD &\footnotesize{RDD} &\footnotesize Ours\\
	\end{tabular}	
	\caption{Visual comparisons of rain removal effects on our proposed LARA dataset with different methods. Each row represents a different scene variation. Our method achieves the best visual results.}
	\label{fig:fig_ours}
	\vspace{-0.3cm}
\end{figure*}	

\begin{figure*}[!t]
	\centering
	\begin{tabular}{c@{\extracolsep{0.5em}}c@{\extracolsep{0.5em}}c@{\extracolsep{0.5em}}c}
		\includegraphics[width=0.235\linewidth]{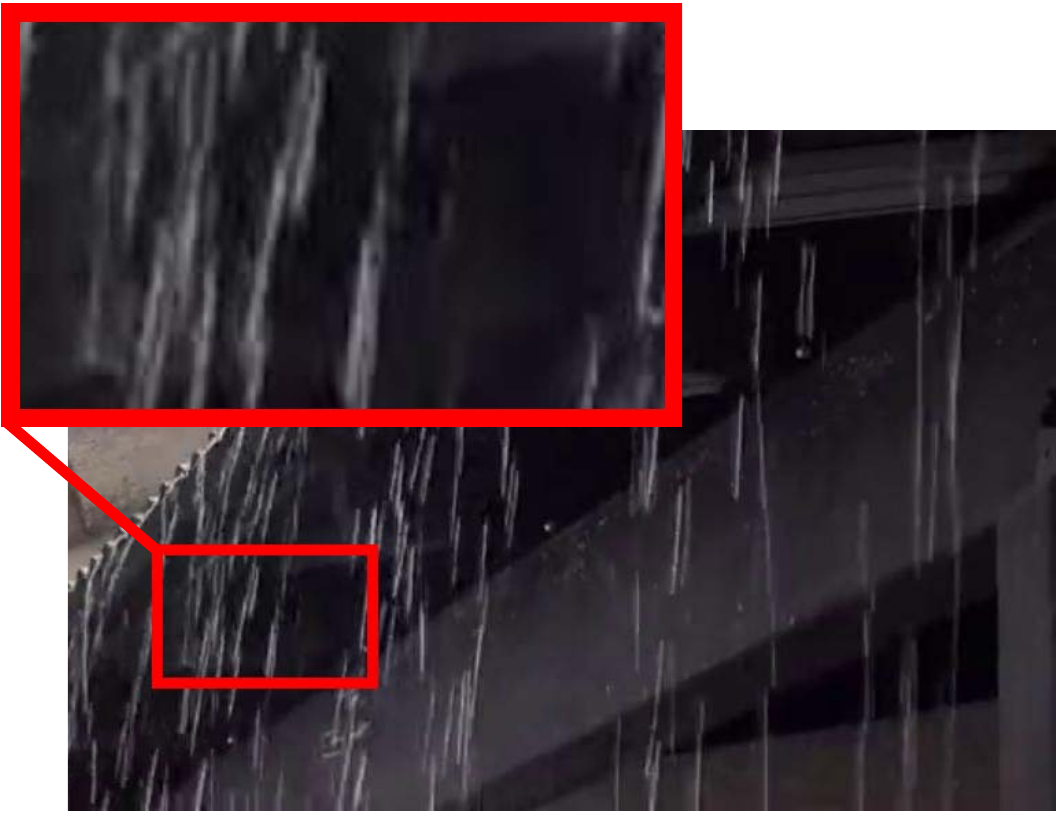}&
		\includegraphics[width=0.235\linewidth]{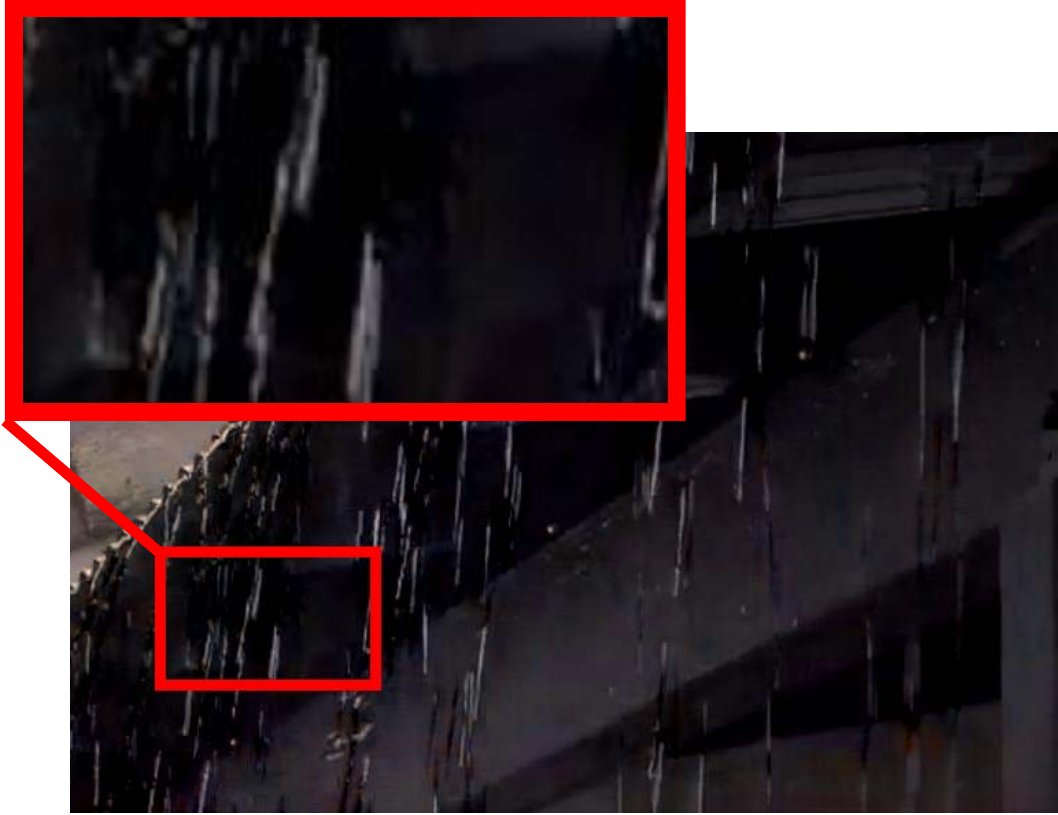}&
		\includegraphics[width=0.235\linewidth]{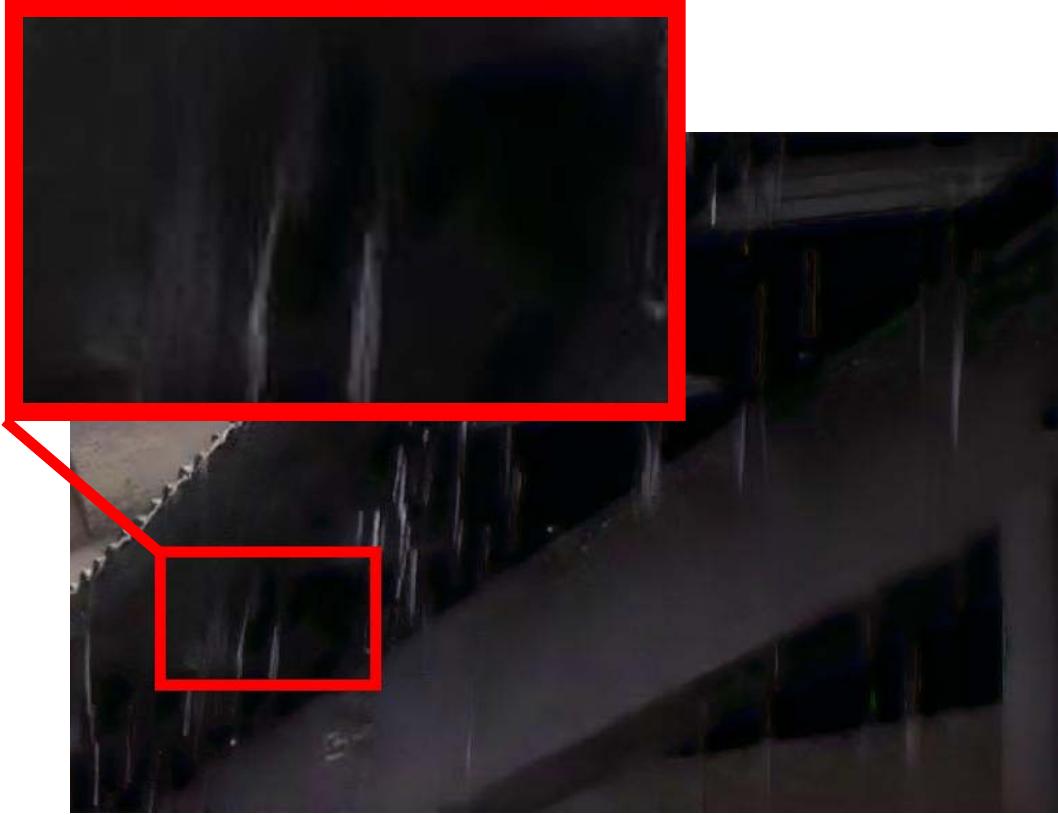}&
		\includegraphics[width=0.235\linewidth]{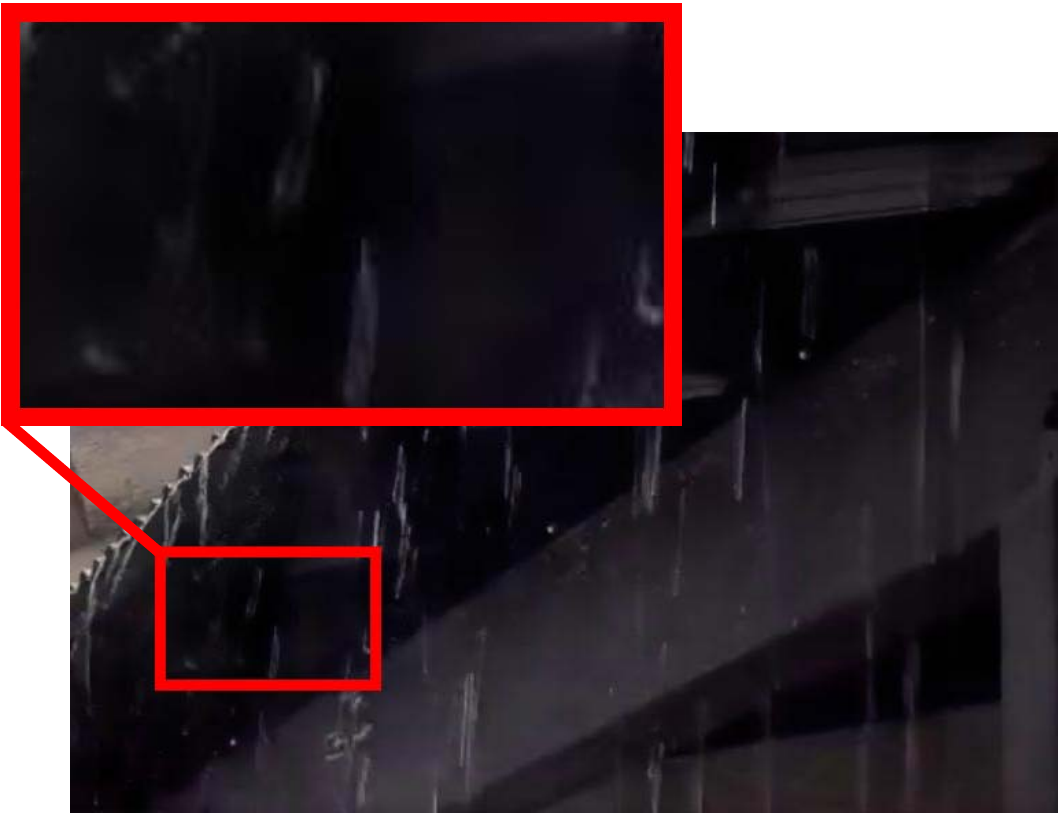}\\
		\footnotesize Input &\footnotesize{SIRR} &\footnotesize Syn2Real &\footnotesize MOSS \\	
		\includegraphics[width=0.235\linewidth]{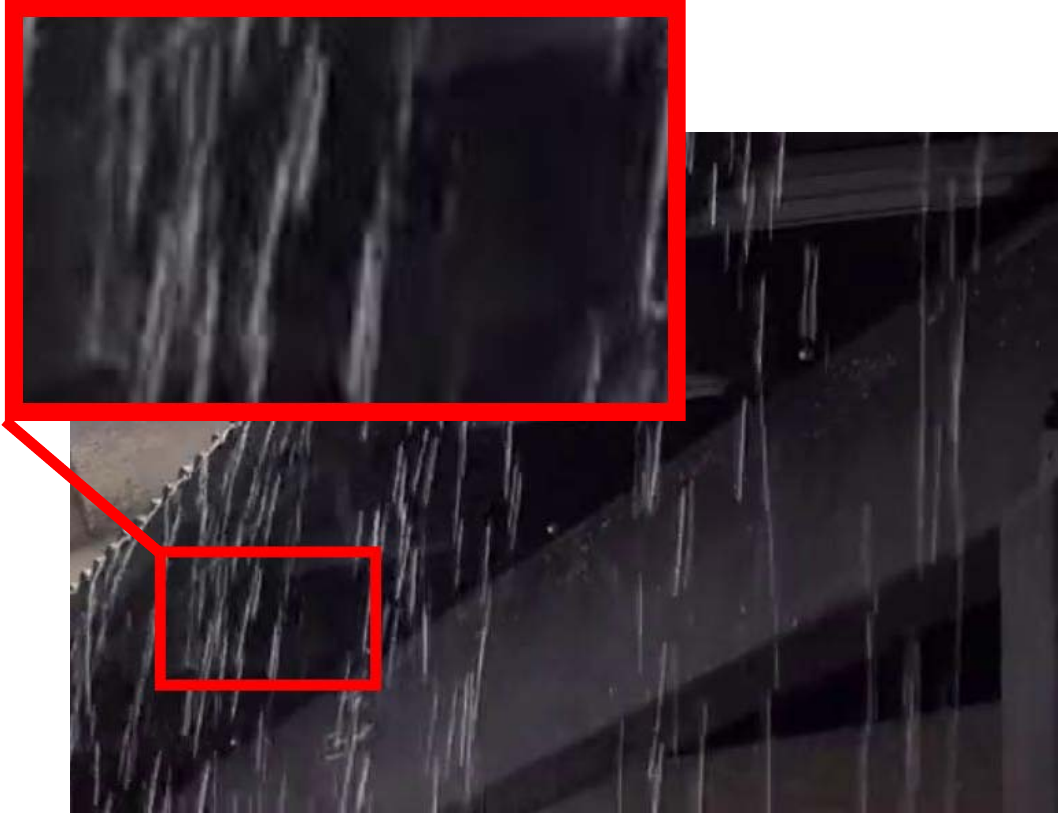}&
		\includegraphics[width=0.235\linewidth]{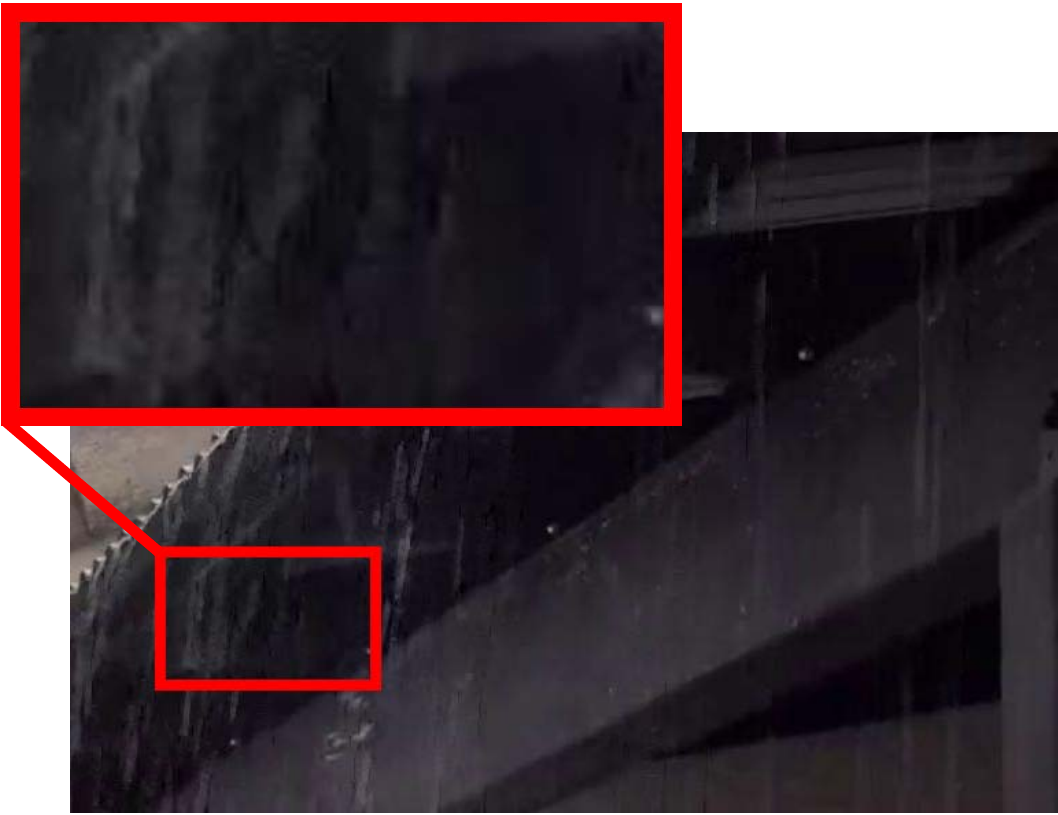}&
		\includegraphics[width=0.235\linewidth]{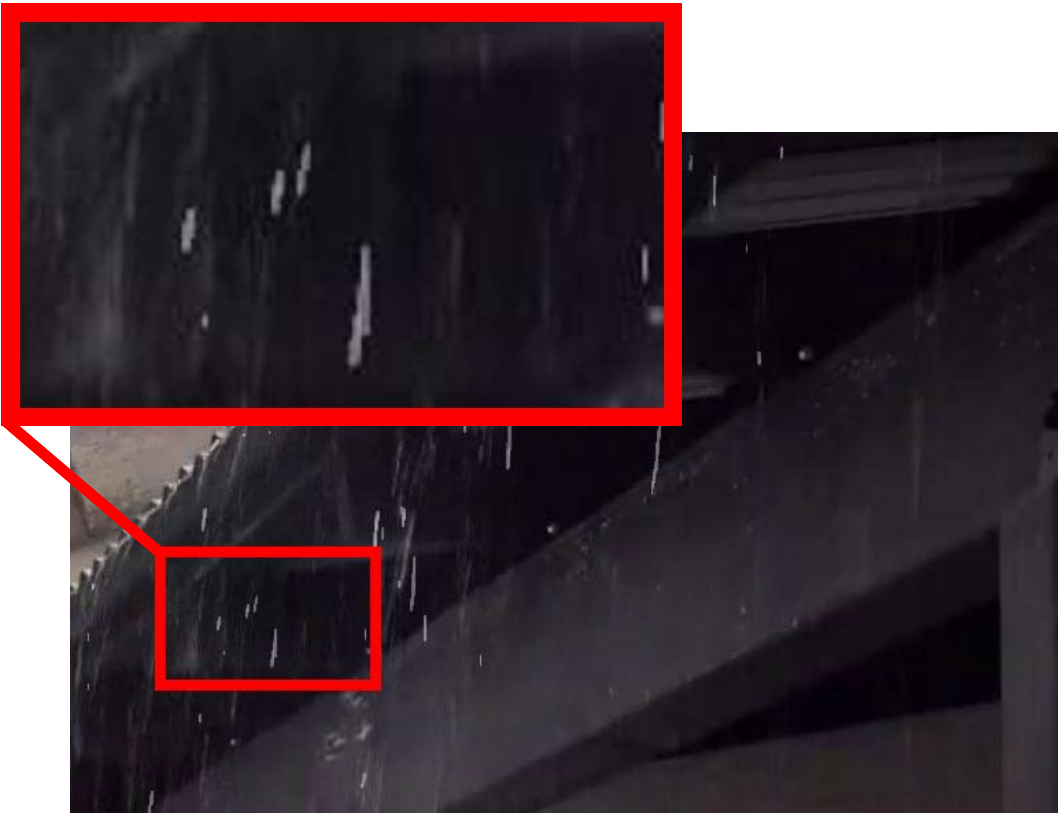} &
		\includegraphics[width=0.235\linewidth]{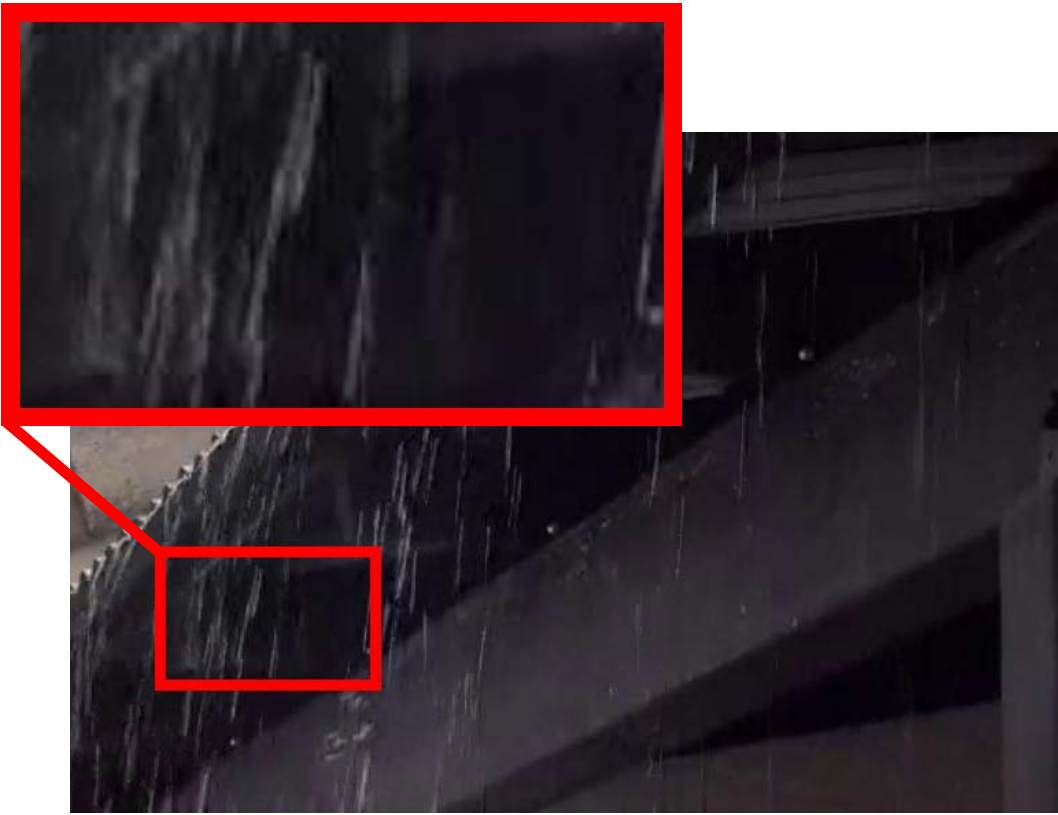}\\
		\footnotesize{SE} &\footnotesize DIP&\footnotesize MSCSC &\footnotesize FastDerain\\	
		\includegraphics[width=0.235\linewidth]{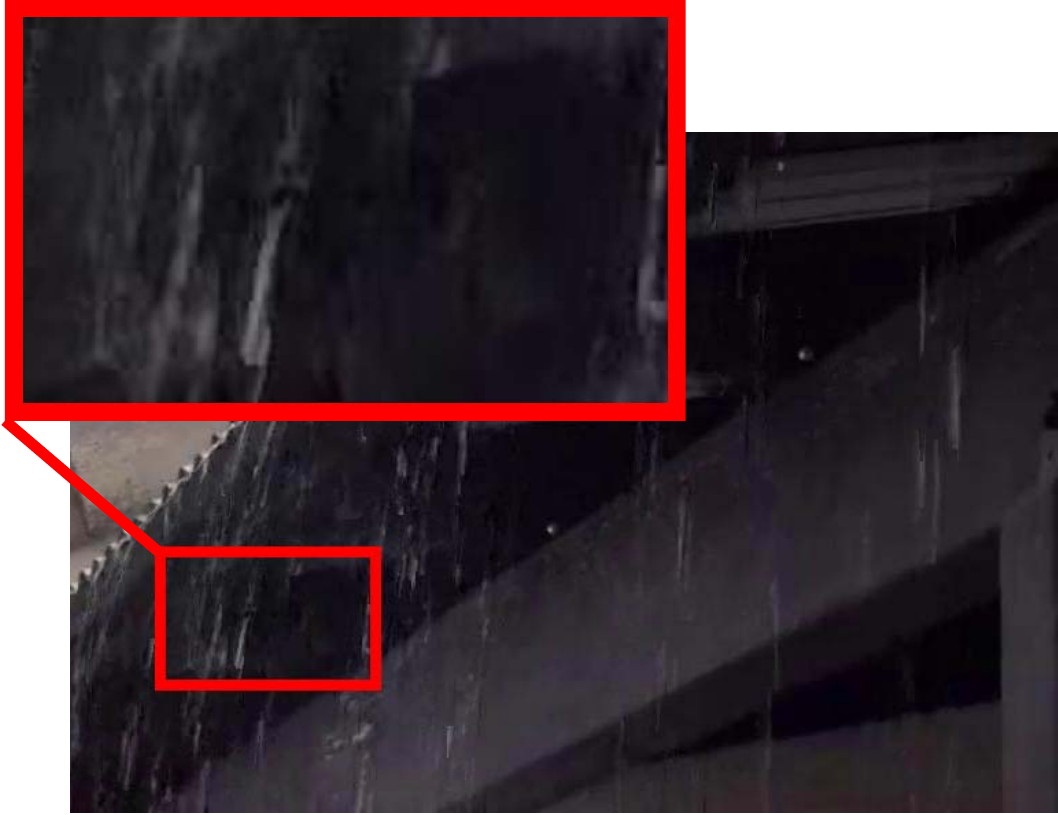}&
		\includegraphics[width=0.235\linewidth]{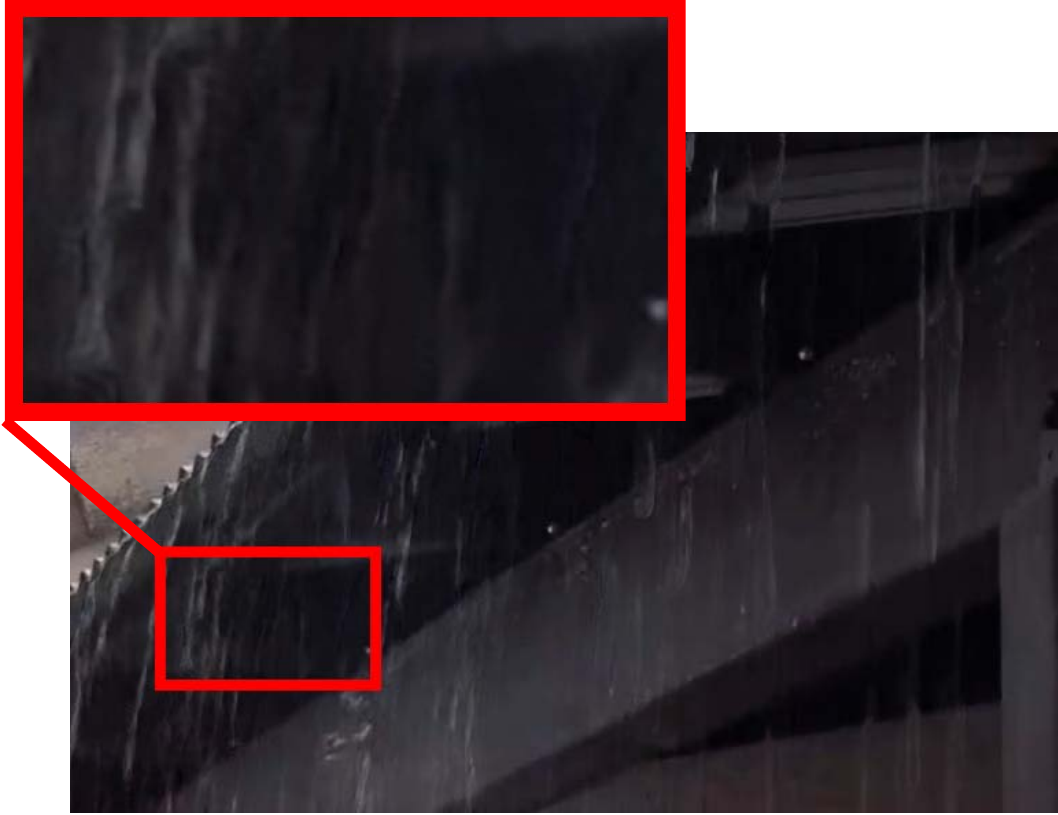}&
		\includegraphics[width=0.235\linewidth]{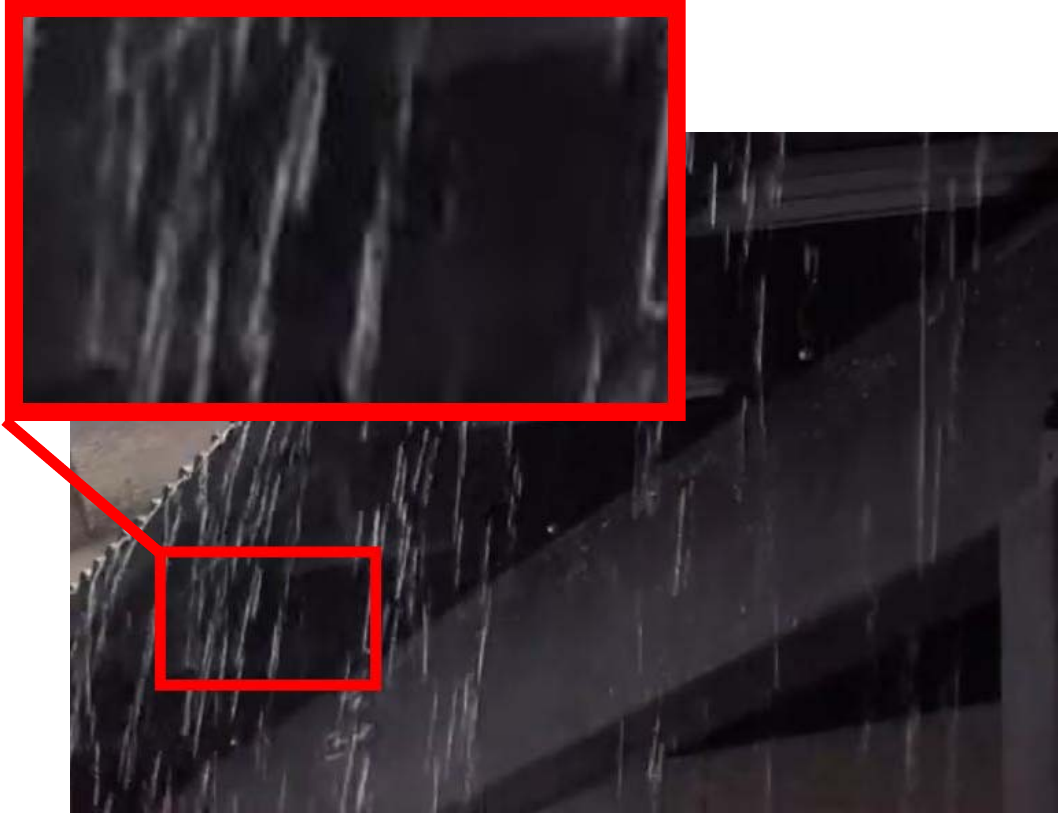}&
		\includegraphics[width=0.235\linewidth]{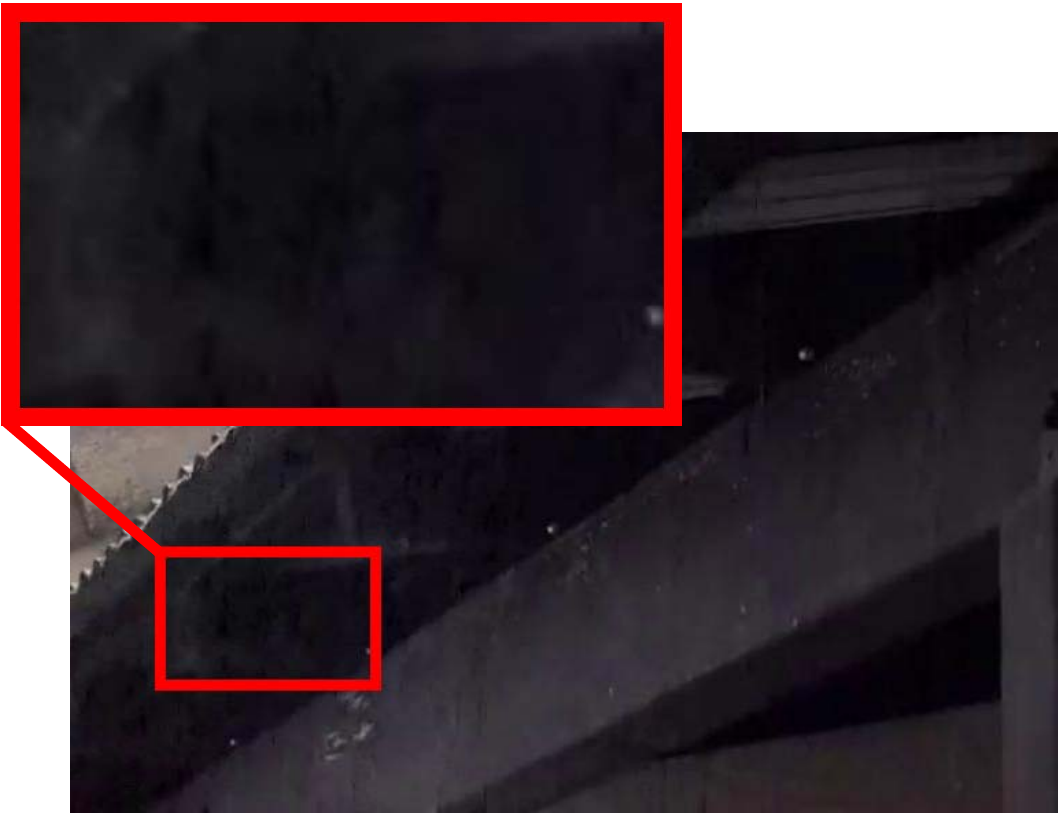}\\
		\footnotesize{SpacCNN}&\footnotesize SLDNet&\footnotesize S2VD&\footnotesize{Ours}\\	
     \end{tabular}
	\caption{Visual comparison among different video deraining methods on a real-world example. Zoom in for better view of the rain removal effect. For the difference between real and synthetic rain streaks, the analysis of the results in real scenarios better demonstrates the superiority of our proposed network and learning strategy.}
	\label{fig:fig_real1}
	\vspace{-0.5cm}
\end{figure*}

\subsection{Quantitative Evaluation}
We utilize PSNR and SSIM metrics to evaluate fidelity quality of the results. We compare our method with the single image and video deraining methods mentioned above in~\cref{tab:value_our}. The proposed method shows significant superiority to previous methods. Compared to latest video deraining approaches, such as MS-CSC, FastDerain, J4RNet, SpacCNN, SLDNet, ESTI, RDD and S2VD, the performance gain is up to about 14dB, 11dB, 7dB, 6dB, 7dB, 7dB, 7dB, 4dB in PSNR and about 0.38, 0.25, 0.11, 0.12, 0.18, 0.14, 0.13, 0.06 in SSIM. We also compare these methods on three commonly-used benchmarks~\cite{liu2018erase,chen2018robust}. As shown in~\cref{tab:value_LH}, our results are consistently much better than previous methods on all three datasets, which shows the superiority of our method. Training on our synthesized dataset and utilizing the proposed learning scene adaptation strategy help us get these amazing results. The analysis of our synthesized LARA dataset and learning strategy will be shown in~\cref{ssec:dataset_study} and~\cref{ssec:strategy_study}. It should be mentioned that we retrained SLDNet on all datasets and retrained the S2VD model with the released code on all datasets except \emph{NTURain} and test S2VD on \emph{NTURain} with the released model parameters\footnote{https://github.com/zsyOAOA/S2VD/tree/master/model\_states}. 

Besides the fidelity quality, we also evaluate perceptual quality with LPIPS and NIQE on our synthesized dataset. As shown in~\cref{tab:value_our}, our method obtains better LPIPS and NIQE scores (the fewer scores, the better perceptual quality) compared with other state-of-the-art approaches, which means our approach can obtain better perceptual quality.

To evaluate temporal quality, we utilize the tLP metric to assess the temporal consistency. As shown in~\cref{tab:value_our}, our method obtains the best performance. Besides, learning-based video rain removal methods generally outperform those single image rain removal methods on tLP metric, which shows the importance of temporal modeling. 

\begin{figure*}[!t]
	\centering
	\begin{tabular}{c@{\extracolsep{0.3em}}c@{\extracolsep{0.3em}}c@{\extracolsep{0.3em}}c@{\extracolsep{0.3em}}c@{\extracolsep{0.3em}}c@{\extracolsep{0.3em}}c}		
		\rotatebox	{90}{$\;\;\;\;\,$\footnotesize Input}&	
		\includegraphics[height=0.088\linewidth]{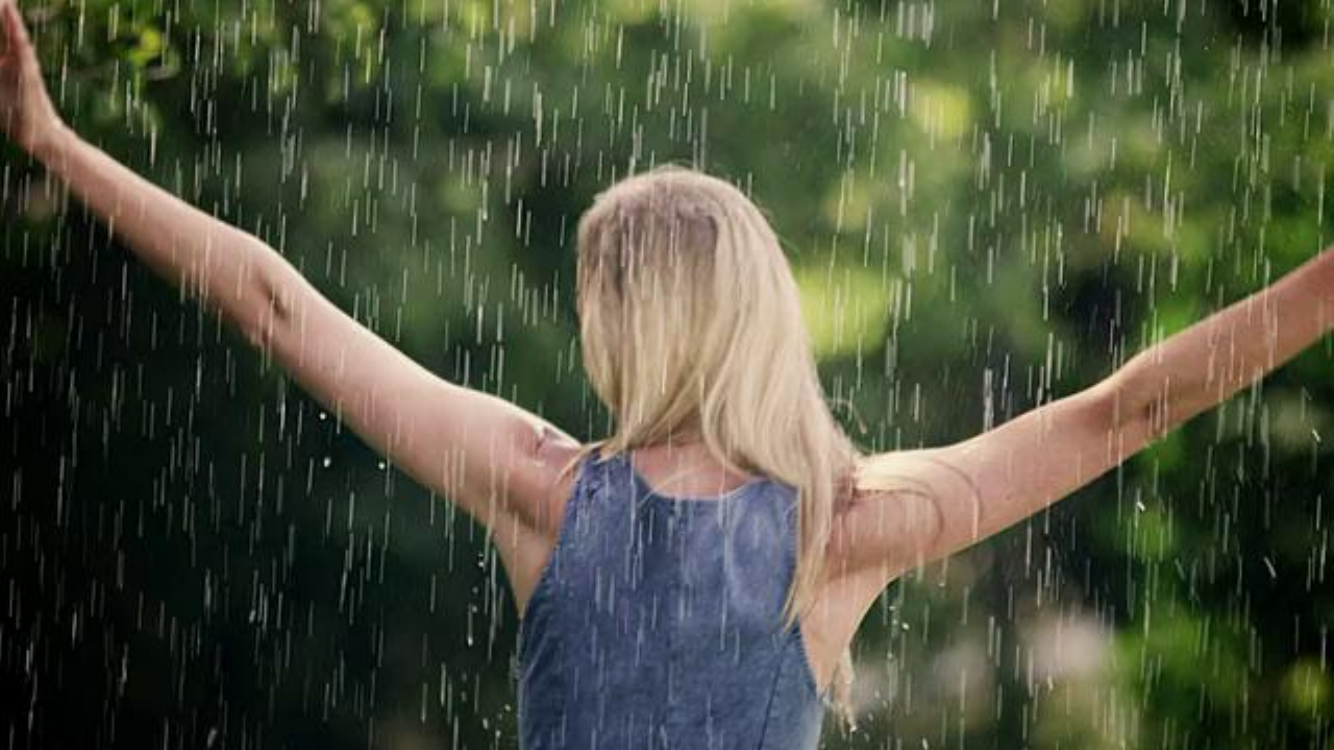}&
		\includegraphics[height=0.088\linewidth]{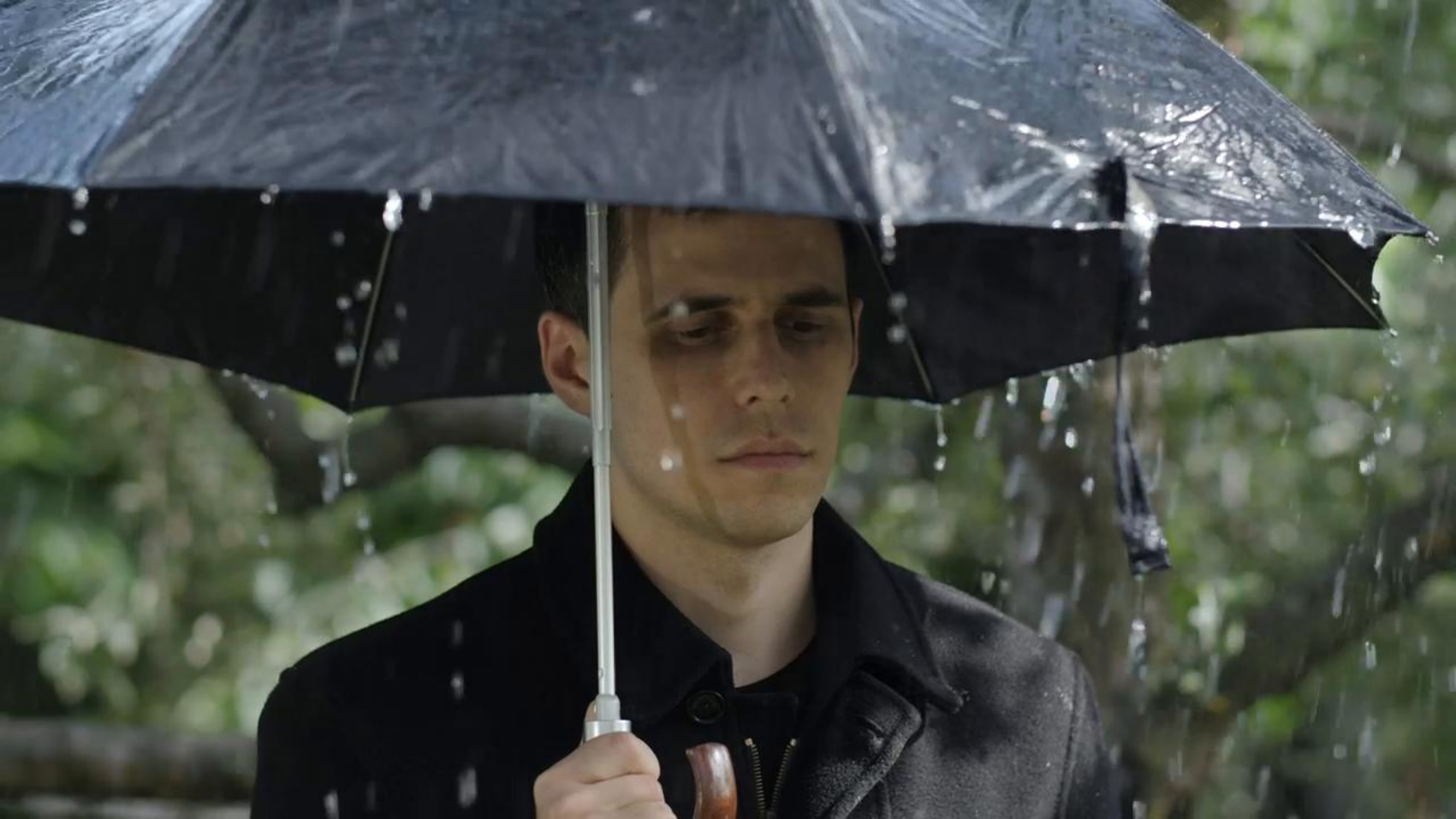}&
		\includegraphics[height=0.088\linewidth]{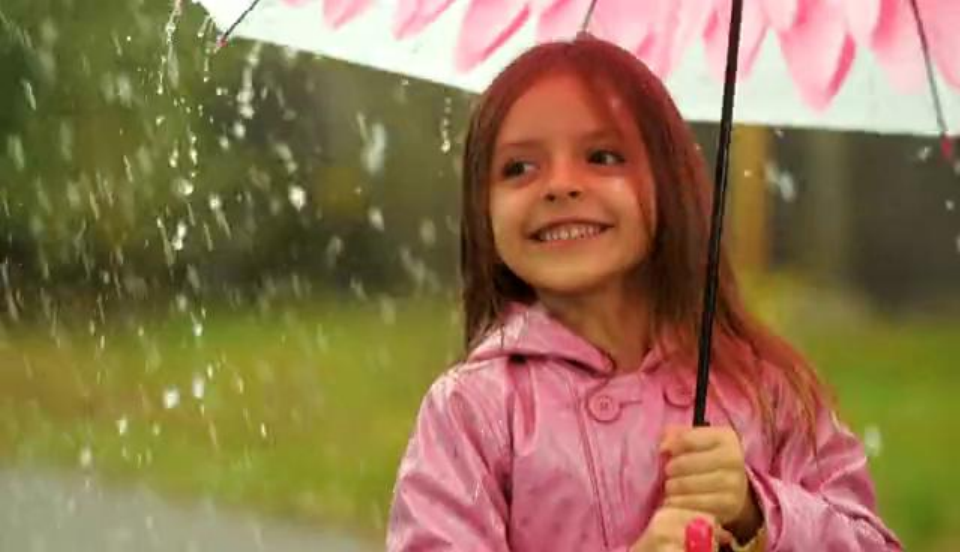}&
		\includegraphics[height=0.088\linewidth]{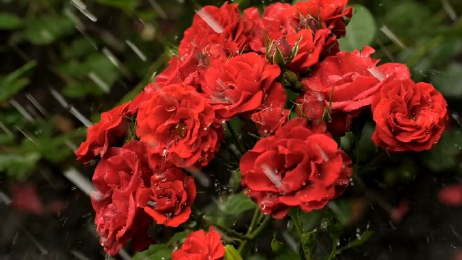}&
		\includegraphics[height=0.088\linewidth]{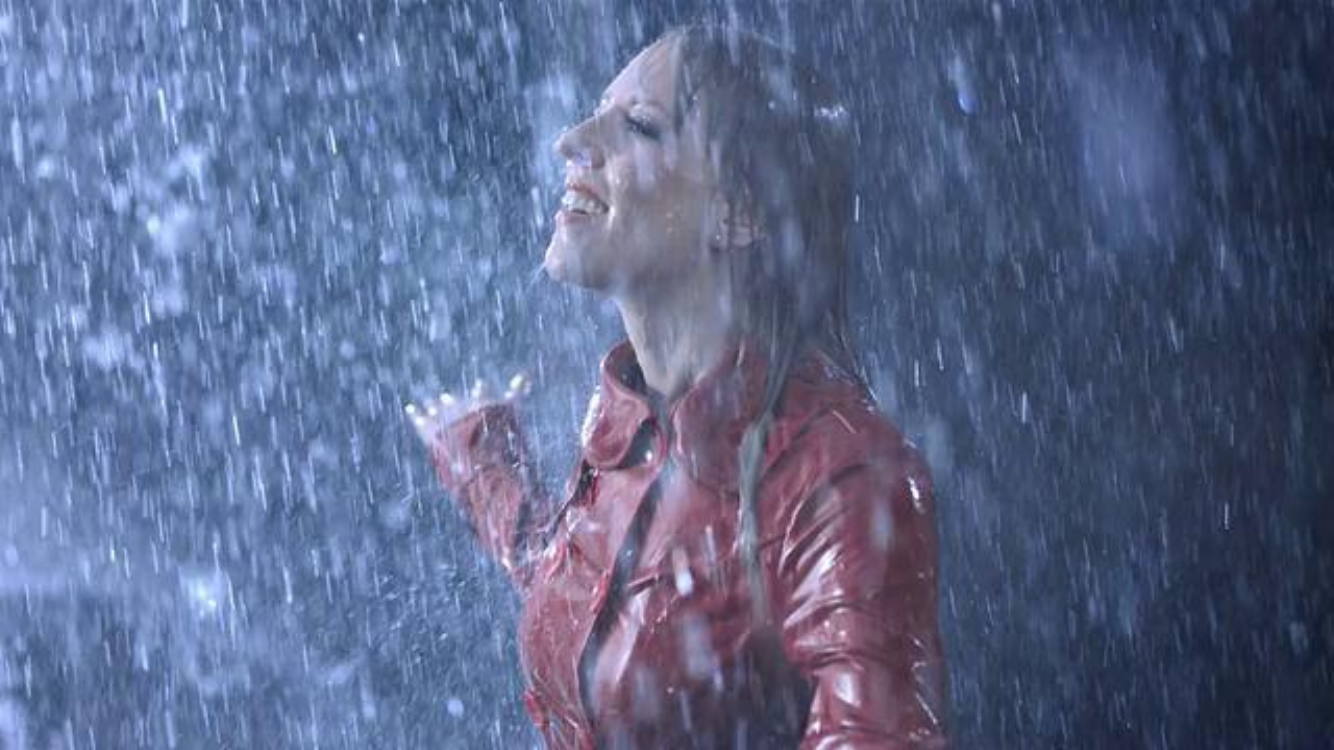}&
		\includegraphics[height=0.088\linewidth]{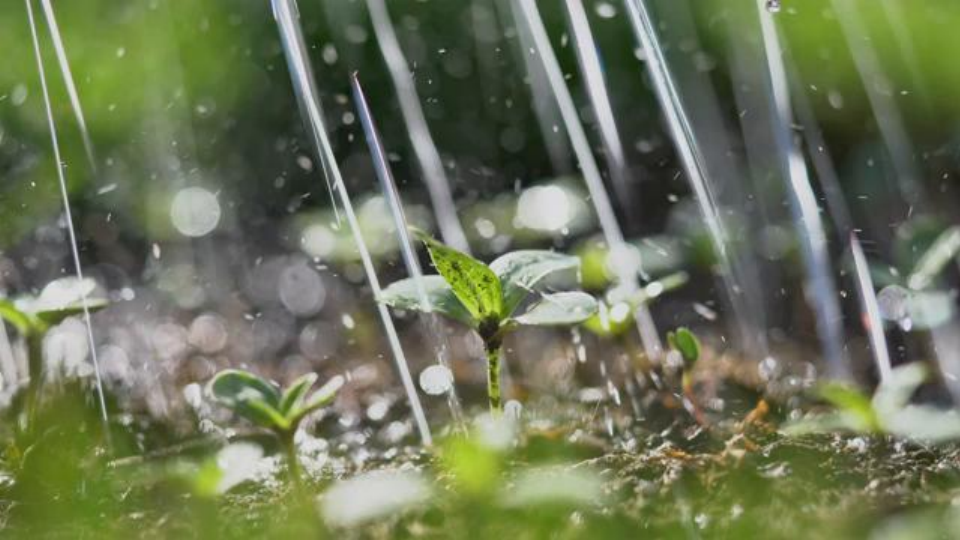}\\		
		\rotatebox	{90}{$\;\;\;\;\,$\footnotesize S2VD}&	
		\includegraphics[height=0.088\linewidth]{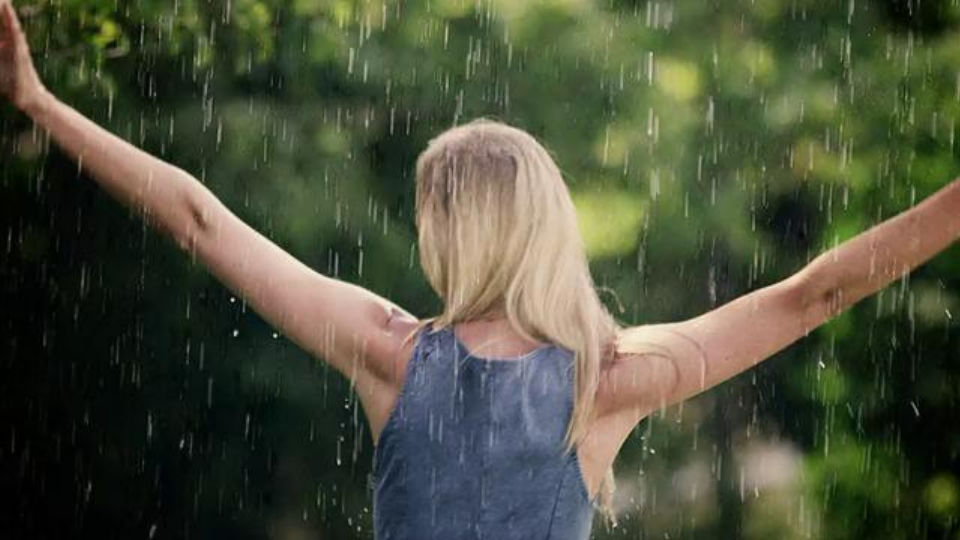}&
		\includegraphics[height=0.088\linewidth]{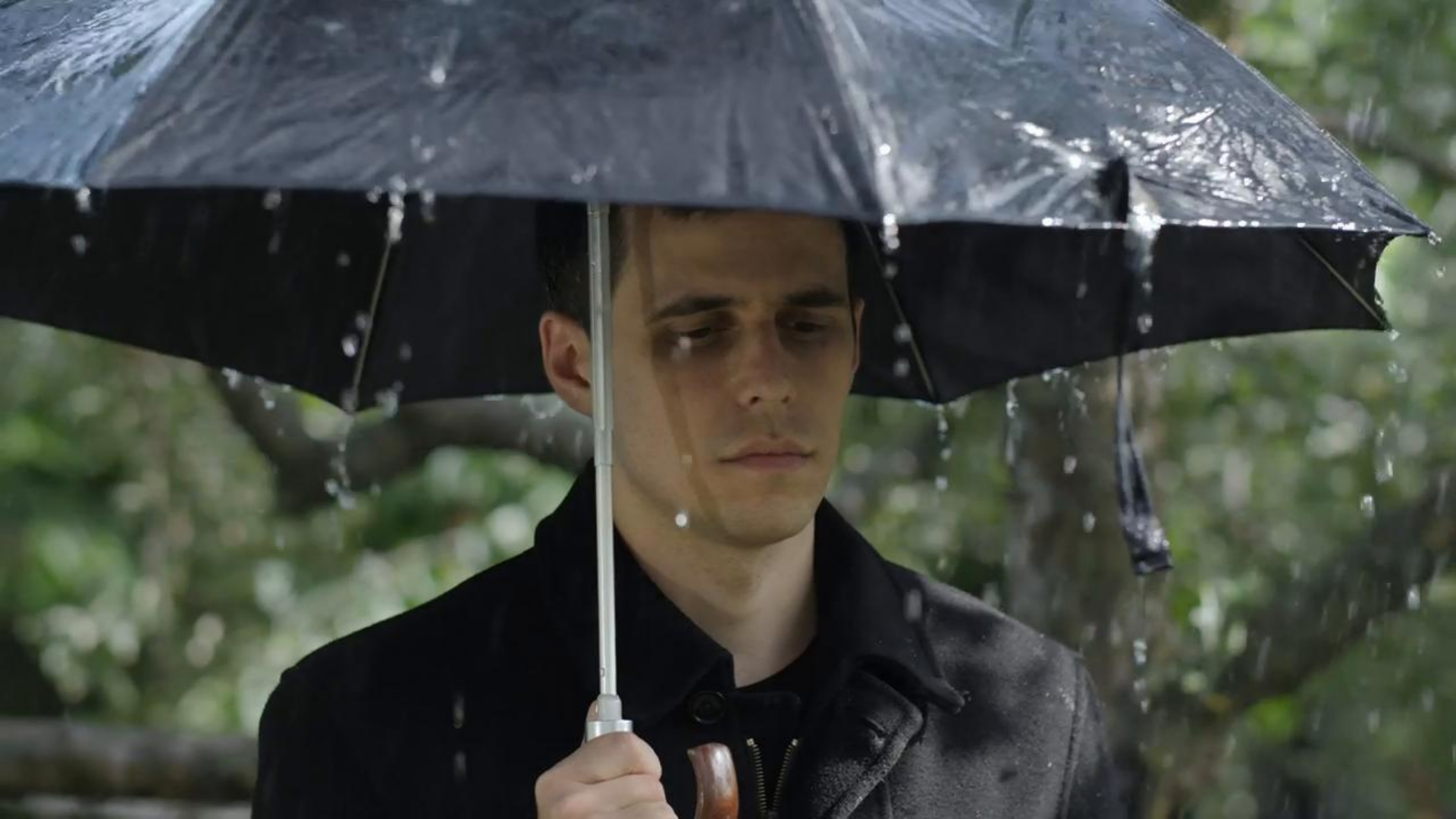}&
		\includegraphics[height=0.088\linewidth]{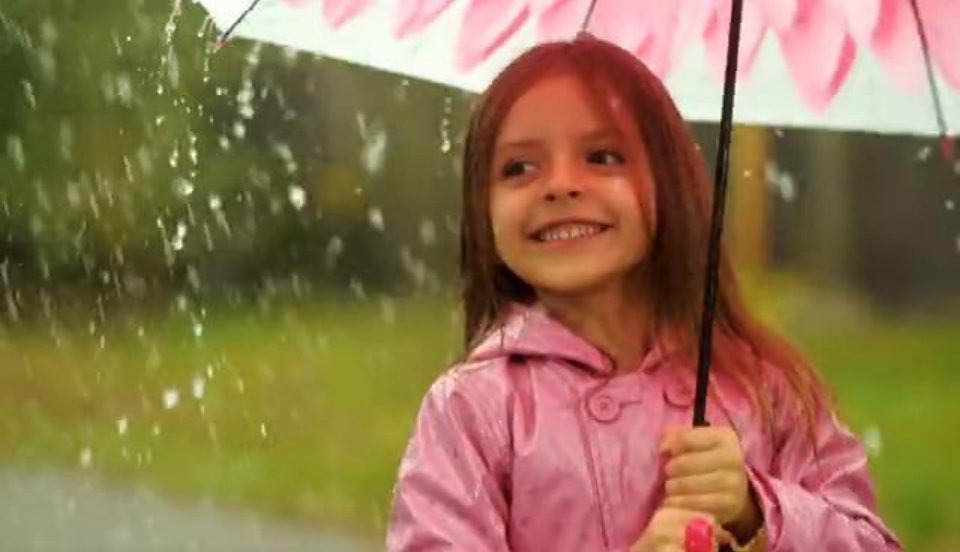}&
		\includegraphics[height=0.088\linewidth]{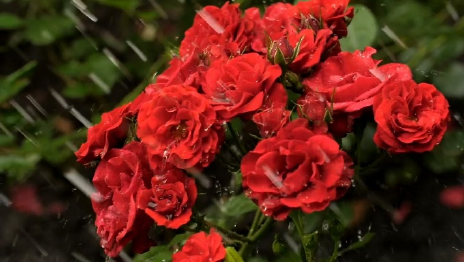}&
		\includegraphics[height=0.088\linewidth]{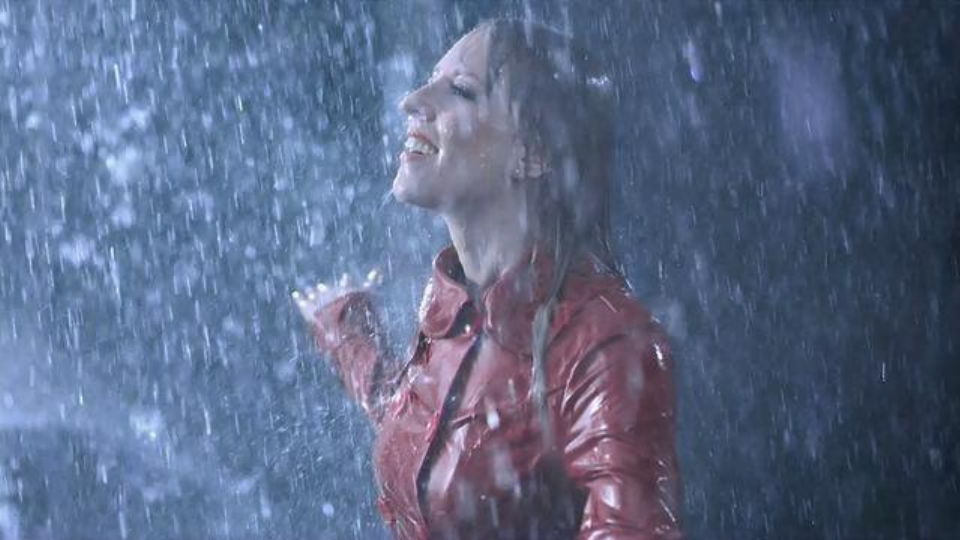}&
		\includegraphics[height=0.088\linewidth]{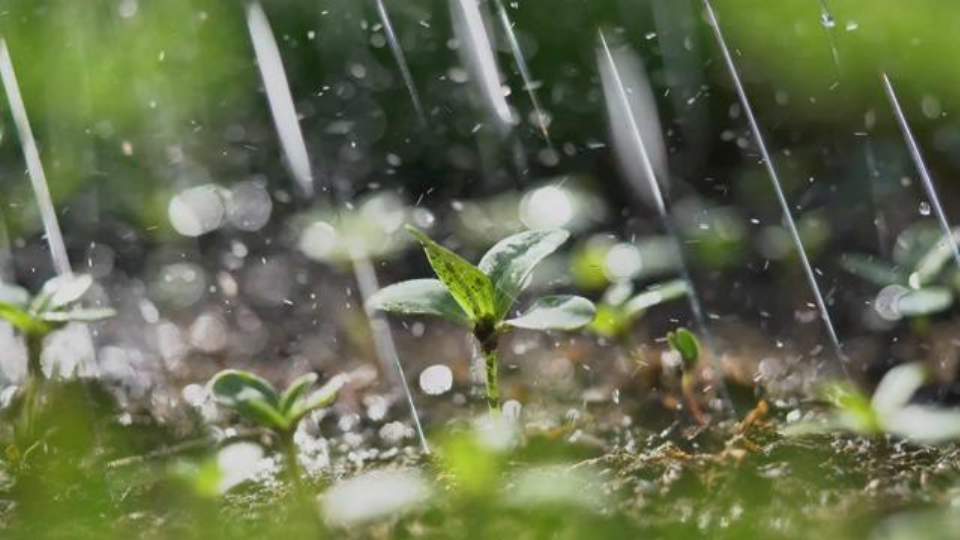}\\		
		\rotatebox	{90}{$\;\;\;\;\;$\footnotesize Ours}&	
		\includegraphics[height=0.088\linewidth]{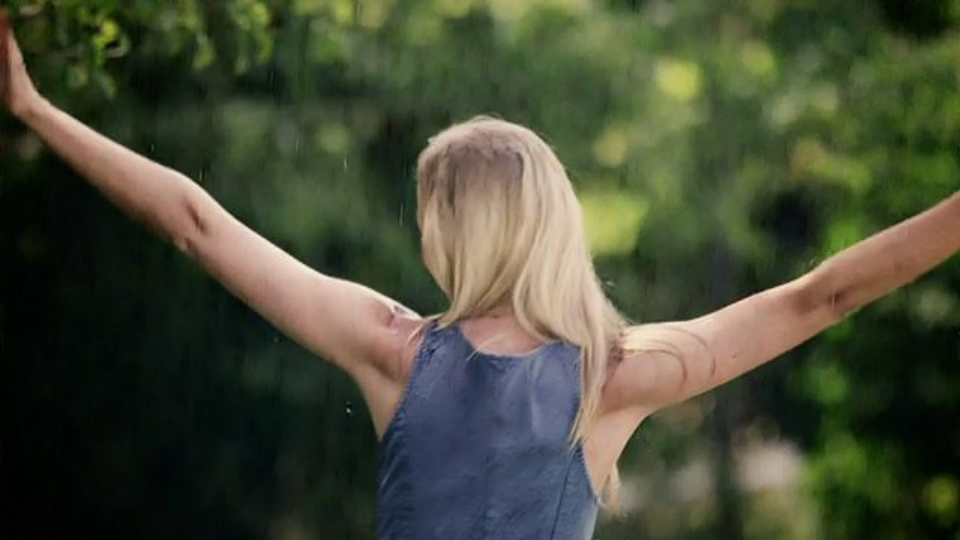}&
		\includegraphics[height=0.088\linewidth]{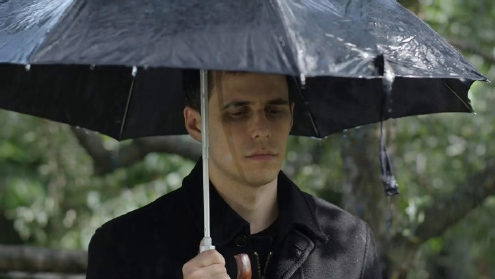}&
		\includegraphics[height=0.088\linewidth]{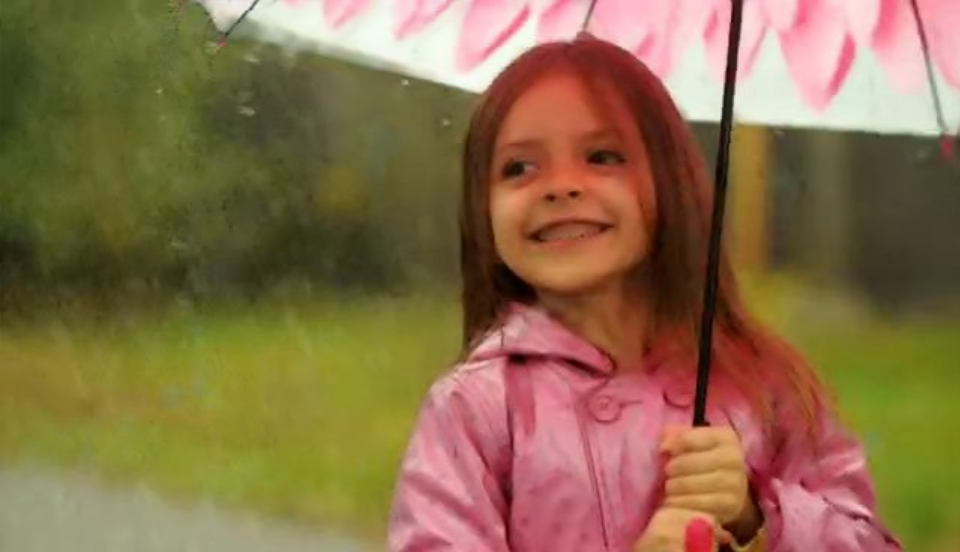}&
		\includegraphics[height=0.088\linewidth]{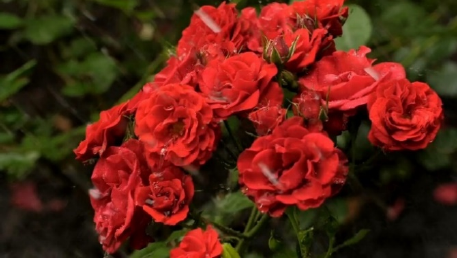}&
		\includegraphics[height=0.088\linewidth]{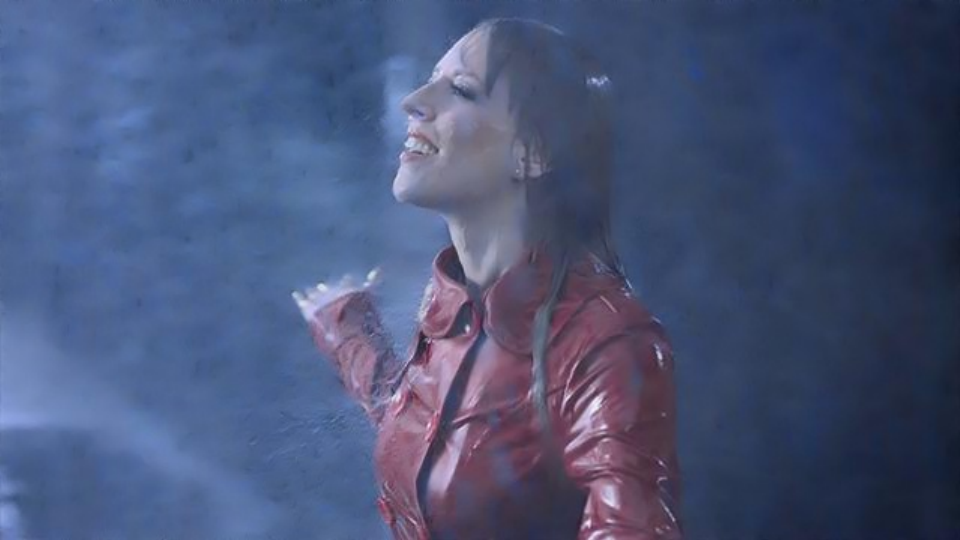}&
		\includegraphics[height=0.088\linewidth]{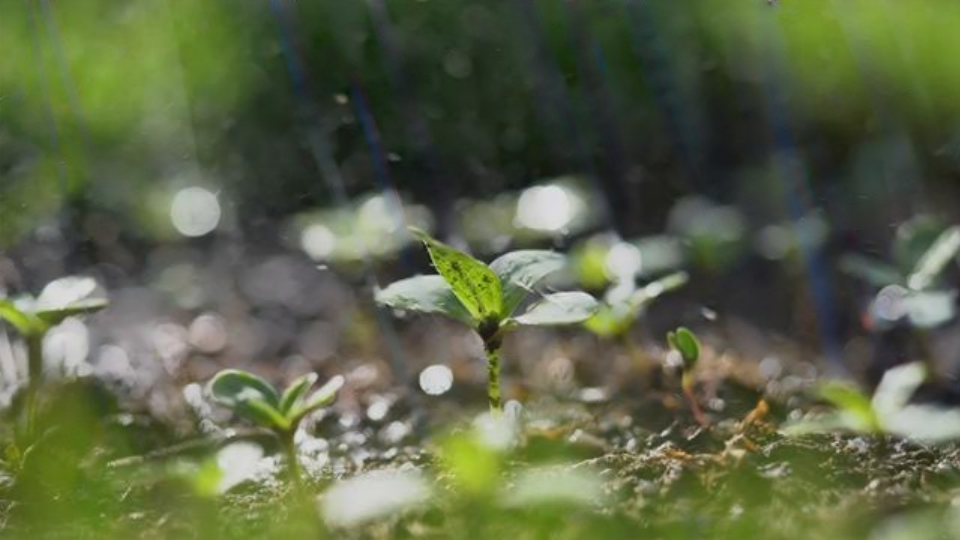}\\

	\end{tabular}	
	\caption{More results on different real scenarios among S2VD and our ASF-Net. These scenarios cover different types of rain patterns and multiple video backgrounds, providing more details for visual comparison.}
	\label{fig:morereal}
	\vspace{-0.4cm}
\end{figure*}

\subsection{Qualitative Evaluation}
Furthermore, in~\cref{fig:fig_other} and~\cref{fig:fig_ours}, we compare our visual results with different methods. 
%The top two rows of ~\cref{fig:fig_our} provide two examples of our \emph{LARA} dataset. Our results remove all the streaks and remain fewer artifacts. 
~\cref{fig:fig_other} shows the results on \emph{Light25} and \emph{Complex25}. Our method  can effectively remove rain streaks, while some other methods, such as SE, MSCSC, and FastDerain work less especially on \emph{Complex25} dataset.~\cref{fig:fig_ours} provides four examples of our \emph{LARA} dataset. Our results remove all the streaks and remain fewer artifacts. And compared with~\cref{fig:fig_other}, our dataset has more obvious effect on various rain removal methods and covers more scenes. The recovered videos under our method have higher quality visual effects. As shown in~\cref{fig:fig_ours}, the deraining performance of latest video deraining method, such as S2VD~\cite{yue2021semi} and RDD~\cite{wang2022rethinking}, is unstable and leaves noticeable rain streaks in certain scenarios. 

As shown in \cref{fig:fig_real1}, our method also outperforms other state-of-the-art video deraining methods in the real-world scenario. Compared to other deraining methods whose results still contain many obvious streaks or remain artifacts, our method provides more effective deraining results with fewer remaining streaks and fewer artifacts. Moreover, \cref{fig:morereal} provides more visual results of S2VD~\cite{yue2021semi} (the most advanced video deraining algorithm) and our method on different real scenarios. 

%Besides, to better verify the generalization ability of the model, we apply this method to the video snow removal task.~\cref{fig:snow} shows the visual effects after snow removal corresponding to different methods under the real snow scenarios. Among them, S2VD~\cite{yue2021semi}, DIP~\cite{jiang2017novel}, FastDerain~\cite{jiang2018fastderain} are rain removal methods, and JSTASR~\cite{2020JSTASR} is the latest image snow removal work.~\cref{fig:snow} shows that our method also performs well on the migration of the snow removal task. It can also be seen that our proposed synthetic-to-real online re-degraded learning strategy has a strong generalization ability and can achieve better results on different tasks. 

\section{Analysis and Discussion}
\label{sec:ablation}

\subsection{The Necessity of LARA}
\label{ssec:dataset_study}
\begin{figure}[t]
	\centering
	\begin{tabular}{c}
		\includegraphics[width=0.98\linewidth]{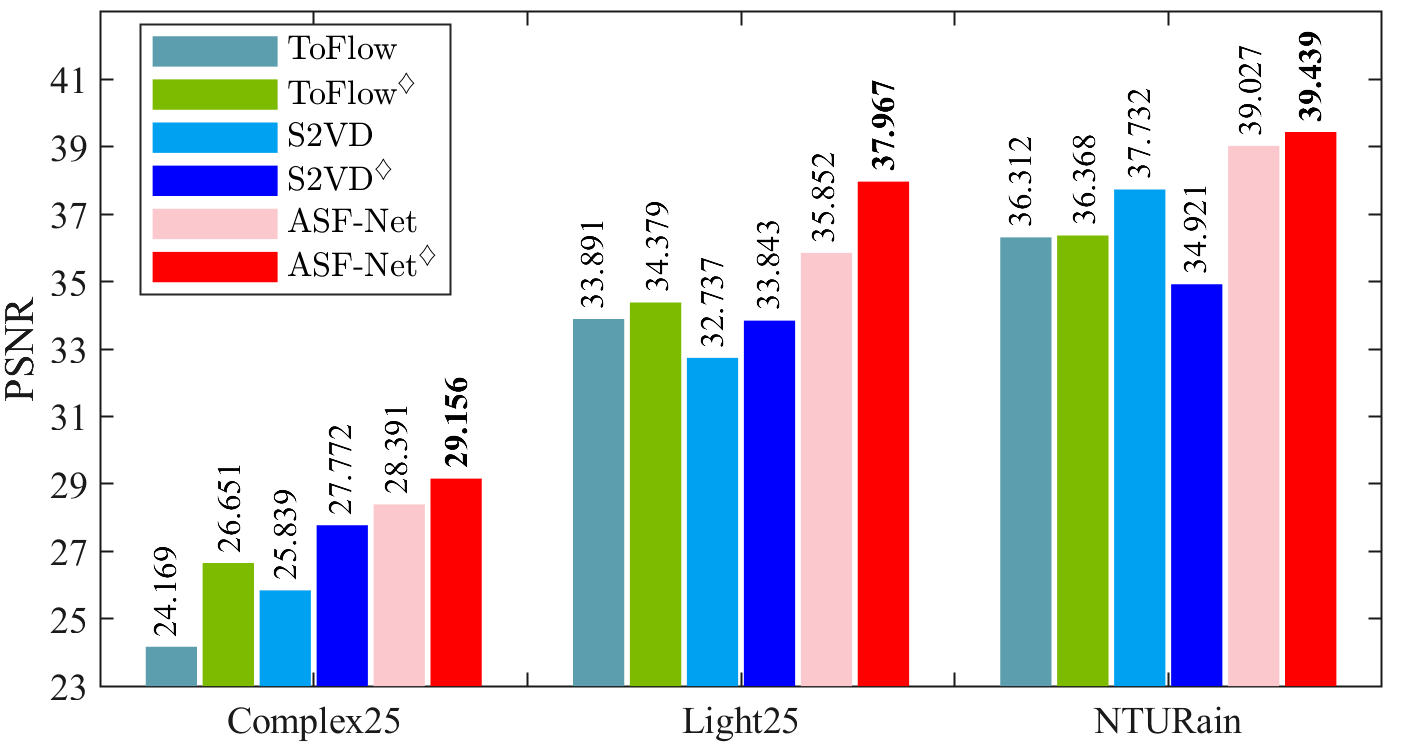}\\
	\end{tabular}
	\caption{Effects of our LARA dataset. ToFlow~\cite{xue2019video}, S2VD~\cite{yue2021semi}, and ASF-Net represent that they are trained on existing datasets. $\Diamond$ denotes that it is trained on our {LARA} dataset. }
	\label{fig:ablation_bar}
	\vspace{-0.3cm}
\end{figure}

\begin{figure}[t]
	\centering
	\begin{minipage}{0.18\textwidth}
			\begin{minipage}{1\textwidth}
				\includegraphics[width=1\textwidth]{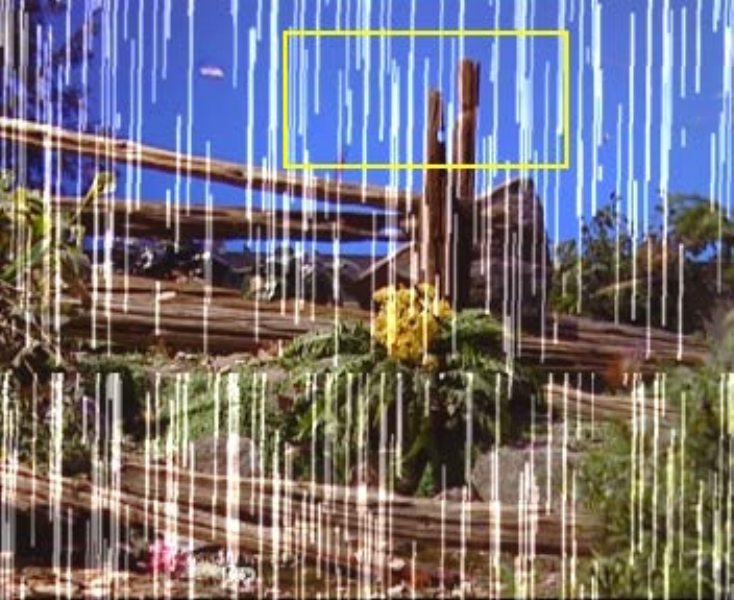}
				\centering \footnotesize Input \\ %\vspace{+0.1em} \\
			\end{minipage}
	\end{minipage} 
	\begin{minipage}{0.135\textwidth}
			\begin{minipage}{1\textwidth}
				\includegraphics[width=1\textwidth]{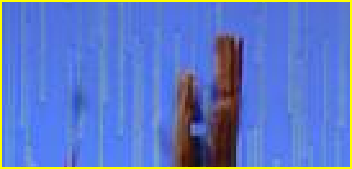}
				\centering \footnotesize ToFlow$^\Diamond$ \\ %\vspace{+0.2em} \\
			\end{minipage}
			\begin{minipage}{1\textwidth}
				\includegraphics[width=1\textwidth]{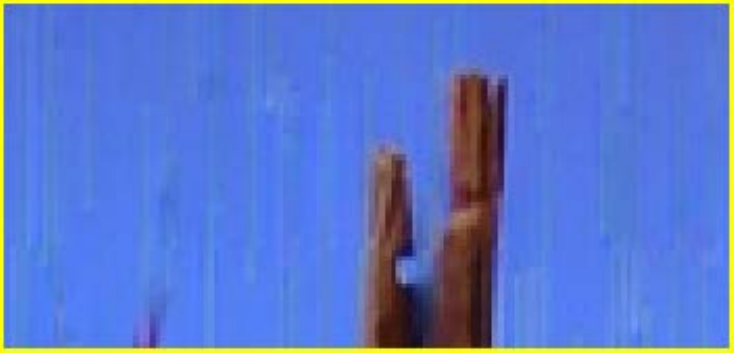}
				\centering \footnotesize Ours \\ %\vspace{+0.2em}\\
			\end{minipage}
	\end{minipage} 
	\begin{minipage}{0.135\textwidth}
			\begin{minipage}{1\textwidth}
				\includegraphics[width=1\textwidth]{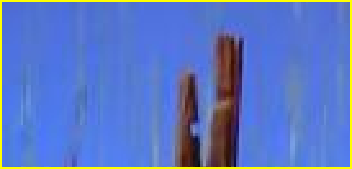}
				\centering \footnotesize S2VD$^\Diamond$ \\ %\vspace{+0.2em}\\
			\end{minipage}
			\begin{minipage}{1\textwidth}
				\includegraphics[width=1\textwidth]{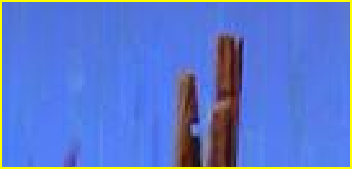}
				\centering \footnotesize Ours$^\Diamond$ \\ %\vspace{+0.2em}\\
			\end{minipage}
	\end{minipage} 
	%\vspace{-0.2cm}
	\caption{Visual comparison among different cases in our LARA dataset. Zoom in for better view of specific rain details. All settings are the same with the form in Figure~\ref{fig:ablation_bar}.}
	\label{fig:ablation_bar_h}
	\vspace{-0.6cm}
\end{figure}

To evaluate the effectiveness of LARA dataset, we analyze the results on existing three datasets. We choose three methods which are trained on LARA dataset or existing datasets to analyze the effectiveness of LARA dataset in~\cref{fig:ablation_bar}. ToFlow~\cite{xue2019video} is designed for video enhancement tasks.  S2VD~\cite{yue2021semi} and our network are video deraining methods. As shown in~\cref{fig:ablation_bar}, all three methods trained on our LARA dataset can outperform the model trained on existing datasets except for S2VD trained on \emph{NTURain}. Because we do not retrain S2VD on \emph{NTURain} but use the released model parameters to test the results of \emph{NTURain}.~\cref{fig:ablation_bar_h} shows that our method trained on our LARA dataset can remove the most rain streaks and remain fewer artifacts. In a word, the results in~\cref{fig:ablation_bar} illustrate the effectiveness of our LARA dataset and also demonstrate that improving the quality of dataset is rational and helpful. 

It should be mentioned that the results of our method in~\cref{fig:ablation_bar} are a little inferior to the results shown in~\cref{tab:value_LH}, since we do not adopt the ORL strategy (~\cref{sec:mda}) in~\cref{fig:ablation_bar}. And the analysis of ORL strategy will be shown in~\cref{ssec:strategy_study}. 

\subsection{Effects of ASF-Net}
\label{ssec:net_study}
Based on the initial analysis of the shift module by~\ref{ssec:shift}, we further quantitatively analyzed our ASF-Net in~\cref{tab:net_study}. It is observed that pre-aligned by optical flow can significantly improve the objective performance ($\mathbf{M}_a$ vs. $\mathbf{M}_b$). Offsets estimated by dilated convolution can also benefit rain streak removal ($\mathbf{M}_b$ vs. $\mathbf{M}_c$). Besides, the temporal shift module also helps achieve better deraining performance ($\mathbf{M}_c$ vs. $\mathbf{M}_d$). 
Comparatively, our full version of ASF-Net can achieve the best deraining performance. 

%\begin{table}[t]	
%	\setlength{\tabcolsep}{2.8mm}
%	\caption{Ablation study on different settings of temporal alignment and shift.}
%	\centering
 %    \renewcommand\arraystretch{1.25}
%	\begin{tabular}{|c|c | c |c |c |}
%		\hline 
%		\footnotesize \footnotesize Baseline &\footnotesize $\mathbf{M}_a$ &\footnotesize $\mathbf{M}_b$  &\footnotesize $\mathbf{M}_c$ &\footnotesize $\mathbf{M}_d$ \\
%		\hline
%		\footnotesize Deformable Conv. &\footnotesize$\checkmark$ &\footnotesize $\checkmark$ &\footnotesize $\checkmark$ &\footnotesize $\checkmark$ \\
%		\footnotesize Optical Flow &\footnotesize &\footnotesize	$\checkmark$ &\footnotesize $\checkmark$ &\footnotesize $\checkmark$  \\
%		\footnotesize Dilate Conv. &\footnotesize &\footnotesize & \footnotesize $\checkmark$ &\footnotesize $\checkmark$	  \\
%		\footnotesize Temporal Shift	&\footnotesize &\footnotesize &\footnotesize &\footnotesize $\checkmark$  \\
%		\hline
%		\footnotesize PSNR   &\footnotesize 27.522 &\footnotesize 28.265 &\footnotesize 28.639   &\footnotesize \textbf{29.156}\\
%		\footnotesize SSIM  &\footnotesize 0.806 &\footnotesize 0.829 &\footnotesize 0.845  &\footnotesize \textbf{0.876}\\
%		\hline
%	\end{tabular}
%	\label{tab:net_study}
%\end{table}

\subsection{Effects of ORL}
\label{ssec:strategy_study}
\begin{figure}[!t]
	\centering
	\begin{tabular}{c@{\extracolsep{0.2em}}c@{\extracolsep{0.2em}}c@{\extracolsep{0.2em}}c}
\includegraphics[width=0.23\linewidth]{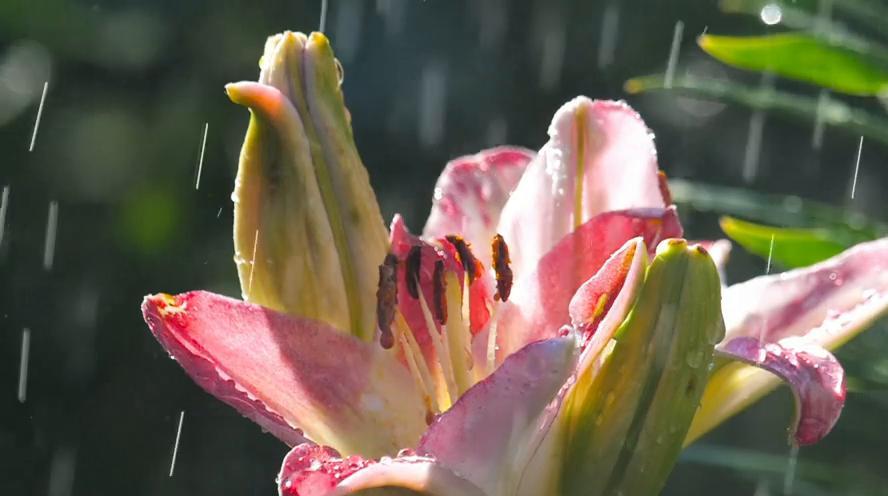}
		&\includegraphics[width=0.23\linewidth]{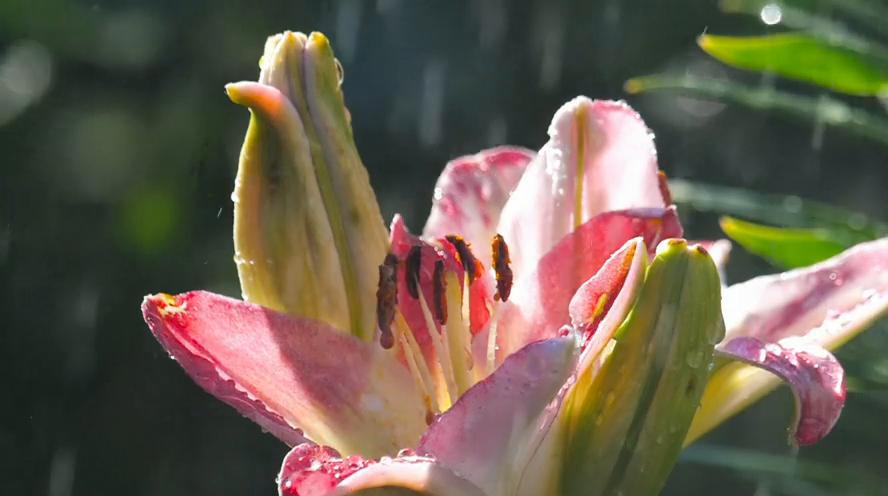}
		&\includegraphics[width=0.23\linewidth]{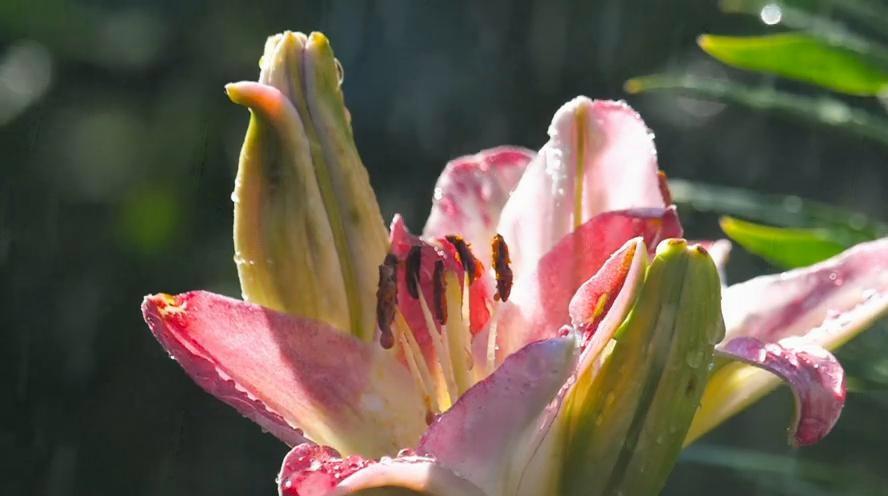}
		&\includegraphics[width=0.23\linewidth]{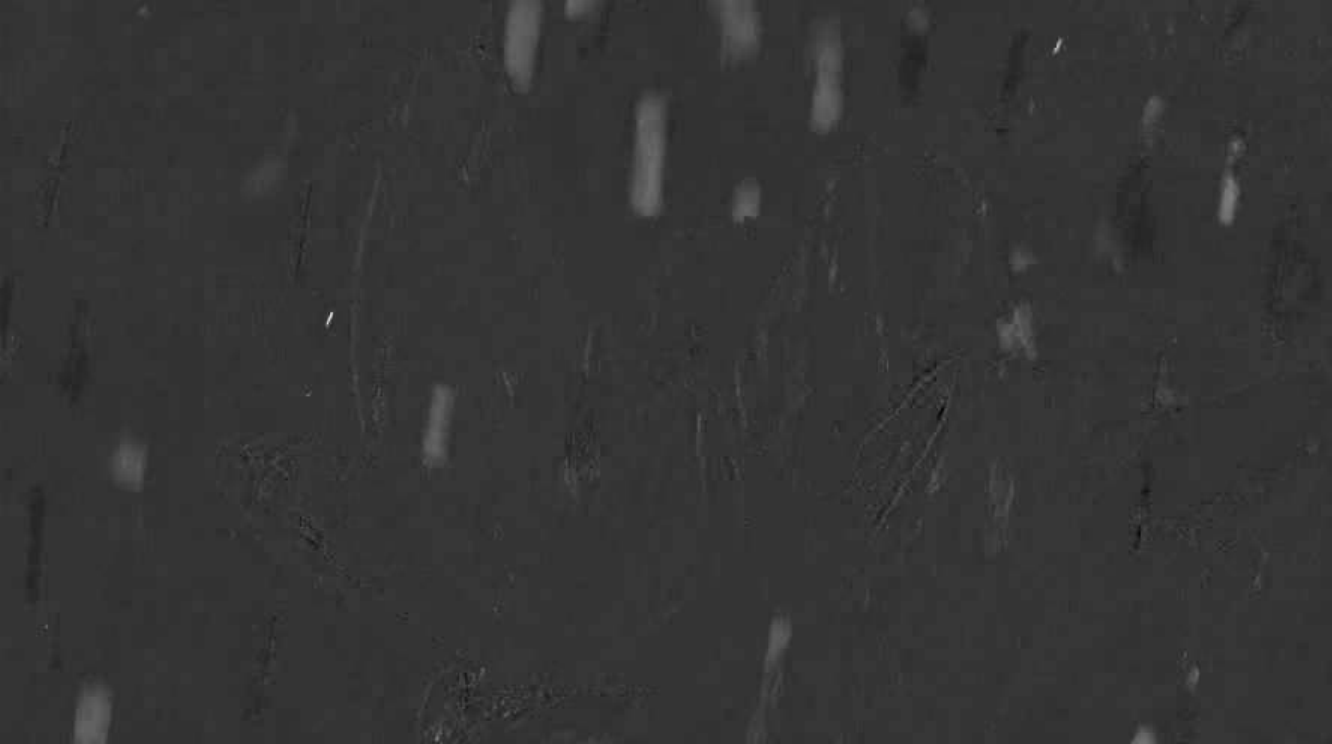}\\
		\includegraphics[width=0.23\linewidth]{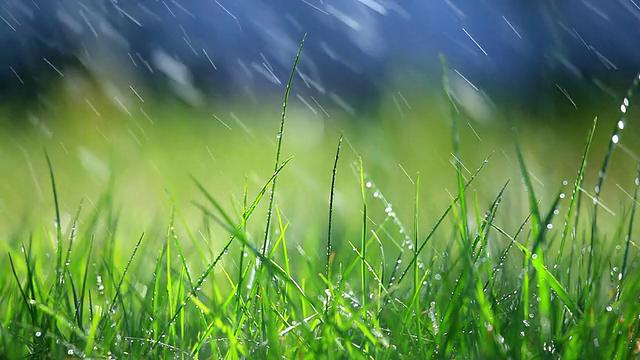}
		&\includegraphics[width=0.23\linewidth]{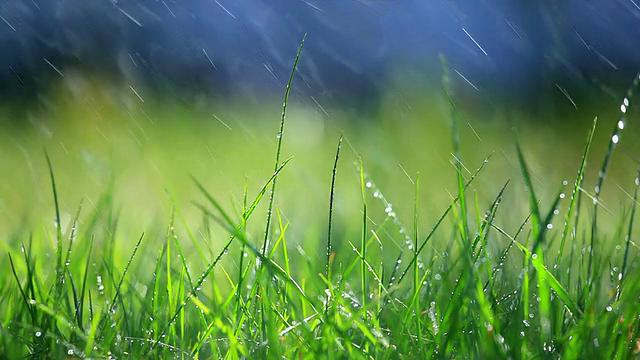}
		&\includegraphics[width=0.23\linewidth]{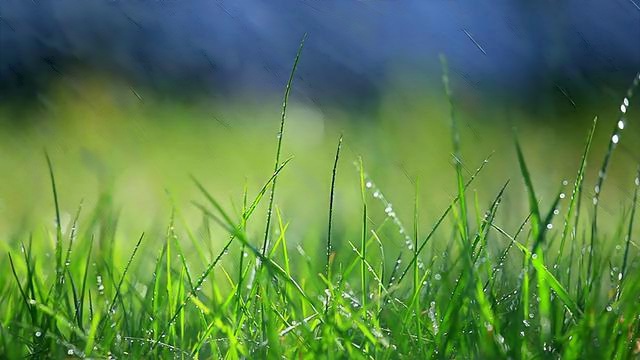}
		&\includegraphics[width=0.23\linewidth]{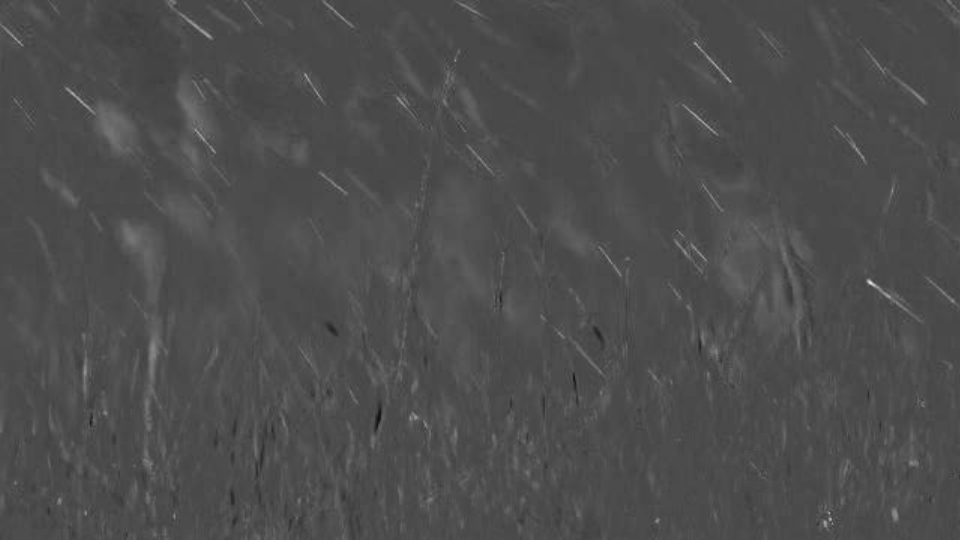}\\
		\footnotesize Input &\footnotesize w/o ORL &\footnotesize w/ ORL &\footnotesize Difference\\			
	\end{tabular}
	\caption{Effects of ORL. The last column shows the difference between the input and w/ ORL. }
	\label{fig:strategy_study}
	\vspace{-0.4cm}
\end{figure}

The ORL strategy can exploit the useful information from both synthesized paired data and real data. As shown in~\cref{tab:strategy_study}, with the help of ORL, our method can obtain about 1.65db gain in PSNR ($\mathbf{P}_a$ vs. $\mathbf{Ours}$). Besides, we also evaluate the influence of real rain data. Method $\mathbf{P}_b$ trains ASF-Net on synthetic data with ORL. Method $\mathbf{P}_c$ denotes that we use real data of \emph{NTURain} to train ASF-Net based on ORL strategy. Our method outperforms $\mathbf{P}_b$ and $\mathbf{P}_c$ in PSNR, which shows the superiority of our real dataset for the ORL strategy. And \cref{fig:strategy_study} shows some visual examples in real scenarios to illustrate that ORL strategy can remove rain streaks effectively. 

\begin{table}[t]
	%	\vspace{-0.2cm}
	%\renewcommand\arraystretch{1.3} 	 	
	\setlength{\tabcolsep}{6mm}
	\caption{Effects of ORL. $\mathbf{P}_a$ only trained ASF-Net on synthetic data without ORL, $\mathbf{P}_b$ trained ASF-Net on synthetic data with ORL, $\mathbf{P}_c$ trained ASF-Net on real-world data of \emph{NTURain} with ORL.}
	\centering
     \renewcommand\arraystretch{1.2}
	\begin{tabular}{|l|@{\extracolsep{2em}}c@{\extracolsep{2em}} c@{\extracolsep{2em}} c@{\extracolsep{2em}} c|}
		\hline 
		\footnotesize Metric &\footnotesize $\mathbf{P}_a$ &\footnotesize $\mathbf{P}_b$ &\footnotesize $\mathbf{P}_c$ &\footnotesize $\mathbf{Ours}$ \\
		\hline
		\footnotesize PSNR &\footnotesize 29.156 &\footnotesize 30.385  &\footnotesize 29.884 &\footnotesize \textbf{30.805}   \\
		\footnotesize SSIM &\footnotesize 0.876 &\footnotesize 0.890 &\footnotesize 0.886 &\footnotesize \textbf{0.904}    \\
		\hline
	\end{tabular}
	\label{tab:strategy_study}
	\vspace{-0.2cm}
\end{table}

\begin{table}[t]	
	\setlength{\tabcolsep}{2.8mm}
	\caption{Ablation study on different settings of temporal alignment and shift.}
	\centering
	\renewcommand\arraystretch{1.25}
	\begin{tabular}{|c|c | c |c |c |}
		\hline 
		\footnotesize \footnotesize Baseline &\footnotesize $\mathbf{M}_a$ &\footnotesize $\mathbf{M}_b$  &\footnotesize $\mathbf{M}_c$ &\footnotesize $\mathbf{M}_d$ \\
		\hline
		\footnotesize Deformable Conv. &\footnotesize$\checkmark$ &\footnotesize $\checkmark$ &\footnotesize $\checkmark$ &\footnotesize $\checkmark$ \\
		\footnotesize Optical Flow &\footnotesize &\footnotesize	$\checkmark$ &\footnotesize $\checkmark$ &\footnotesize $\checkmark$  \\
		\footnotesize Dilate Conv. &\footnotesize &\footnotesize & \footnotesize $\checkmark$ &\footnotesize $\checkmark$	  \\
		\footnotesize Temporal Shift	&\footnotesize &\footnotesize &\footnotesize &\footnotesize $\checkmark$  \\
		\hline
		\footnotesize PSNR   &\footnotesize 27.522 &\footnotesize 28.265 &\footnotesize 28.639   &\footnotesize \textbf{29.156}\\
		\footnotesize SSIM  &\footnotesize 0.806 &\footnotesize 0.829 &\footnotesize 0.845  &\footnotesize \textbf{0.876}\\
		\hline
	\end{tabular}
	\label{tab:net_study}
	\vspace{-0.2cm}
\end{table}
%\subsection{Limitation}
%Although we have achieved superior video deraining performance as shown above, our method may fail in some heavy rain scenarios. As shown in~\cref{fig:limitation}, Our method and other latest video deraining method~\cite{yang2020self,yue2021semi} can  not effectively remove the real heavy rain streaks and obvious artifacts are left in the results.

%\begin{figure}[!t]
%	\centering
%	\begin{tabular}{c@{\extracolsep{0.3em}}c@{\extracolsep{0.3em}}c@{\extracolsep{0.3em}}c}
%		\includegraphics[width=0.232\linewidth]{Figures/Real16/rfc5}
%		&\includegraphics[width=0.232\linewidth]{Figures/Real16/sldnet}
%		&\includegraphics[width=0.232\linewidth]{Figures/Real16/s2vd}
%		&\includegraphics[width=0.232\linewidth]{Figures/Real16/cvpr22}
%		\\
%		\footnotesize Input &\footnotesize SLDNet &\footnotesize S2VD &\footnotesize  Ours \\			
%	\end{tabular}
%	\caption{Limitation in heavy rain scenarios. There are still rain streaks remnants and blurring of detail information. }
%	\label{fig:limitation} 
%	\vspace{-0.6cm}
%\end{figure}

\section{Conclusion}
%In this paper, we establish a new video deraining paradigm, alignment-shift-fusion, to effectively exploit temporal information among neighbor frames. And we  propose a new online synthetic-to-real learning strategy to compensate for domain gap between synthetic and real scenarios.
We introduce a new video deraining paradigm, alignment-shift-fusion, which effectively utilizes the temporal information among adjacent frames. Additionally, we propose a novel online re-degraded learning strategy to address the domain gap between synthetic and real scenarios. 
%Extensive experiments have verified the superiority of the ASF-Net and the online synthetic-to-real learning strategy. 
Our extensive experiments have demonstrated the superior performance of ASF-Net, the architecture built on the proposed paradigm, and the online re-degraded learning strategy. 
%We believe that our proposed paradigm and the built dataset can boost the development of video deraining research. Last but not least, our online synthetic-to-real learning strategy can also be applied to various image or video degradation problems to help compensate for the domain gap from different scenes. 
In summary, with our proposed paradigm and the conducted dataset, it is expected to make a significant contribution to the field of video deraining, and the broad potential in our online synthetic-to-real learning strategy can be useful in a variety of related research areas. 

%\begin{thebibliography}{1}
\bibliographystyle{IEEEtran}
\bibliography{egbib}

\end{document}